\documentclass[preprints,review,accept,moreauthors,pdftex]{Definitions/mdpi} 
\usepackage{longtable}
\usepackage{supertabular}

 \firstpage{1} 
\pubvolume{xx}
\issuenum{1}
\articlenumber{5}
\pubyear{2020}
\copyrightyear{2020}
\history{Received: date; Accepted: date; Published: date}

\pdfoutput=1

\Title{A Survey on Machine Reading Comprehension: Tasks, Evaluation Metrics and Benchmark Datasets}

\Author{Changchang Zeng $^{1,2,6}$, Shaobo Li $^{1,3,*}$, Qin Li $^{4}$, Jie Hu $^{4}$, and Jianjun  Hu $^{5} \orcidA{}$*}

\AuthorNames{Chengchang Zeng, Shaobo Li, Qin Li, Jie Hu, and Jianjun Hu}

\address{%
$^{1}$ \quad Chengdu Institute of Computer Application, Chinese Academy of Sciences, Chengdu 610041, China; zengchangchang16@mails.ucas.ac.cn\\
$^{2}$ \quad University of Chinese Academy of Sciences, Beijing 100049, China\\
$^{3}$ \quad School of Mechanical Engineering, Guizhou University, Guiyang 550025, China; lishaobo@gzu.edu.cn\\
$^{4}$ \quad College of Big Data Statistics, GuiZhou University of Finance and Economics, Guiyang, Guizhou 550025, China; jiehu@mail.gufe.edu.cn\\
$^{5}$ \quad Department of Computer Science and Engineering, University of South Carolina, Columbia, SC 29208, USA; jianjunh@cse.sc.edu\\
$^{6}$ \quad Department of Computer Science and Engineering, Chengdu Neusoft University, Chengdu 611844, China; zengchangchang@nsu.edu.cn

}

\corres{Correspondence: jianjunh@cse.sc.edu; lishaobo@gzu.edu.cn; Tel.:  +1-803-777-7304(J.H.)}

\abstract{Machine Reading Comprehension (MRC) is a challenging Natural Language Processing(NLP) research field with wide real-world applications. The great progress of this field in recent years is mainly due to the emergence of large-scale datasets and deep learning. At present, a lot of MRC models have already surpassed human performance on various benchmark datasets despite the obvious giant gap between existing MRC models and genuine human-level reading comprehension. This shows the need for improving existing datasets, evaluation metrics, and models to move current MRC models toward "real" understanding. To address the current lack of comprehensive survey of existing MRC tasks, evaluation metrics, and datasets, herein, (1) we analyze 57 MRC tasks and datasets and propose a more precise classification method of MRC tasks with 4 different attributes; (2) we summarized 9 evaluation metrics of MRC tasks, 7 attributes and 10 characteristics of MRC datasets; (3) We also discuss key open issues in MRC research and highlighted future research directions. In addition, we have collected, organized, and published our data on the companion website(\href{https://mrc-datasets.github.io/}{https://mrc-datasets.github.io/}) where MRC researchers could directly access each MRC dataset, papers, baseline projects, and the leaderboard. }

\keyword{machine reading comprehension; survey; dataset}

\begin{document}



\section{Introduction}
\subsection{Overview}
In the long history of Natural Language Processing (NLP), teaching computers to read the text and understand the meaning of the text is a major research goal that has not been fully realized. In order to accomplish this task, researchers have conducted machine reading comprehension (MRC) research in many aspects recently with the emergence of the large-scale datasets, higher computing power, and the deep learning techniques, which have boosted the whole NLP research \cite{young2018recent,li2018deeppatent, li2018tourism}. The concept of MRC comes from the human understanding of text. The most common way to test whether a person can fully understand a piece of text is to require she/he answer questions about the text. Just like the human language test, reading comprehension is a natural way to evaluate a computer's language understanding ability.

In the NLP community, machine reading comprehension has received extensive attention in recent years \cite{boerma2016reading,sugawara2019assessing,zhang2019machine,baradaran2020survey, gupta2020conversational}. The goal of a typical MRC task is to require a machine to read a (set of) text passage(s) and then answers questions about the passage(s), which is very challenging \cite{gao-etal-2018-neural}.

\begin{figure}[H]
	\centering
	\includegraphics[width=15cm]{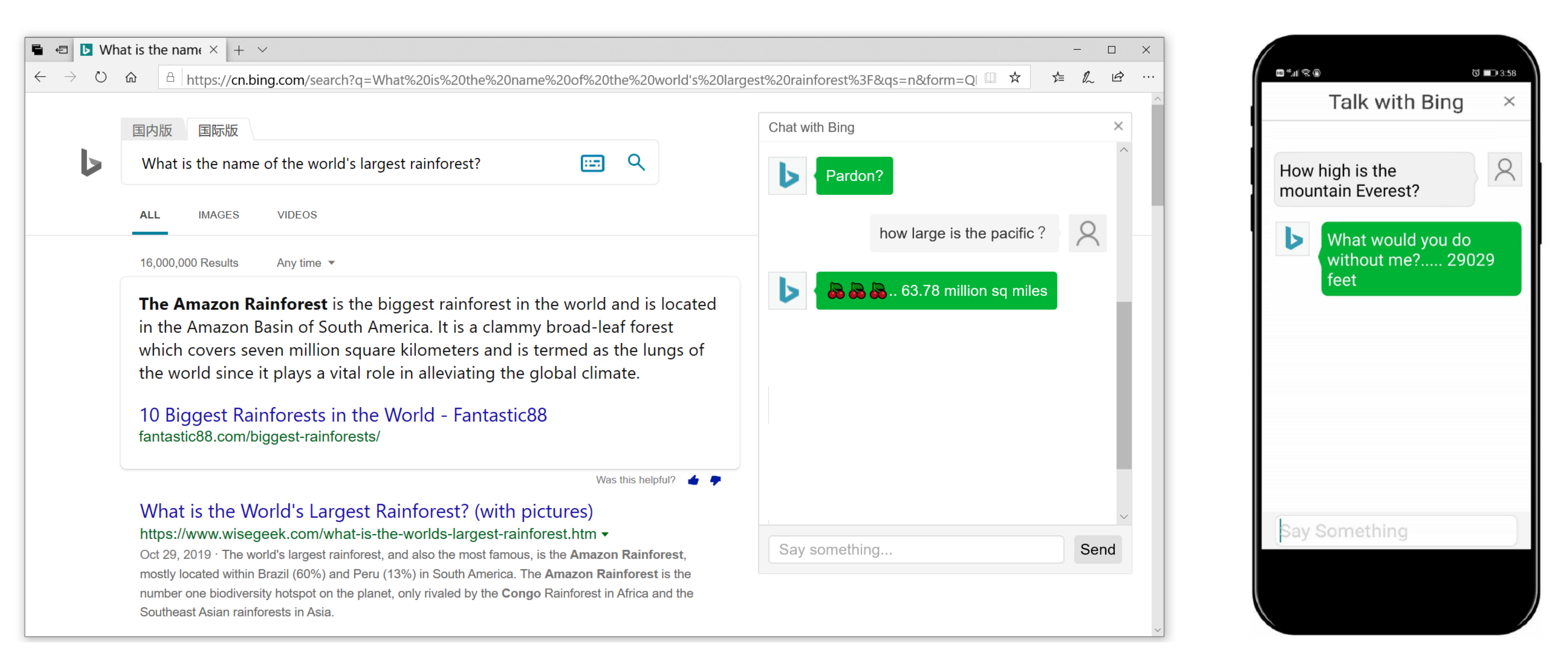}
	\caption{Examples of machine reading comprehension applied to search engine and dialogue system.}
	\label{figure:application}
\end{figure}

Machine reading comprehension could be widely applied in many NLP systems such as search engines and dialogue systems. For example, as shown in Figure \ref{figure:application}, nowadays, when we enter a question into the search engine Bing, sometimes the Bing can directly return the correct answer by highlight it in the context (if the question is simple enough). Moreover, if we open the "Chat with Bing" in the website of Bing, as shown in the right part of the browser in Figure \ref{figure:application}, we can also ask it questions such as "How large is the pacific?", the Bing chatbot will directly give the answer "63.78 million square miles". And on Bing's App, we can also open this "Chat with Bing", as shown in the right part of Figure \ref{figure:application}. It is clear that MRC can help improve the performances of search engines and dialogue systems, which can allow users to quickly get the right answer to their questions, or to reduce the workload of customer service staff. 

\subsection{History}
\begin{figure}[H]
	\centering
	\includegraphics[width=15cm]{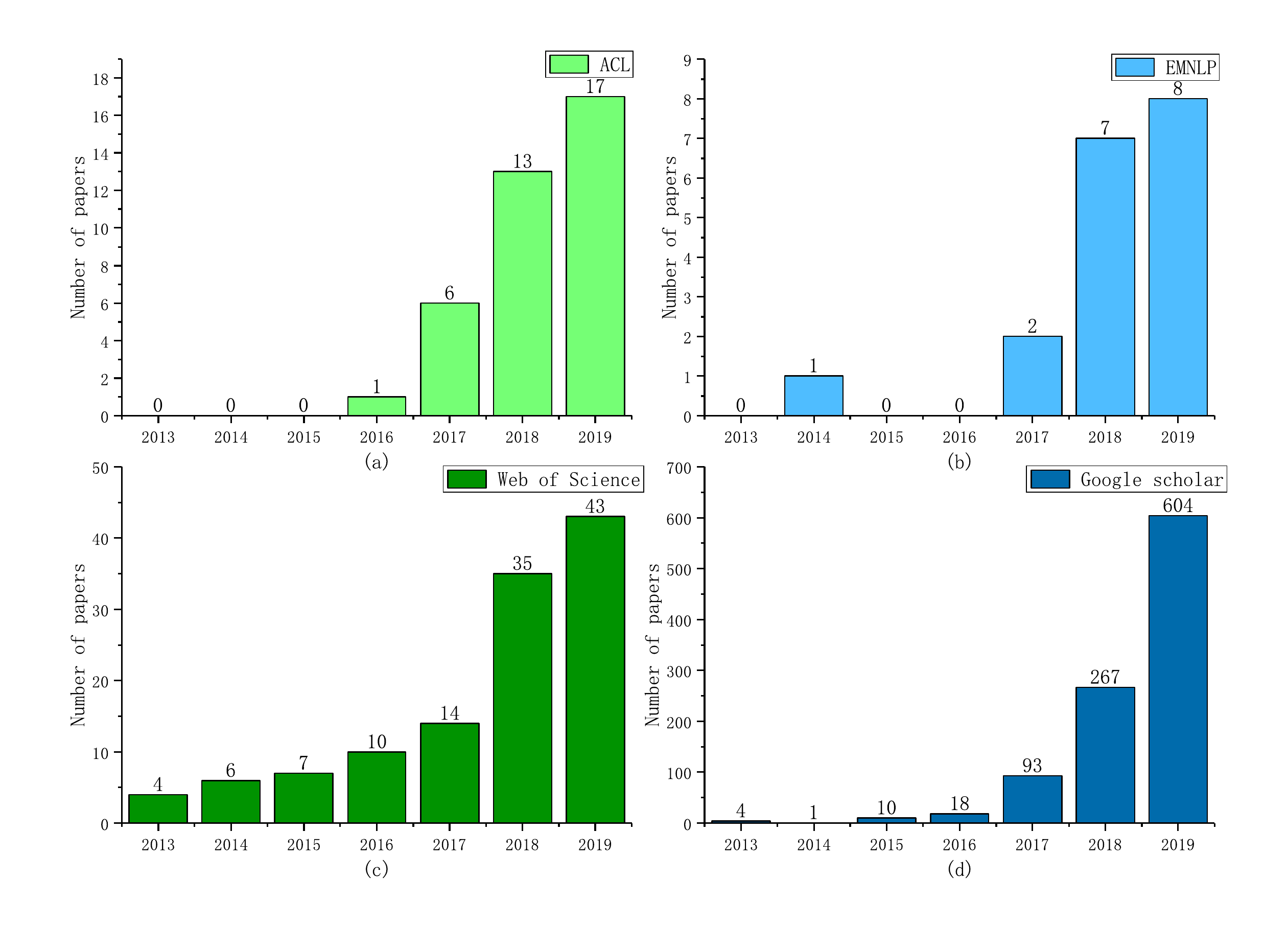}
	\caption{The number of research papers for machine reading comprehension each year: (\textbf{a}) The number of research papers on MRC in ACL from 2013 to 2019. (\textbf{b}) The number of research papers on MRC in ENMLP from 2013 to 2019. (\textbf{c}) The number of research papers on MRC in Web of Science from 2013 to 2019.(\textbf{d}) The number of research papers on MRC in Google scholar from 2013 to 2019.}
	\label{figure: paper }
\end{figure}

Machine reading comprehension is not newly proposed. As early as 1977, Lehnert et al. \cite{10.5555/908336} had already built a question answering program called the QUALM which was used by two story understanding systems. In 1999, Hirschman et al. \cite{hirschman-etal-1999-deep} constructed a reading comprehension system with a corpus of 60 development and 60 test stories of 3rd to 6th-grade material. The accuracy of the baseline system is between 30\% and 40\% on 11 sub-tasks. Most of MRC systems in the same period were rule-based or statistical models \cite{10.3115/1117595.1117598, charniak-etal-2000-reading}. However, due to the lack of high quality MRC datasets, this research field has been neglected for a long time \cite{chen2018neural}. In 2013, Richardson et al. \cite{richardson2013mctest} created the MCTest \cite{richardson2013mctest} dataset which contained 500 stories and 2000 questions. Later, many researchers began to apply machine learning models on MCTest \cite{richardson2013mctest,wang2015machine, sachan2015learning, narasimhan2015machine} despite that the original baseline of MCTest \cite{richardson2013mctest} is a rule-based model and the number of training samples in the MCTest \cite{richardson2013mctest} dataset is not large. A turning point for this field came in 2015 \cite{chen2018neural}. In order to resolve these bottlenecks, Hermann et al. \cite{hermann2015teaching} defined a new dataset generation method that provides large-scale supervised reading comprehension datasets in 2015. They also developed a class of attention based deep neural networks that learn to read real documents and answer complex questions with minimal prior knowledge of language structure. Since 2015, with the emergence of various large-scale supervised datasets and neural network models, the field of machine reading comprehension has entered a period of rapid development. Figure \ref{figure: paper } shows the numbers of research papers on MRC since 2013. As is seen, the number of papers on MRC has been growing at an impressive rate.

\subsection{Motivation}
\begin{figure}[H]
	\centering
	\includegraphics[width=15cm]{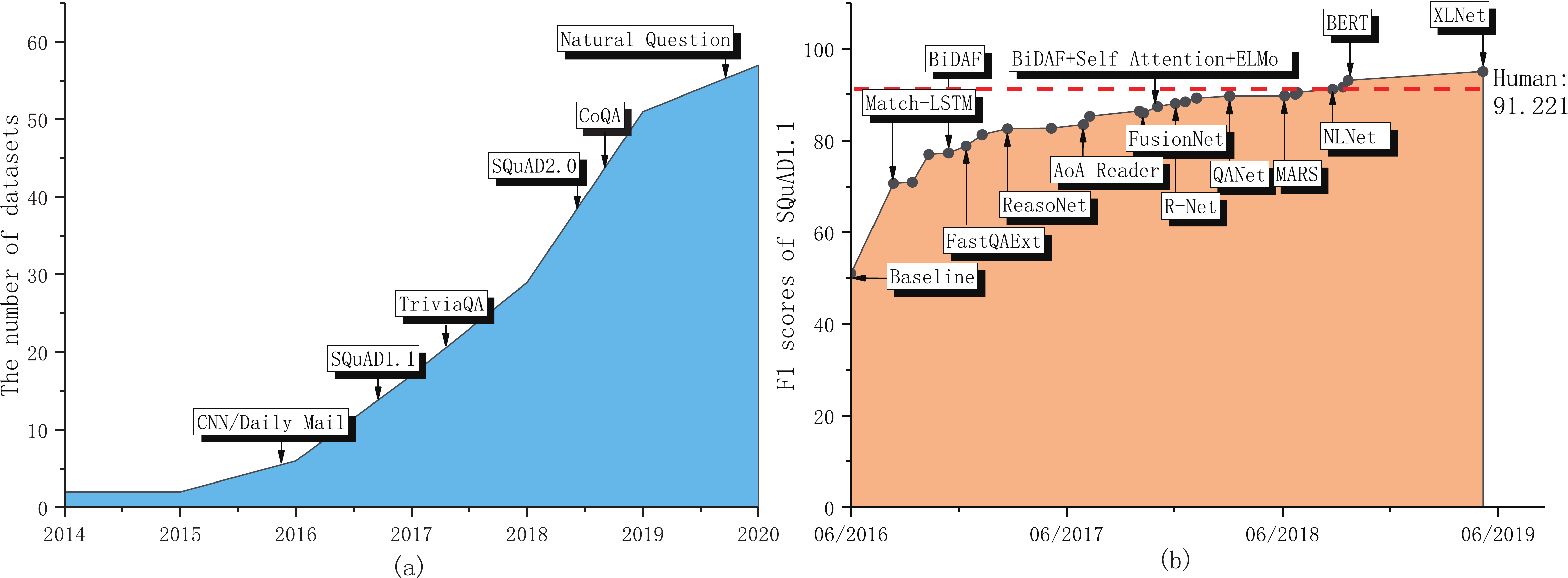}
	\caption{The number of MRC datasets created in recent years and the F1 scores of state-of-the-art models on SQuAD 1.1 \cite{rajpurkar2016squad}: (\textbf{a}) The cumulative number of MRC datasets from the beginning of 2014 to the end of 2019. (\textbf{b}) The progress of state-of-the-art models on SQuAD 1.1 since this dataset was released. The data points are taken from the leaderboard at \href{https://rajpurkar.github.io/SQuAD-explorer/}{https://rajpurkar.github.io/SQuAD-explorer/}.}
	\label{figure: model}
\end{figure}

The benchmark datasets play a crucial role in speeding up the development of better neural models. In the past few years, we have witnessed an explosion of work that brings various MRC benchmark datasets \cite{boerma2016reading,sugawara2019assessing,zhang2019machine,baradaran2020survey, gupta2020conversational}. Figure \ref{figure: model} (a) shows the cumulative number of MRC datasets from the beginning of 2014 to the beginning of 2020. It shows that the number of MRC datasets has increased exponentially in recent years. And these novel datasets inspired a large number of new neural MRC models, such as those shown in Figure \ref{figure: model} (b), just take SQuAD 1.1 \cite{rajpurkar2016squad} for example, we can see that many neural network models were created in recent years, such as BiDAF \cite{seo2016bidirectional}, ELMo \cite{peters2018deep}, BERT \cite{devlin2018bert}, RoBERTa \cite{liu2019roberta} and XLNet \cite{yang2019xlnet}. The performance of the state-of-the-art neural network models have already exceeded human performance over the related MRC benchmark datasets.

Despite the critical importance of MRC datasets, most of the existing MRC reviews have focused on MRC algorithms for improving system performance \cite{grail2018reviewqa,Qiu2019ASO}, performance comparisons \cite{baradaran2020survey}, or general review that has limited coverage of datasets \cite{zhang2019machine}.  In addition, there is also a need for systematic categorization/classification of task types. For example, MRC tasks are usually divided into four categories: cloze style, multiple-choice, span prediction and free form \cite{chen2018neural,liu2019neural,Qiu2019ASO}. But this classification method is not precise because the same MRC task could belong to both cloze style and multiple-choice style at the same time, such as the CBT \cite{hill2016cbt} task in the Facebook bAbi project \cite{weston2016babi}. Moreover, most researchers focus on few popular MRC datasets while most other MRC datasets are not widely known and studied by the community. To address these gaps, a comprehensive survey of existing MRC benchmark datasets, evaluation metrics and tasks is strongly needed.

At present, a lot of neural MRC models have already surpassed human performance on many MRC datasets, but there is still a giant gap between existing MRC and real human comprehension \cite{jia-liang-2017-adversarial}. This shows the need of improving existing MRC datasets in terms of both question and answer challenges and related evaluation criteria. In order to build more challenging MRC datasets, we need to understand existing MRC tasks, evaluation metrics and datasets better.

\subsection{Outline}	
In section~\ref{sec:Tasks}, we focus on the MRC tasks. We first give a definition of typical MRC task. Then we compare multi-modal MRCs with textual MRCs, and discuss the differences between question answering tasks and machine reading comprehension tasks. Next, we analyze the existing classification method of MRC tasks which is widely used by the community. We argue that the existing classification method is inadequate and has potential problems. In order to solve the above problems, we propose a more adequate classification method of MRC tasks. We summarize 4 different attributes of MRC tasks. Each of these attributes can be divided into several categories. We give a detailed definition of each category with examples and explain why the new classification method is more adequate. After that, we collect totally 57 different MRC tasks and categorize them according to the new classification method. Finally, we analyze these MRC tasks and make statistical tables and charts of them.

In section~\ref{sec:Metrics}, we discuss the MRC evaluation metrics. Nine evaluation metrics of MRC tasks have been analyzed. We begin by presenting an overview of MRC evaluation metrics. Then we discuss the computing methods of each evaluation metric, including several sub-metrics such as token-level F1 and question-level F1. Next, we analyze the usage of each evaluation metric in different MRC tasks. After that, we make statistics on the usages of different evaluation metrics in the 57 MRC tasks. Finally, we analyze the relationship between the MRC task types and the evaluation metrics they used. 

In section~\ref{sec:Dataset}, we present the family of MRC datasets. We begin by analyzing the size of each MRC datasets. Here, we have counted the total number of questions in each MRC dataset along with the sizes of its training set, development set and testing set, as well as the proportion of training set. Then we discuss the generation method of datasets which can be roughly described as several categories: Crowdsourcing, Expert, and Automated. Next, we conduct an in-depth analysis of the source of corpus and the type of context of MRC datasets. After that, we try to find all the download links, leaderboards and baseline projects of MRC datasets, all of which have been published on our website. Then, we present a statistical analysis of prerequisite skills and citations of the papers in which each dataset was proposed. Next, we summarize 10 characteristics of MRC datasets. Finally, we give a detailed description of each MRC dataset.

In section~\ref{sec:Issues}, we discuss several open issues that remain unsolved in this field. Firstly, We believe that many important aspects have been overlooked which merit additional research, such as multi-modal MRC, commonsense and world knowledge, complex reasoning, robustness, interpretability, evaluation of the quality of MRC datasets. Secondly, we talk about understanding from the perspective of cognitive neuroscience. Finally, we share some of the latest research results of cognitive neuroscience and the inspiration of these results for NLP research. 

In section~\ref{sec:Conclusions}, we present a comprehensive conclusion of this survey.

Finally, we have published all the data on the website, the researchers could directly access each MRC datasets, papers, and baseline projects, or browse the leaderboards by clicking the hyperlinks. It is hoped that the research community could quickly access the comprehensive information of MRC datasets and their tasks. The address of the website is \href{https://mrc-datasets.github.io/}{https://mrc-datasets.github.io/}.

\section{Tasks}
\label{sec:Tasks}
\subsection{Definition of Typical MRC Tasks}
In our survey, machine reading comprehension is considered as a special research field, which includes some specific tasks, such as multi-modal machine reading comprehension, textual machine reading comprehension, etc. Since most of the existing machine reading comprehension tasks are in the form of question answering, the textual QA-based machine reading comprehension task is considered to be the typical machine reading comprehension task.
According to previous review papers on MRC \cite{chen2018neural,liu2019neural}, the definition of a typical MRC task is:

\begin{table}[H]
	\centering
	\begin{tabular}{p{12cm}c}
		\textbf{Definition 1.} Typical machine reading comprehension task could be formulated as a supervised learning problem. Given a collection of textual training examples $\big\{\big(p_{i}, q_{i}, a_{i}\big)\big\}_{i=1}^{n}$, where $p$ is a passage of text, and $q$ is a question regarding the text $p$. The goal of typical machine reading comprehension task is to learn a predictor $f$ which takes a passage of text $p$ and a corresponding question $q$ as inputs and gives the answer $a$ as output, which could be formulated as the following formula \cite{chen2018neural}:		
		\begin{equation}a=f(p, q)\end{equation}		
		and it is necessary that a majority of native speakers would agree that the question $q$ does regarding that text $p$, and the answer $a$ is a correct one which does not contain information irrelevant to that question.\\
	\end{tabular}
\end{table}

\subsection{Discussion on MRC Tasks}
In this section, we first compare multi-modal MRCs with textual MRCs, and then discuss the relationship between question answering tasks and machine reading comprehension tasks. 

\subsubsection{Multi-modal MRC vs. Textual MRC}
Multi-modal MRC is a new challenging task that has received increasing attention from both the NLP and the CV communities. Compared with existing MRC tasks which are mostly textual, multi-modal MRC requires a deeper understanding of the text and visual information such as images and videos. When human reads, illustrations can help to understand the text. Experiments showed that children with higher mental imagery skills outperformed children with lower mental imagery skills on story comprehension after reading the experimental narrative \cite{boerma2016reading}. These results emphasize the importance of mental imagery skills for explaining individual variability in reading development \cite{boerma2016reading}.
Therefore, if we want the machine to acquire human-level reading comprehension ability, multi-modal machine reading comprehension is a promising research direction.

\begin{figure}[H]
    \centering
	\begin{tabular} {m{6cm} m{6cm}}
		\toprule
		\textbf{Passage with illustration:}                & \multirow{2}{*}{\includegraphics[width=6cm]{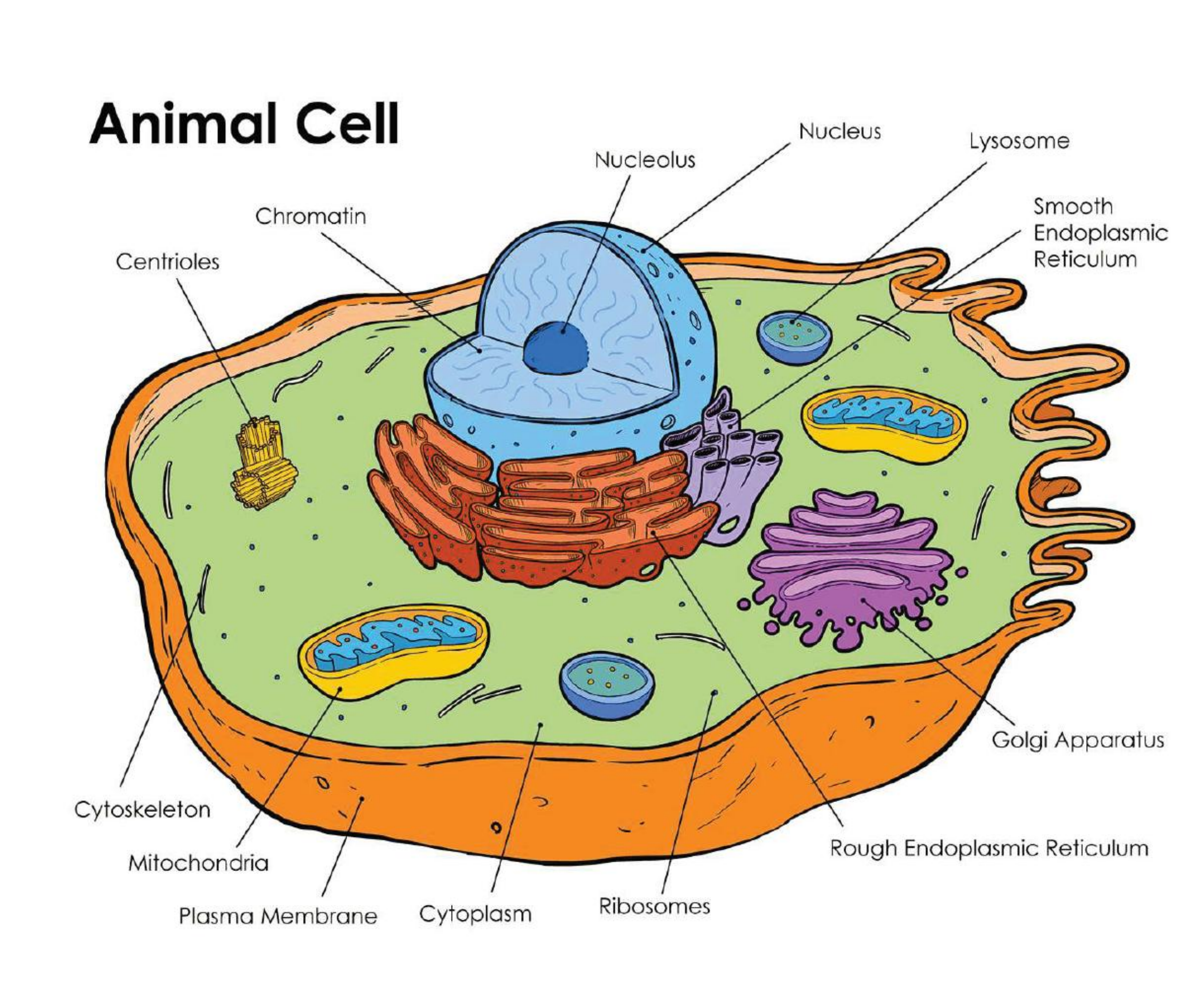}} \\
		This diagram shows the anatomy of an Animal cell. Animal Cells have an outer boundary known as the plasma membrane. The nucleus and the organelles of the cell are bound by this membrane. The cell organelles have a vast range of functions to perform like hormone and enzyme production to providing energy for the cells. They are of various sizes and have irregular shapes. Most of the cells size range between 1 and 100 micrometers and are visible only with help of microscope. &                      \\
		
        \midrule	
			\textbf{Question with illustration:}               & \multirow{8}{*}{\includegraphics[width=6cm]{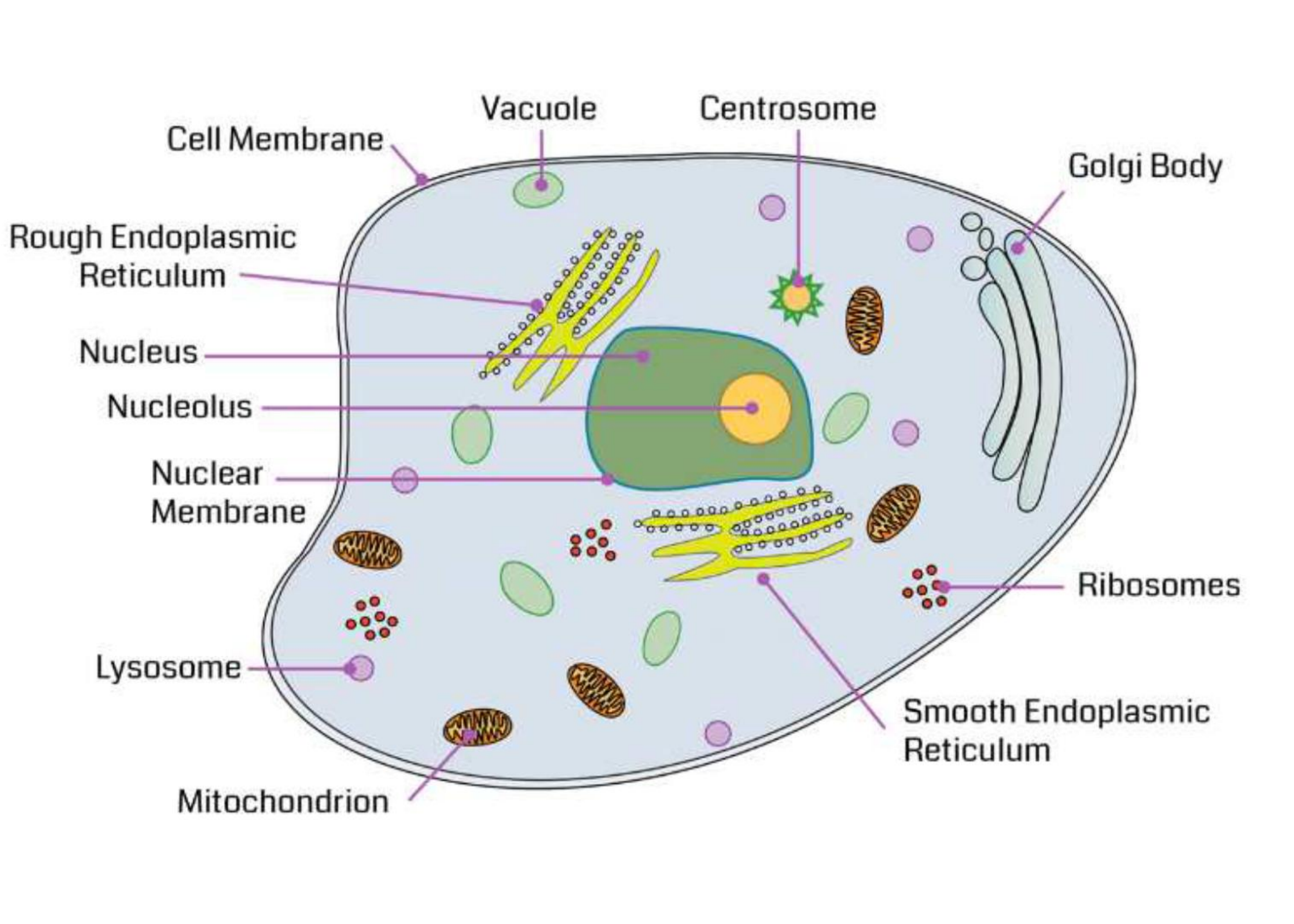}} \\
		What is the outer surrounding part of the Nucleus? &                      \\
		                                                   &                     \\
		\textbf{Choices:}                                  &                      \\
		(1) Nuclear Membrane  $\surd$                      &                      \\
		(2) Golgi Body                                     &                      \\
		(3) Cell Membrane                                  &                      \\
		(4) Nucleolus                                      &                     \\
		                                                   &                 	 \\		
        \bottomrule
		\end{tabular}
		\caption{An example of multi-modal MRC task. The illustrations and questions are taken from the TQA \cite{kembhavi2017tqa} dataset.}
		\label{figure: mmtasks }
\end{figure}

In fact, there are already many tasks and datasets in this field, such as the TQA \cite{kembhavi2017tqa}, MovieQA \cite{tapaswi2016movieqa}, COMICS \cite{iyyer2017COMICS} and RecipeQA \cite{yagcioglu-etal-2018-recipeqa}. As seen in Figure \ref{figure: mmtasks }, TQA is a multi-modal MRC dataset that aims at answering multi-modal questions given a context of text, diagrams and images.

\subsubsection{Machine Reading Comprehension vs. Question Answering}
The relationship between question answering and machine reading comprehension is very close. Some researchers consider MRC as a kind of specific QA task \cite{chen2018neural, liu2019neural}. Compared with other QA tasks such as open-domain QA, MRC is characterized by that the computer is required to answer questions according to the specified text. However, other researchers regard the machine reading comprehension as a kind of method to solve QA tasks. For example, in order to answer open-domain questions, Chen et al. \cite{chen-etal-2017-reading} first adopted document retrieval to find the relevant articles from Wikipedia, then used MRC to identify the answer spans from those articles. Similarly, Hu \cite{hu2019mrc} regarded machine reading as one of the four methods to solve QA tasks. The other three methods are rule-based method, information retrieval method and knowledge-based method.

However, although the typical machine reading comprehension task is usually in the form of textual question answering, the forms of MRC tasks are usually diverse. Lucy Vanderwende \cite{vanderwende2007answering} argued that machine reading could be defined as an automatic understanding of text. "One way in which human understanding of text has been gauged is to measure the ability to answer questions pertaining to the text. An alternative way of testing human understanding is to assess one's ability to ask sensible questions for a given text". 

In fact, there are many such benchmark datasets for evaluating such techniques. For example, ShARC \cite{saeidi2018sharc} is a conversational MRC dataset. Unlike other conversational MRC datasets, when answering questions in the ShARC, the machine needs to use background knowledge that is not in the context to get the correct answer. The first question in a ShARC conversation is usually not fully explained and does not provide enough information to answer directly. Therefore, the machine needs to take the initiative to ask the second question, and after the machine has obtained enough information, it then answers the first question. 

Another example is RecipeQA \cite{yagcioglu-etal-2018-recipeqa} which is a dataset for multi-modal comprehension of illustrated recipes. There are four sub-tasks in RecipeQA, one of which is ordering task. Ordering task tests the ability of a model in finding a correctly ordered sequence given a jumbled set of representative images of a recipe \cite{yagcioglu-etal-2018-recipeqa}. As in previous visual tasks, the context of this task consists of the titles and descriptions of a recipe. To successfully complete this task, the model needs to understand the temporal occurrence of a sequence of recipe steps and infer temporal relations between candidates, i.e. boiling the water first, putting the spaghetti next, so that the ordered sequence of images aligns with the given recipe. In addition, in the MS MARCO \cite{tri2016MARCO}, ordering tasks are also included.

\begin{figure}[H]
	\centering
	\includegraphics[width=10 cm]{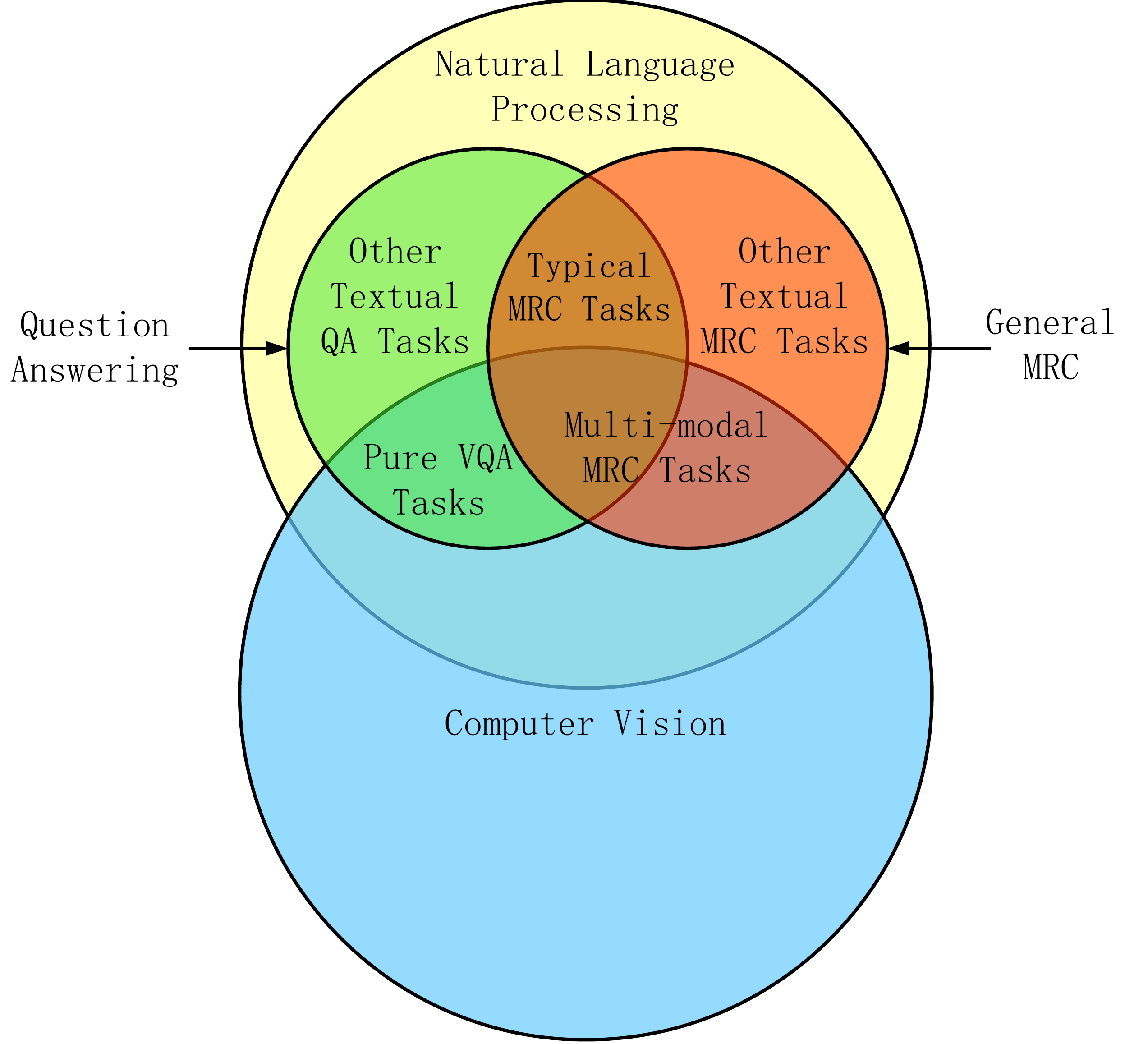}
	\caption{The relations between machine reading comprehension (MRC), question answering (QA), natural language processing (NLP) and computer version (CV). }
	\label{figure: relation }
\end{figure}

In summary, although most machine reading comprehension tasks are in the form of question answering, it does not mean that machine reading comprehension tasks belong to the question answering. In fact, as mentioned above, the forms of MRC tasks are diverse. Question answering also includes a lot of tasks that do not emphasize that the system must read a specific context to get an answer, such as rule-based question answering systems and knowledge-based question answering systems (KBQA). Figure \ref{figure: relation } illustrates the relation between machine reading comprehension (MRC) tasks and question answering (QA) tasks. As shown in Figure \ref{figure: relation }, we regard the general machine reading comprehension and the question answering as two subfields in the research field of natural language processing, both of which contain various specific tasks, such as Visual Question Answering (VQA) tasks, multi-modal machine reading comprehension tasks, etc.
Among them, some of these tasks belong to both natural language processing and computer vision research fields, such as the VQA task and the multi-mode reading comprehension task.
Lastly, most of the existing MRC tasks are textual question answering tasks, so we regard this kind of machine reading comprehension task as a typical machine reading comprehension task, and its definition is shown in Definition 1 above.

\subsubsection{Machine Reading Comprehension vs. Other NLP tasks}
There is a close and extensive relationship between machine reading comprehension and other NLP tasks. First of all, many useful methods in the field of machine reading comprehension can be introduced into other NLP tasks. For example, the stochastic answer network (SAN) \cite{liu2018stochasticMRC,liu2018stochasticSQuAD} is first applied to MRC tasks and achieved results competitive to the state of the art on many MRC tasks such as the SQuAD and the MS MARCO. At the same time, the SAN can also be used in natural language processing (NLP) benchmarks \cite{liu2018stochasticNLI}, such as Stanford Natural Language Inference (SNLI), MultiGenre Natural Language Inference (MultiNLI), SciTail, and Quora Question Pairs datasets. For another example, Yin et al.(2017) \cite{2017Document} regards the document-level multi-aspect sentiment classification task as a machine understanding task, and proposed a hierarchical iterative attention model. The experimental result of this model outperforms the classical baseline in TripAdvisor and BeerAdvocate datasets.

Secondly, some other NLP research results can also be introduced into the MRC area. Asai et al. (2018) \cite{2018Multilingual} solved the task of non-English reading comprehension through a neural network translation (NMT) model based on attention mechanism. In detail, the paragraph question pair of non-English language is translated into English using the neural machine translation model, so that the English extraction reading comprehension model can output its answer, and then use the attention weights of the neural machine translation model to align the answers in the target text. Extra knowledge can also be introduced into MRC tasks. The authors of SG-Net \cite{Zhang2020SG} used syntax information to constrain attention in the MRC task. They used the syntactic dependency of interest (SDOI) to form an SDOI-SAN and have achieved state-of-the-art results on SQuAD 2.0 challenge. Minaee et al. (2020) \cite{minaee2020deep} summarized more than 150 deep learning text classification methods and their performance on more than 40 popular datasets. Many of the methods mentioned in this article have been applied to MRC tasks. 

Thirdly, MRC can be used as a step or component in the pipeline of some complex NLP tasks. For example, machine reading comprehension can be used as a step in open domain QA \cite{chen-etal-2017-reading}. And in many dialogue tasks, machine reading comprehension can also be regarded as a part of pipeline \cite{saeidi2018sharc,reddy2019coqa,gupta2020conversational}.

\subsection{Classification of MRC Tasks}
In order to have a better understanding of MRC tasks, in this section, we analyze existing classification methods of tasks and identify potential limitations of these methods. After analyzing 57 MRC tasks and datasets, we propose a more precise classification method of MRC tasks which has 4 different attributes and each of them could be divided into several types. The statistics of the 57 MRC tasks are shown in the table in this section.   

\subsubsection{Existing Classification Methods of MRC tasks}
 In many research papers \cite{chen2018neural,liu2019neural,Qiu2019ASO}, MRC tasks are divided into four categories: cloze style, multiple-choice, span prediction, and free-form answer. Their relationship is shown in Figure \ref{figure:existingclass}:

\begin{figure}[H]
	\centering
	\includegraphics[width=11 cm]{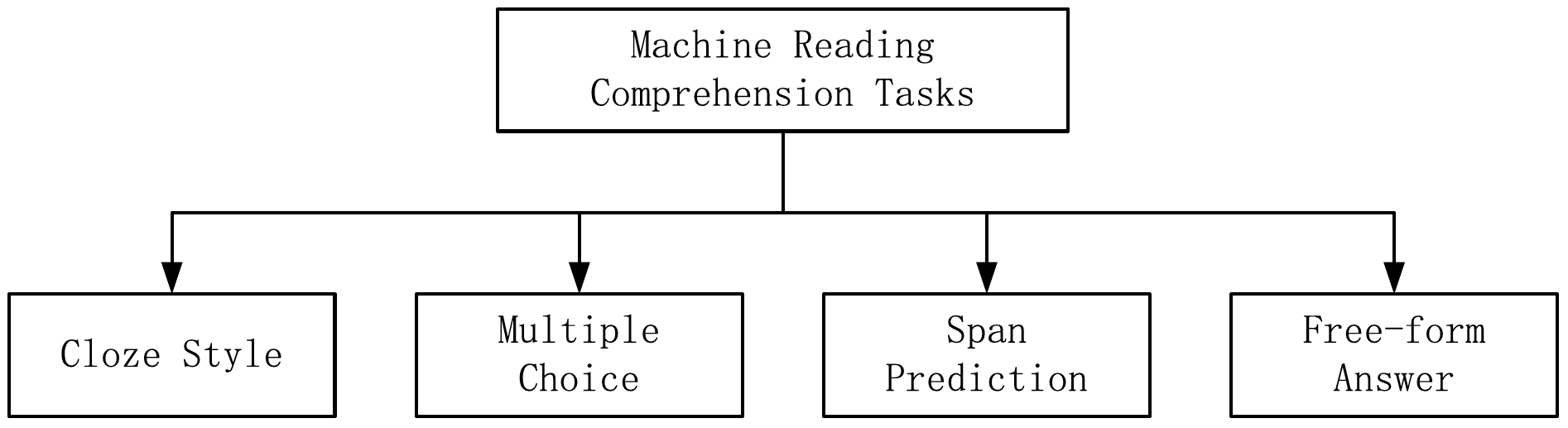}
	\caption{Existing classification method of machine reading comprehension tasks.}
	\label{figure:existingclass}
\end{figure}

\begin{itemize}[leftmargin=*,labelsep=4.9mm]
	\item	Cloze style
\end{itemize}

In a cloze style task, there are some placeholders in the question. The MRC system needs to find the most suitable words or phrases which can be filled in these placeholders according to the context content.

\begin{itemize}[leftmargin=*,labelsep=4.9mm]	
	\item	Multiple-choice
\end{itemize}

In a multiple-choice task, the MRC system needs to select a correct answer from a set of candidate answers according to the provided context.

\begin{itemize}	[leftmargin=*,labelsep=4.9mm]	
	\item	Span prediction
\end{itemize}

In a span prediction task, the answer is a span of text in the context. That is, the MRC system needs to select the correct beginning and end of the answer text from the context.

\begin{itemize}[leftmargin=*,labelsep=4.9mm]	
	\item	Free-form answer
\end{itemize}

This kind of tasks allows the answer to be any free-text forms, that is, the answer is not restricted to a single word or a span in the passage \cite{chen2018neural}.\\

\subsubsection{Limitations of Existing Classification Method}
However, the above task classification method does have certain limitations. Here are the reasons:

First, an adequate classification method should be precise at least or can classify each MRC task distinctly. But the existing classification method is a bit ambiguous or indistinct, that is, according to this classification method, a MRC task may belong to multiple task types. 
For instance, as seen in Figure \ref{figure: textual_cloze }, a sample in the "Who did What" task \cite{onishi2016whodidwhat} are both in the form of "Cloze style" and "Multiple-choice", and we can see that the answer is a span of a text in the context so that it can also be classified to "Span prediction". 

\begin{figure}[H]
	\centering
	\begin{tabular}{p{13cm}l}
		\toprule
		\textbf{Passage:}\quad Tottenham won 2-0 at Hapoel Tel Aviv in UEFA Cup action on Thursday night in a defensive display which impressed Spurs skipper Robbie Keane. ... Keane scored the first goal at the Bloomfield Stadium with \textbf{\textit{\color{red}Dimitar Berbatov}}, who insisted earlier on Thursday he was happy at the London club, heading a second. The 26-year-old Berbatov admitted the reports linking him with a move had affected his performances ... Spurs manager Juande Ramos has won the UEFA Cup in the last two seasons ... \\
		\midrule
		\textbf{Question:}\quad Tottenham manager Juande Ramos has hinted he will allow \underline{\hbox to 18mm{}} to leave if the Bulgaria striker makes it clear he is unhappy. \\
		\midrule
		\textbf{Choices: } \quad (A) Robbie Keane \quad(B)	\textbf{\color{red}Dimitar Berbatov }$\surd$\\
		\bottomrule
	\end{tabular}
 	\caption{An example of MRC task. The question-answer pair and passage are taken from the "Who did What" \cite{onishi2016whodidwhat}. } 	
 	\label{figure: textual_cloze }		
\end{figure}	

Secondly, with the rapid development of MRC, a large number of novel MRC tasks have emerged in recent years. One example is multi-modal MRC, such as MovieQA \cite{tapaswi2016movieqa}, COMICS \cite{iyyer2017COMICS}, TQA \cite{kembhavi2017tqa} and RecipeQA \cite{yagcioglu-etal-2018-recipeqa}. Compared with the traditional MRC task which only requires understanding a text, the multi-modal MRC task requires the model to understand the semantics behind the text and visual images at the same time. A fundamental characteristic of human language understanding is multimodality. Our observation and experience of the world bring us a lot of common sense and world knowledge, and the multi-modal information is extremely important for us. In essence, real world information is multi-modal and widely exists in texts, voices, and images. But these multi-modal tasks are ignored by the existing classification method.

\begin{figure}[H]
	\centering
	\includegraphics[width=12 cm]{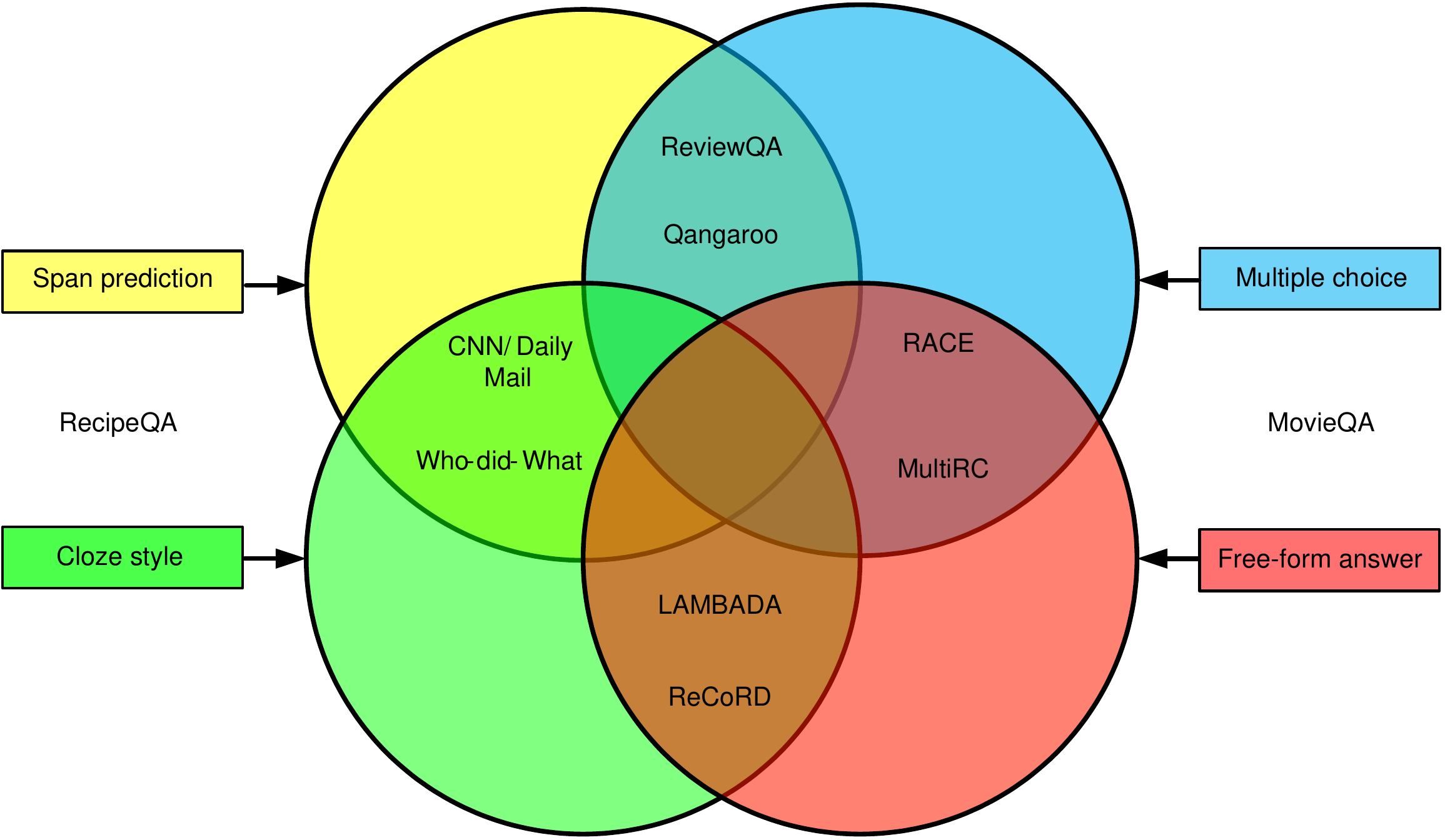}
	\caption{The indistinct classification caused by existing classification method.}
	\label{figure:fuzzy}
\end{figure}

In addition, as seen in Figure \ref{figure:fuzzy}, we list several tasks that belong to the fuzzy classification mentioned above, such as ReviewQA, Qangaroo, Who-did-What, MultiRC, LAMBADA, ReCoRD. Due to the limited space, we only list a few of them in the figure. According to our statistics, among the 57 MRC tasks we collected, 29 tasks fall into this situation. 

\subsubsection{A new Classification Method}
\begin{figure}[H]
	\centering
	\includegraphics[width=15 cm]{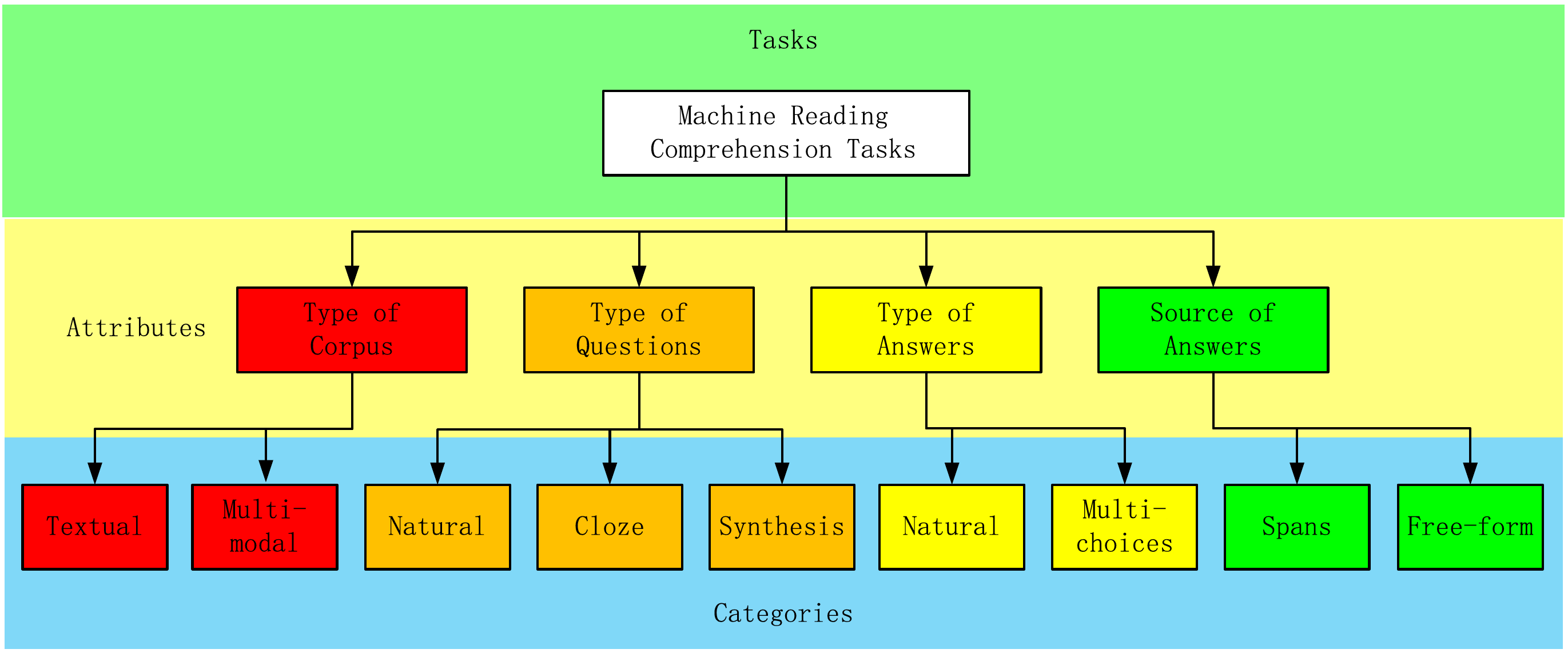}
	\caption{A new classification method of machine reading comprehension tasks.}
	\label{figure: new_method }
\end{figure}

In this section, we propose a new classification method of MRC tasks. As shown in Figure \ref{figure: new_method }, we summarize four different attributes of MRC tasks, including the type of corpus, the type of questions, the type of answers, and the source of answers. Each of these attributes can be divided into several different categories. These categories are:
(1) Type of corpus: textual, multi-modal.
(2) Type of questions: natural form, cloze style, synthetic form.
(3) Type of answers: natural form, multiple-choices.
(4) Source of answers: spans, free-form.

\begin{figure}[H]
	\centering
	\includegraphics[width=13cm]{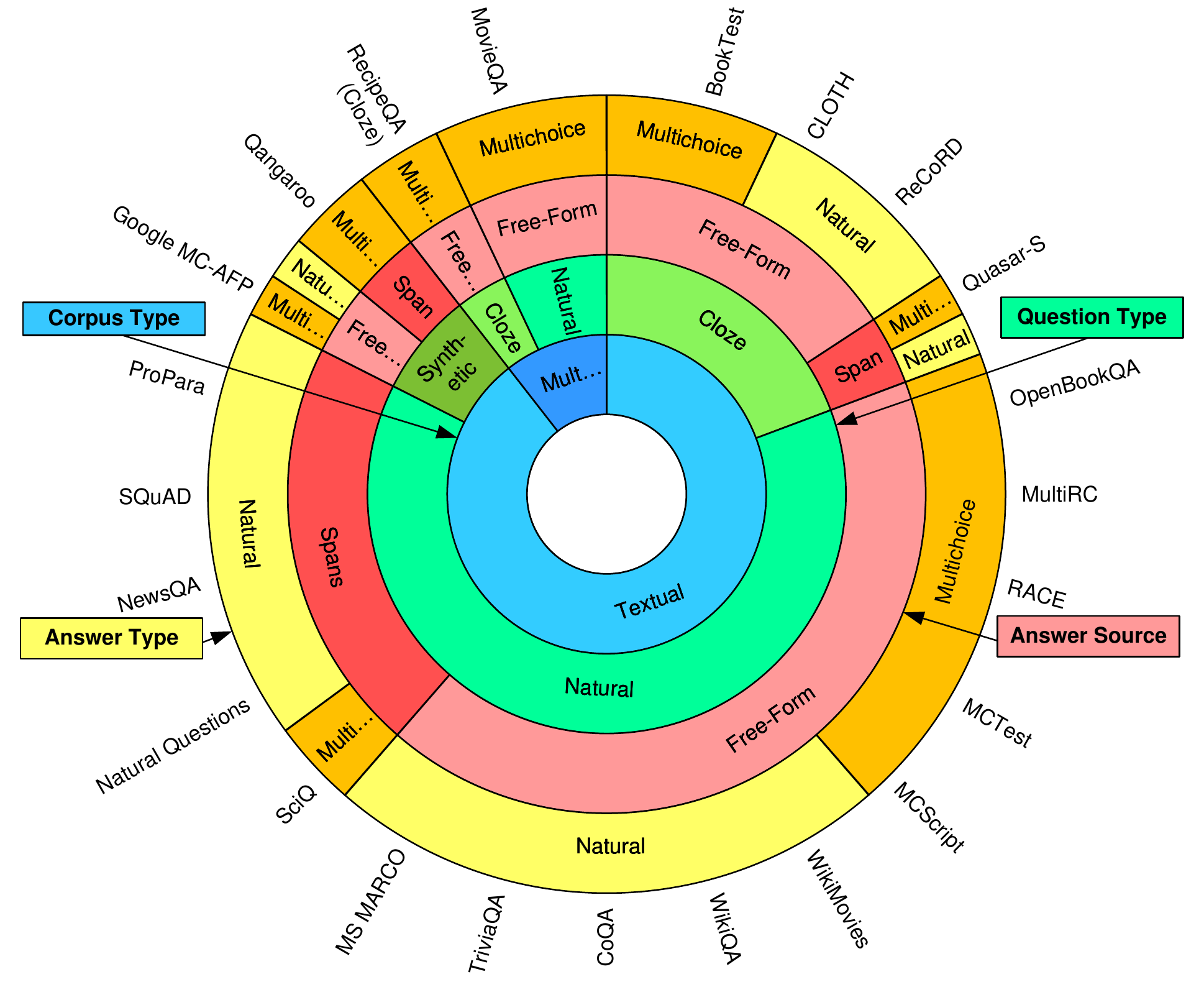}
	\caption{A sunburst chart on the proportion of different types of machine reading comprehension tasks.} 
	\label{figure: sunchart_tasks }
\end{figure}

In order to explain the new classification method more clearly, we make a sunrise statistical chart for the MRC task classification, as seen in Figure \ref{figure: sunchart_tasks }. We collect 57 different specific MRC tasks. Finally, according to the new classification method, the sunrise chart is divided into four layers of rings, representing the four attributes of tasks. The most central blue layer represents the 'Type of Corpus'. Among them, light blue indicates that the type of the task's corpus belongs to 'Textual', and dark blue means that the type of the task's corpus belongs to 'Multi-modal'. The magnitude of different color blocks is set according to the proportion of 57 MRC tasks we collected. Among them, the 'Textual' tasks still account for the vast majority of tasks (89.47\%). Currently, the proportion of MRC tasks is still very small, about 10.53\%. Therefore, as can be seen, the range of light blue color blocks is large, while the range of dark blue color blocks is small. The second green layer represents the 'Type of Question', the third red and pink layer represents the 'Type of Answer', and the outermost yellow layer represents the 'Source of Answer'.

Take the BookTest for an example, as seen in the top of Figure \ref{figure: sunchart_tasks }, the BookTest is a 'Textual' MRC task, its question is in the 'Cloze' form, and the answer comes from 'Spans' in the context, its answer form is 'Multiple-choice'. Another example is the cloze subtask in the RecipeQA dataset, which is also in the top of Figure\ref{figure: sunchart_tasks }. In this task, the answer form is 'Multiple-choice', and question types is 'Cloze' form. Moreover, the context corpus of RecipeQA contains images, so it is a 'Multi-modal' task. The answer types include textual 'Multiple-choice' and image 'Multiple-choice'. 
Therefore, the type of each MRC task is determined according to four different attributes, which eliminates the fuzzy situation that the same task belongs to multiple types in the traditional classification method. 

However, it must be pointed out that although the new classification method fits precisely to the existing datasets, it may suffer from the lack of future generalization. We believe that with the continuous development of MRC field, new MRC tasks will certainly appear, and the classification methods of MRC tasks will also keep pace with them.

\subsection{Definition of each category in the new classification method} 
As mentioned above, we propose a new classification method of MRC tasks. As shown in Figure \ref{figure: new_method } above, we summarize four different attributes of MRC tasks, including the type of corpus, the type of questions, the type of answers, and the source of answers. Each of these attributes can be divided into several different categories. In this subsection, we will give detailed definitions of each category with examples.  

Here are some assumptions or notations we need before the formal definitions:

\begin{table}[H]
	\centering
	\label{table: Q}
	\begin{tabular}{p{13cm}l}
	    \textbf{Assumption 1.} Suppose $V$ is a pure textual vocabulary, and $M$ is a multi-modal dataset which consists of images or other non-text imformation.	\\
	    
	   	\textbf{Assumption 2.} Suppose in a MRC corpus, $C_{i}$ is the i-th context, $Q_{i}$ is the i-th question, and $A_{i}$ is the answer to question $Q_{i}$ according to context $C_{i}$. Let the context $C_{i}=\big\{c_{0}, c_{1}, \ldots , c_{l c i}\big\}$, and the question $Q_{i}=\big\{q_{0}, q_{1},\ldots , q_{l q i}\big\}$, and $A_{i}=\big\{a_{0}, a_{1}, \ldots, a_{\text {lai }}\big\}$, where $l_{ci}$, $l_{qi}$ and $l_{ai}$ denote the length of the i-th context $C_{i}$, question $Q_{i}$, and answer $A_{i}$ respectively. While $c_{i}$, $q_{i}$ and $a_{i}$ is usually a word or a image, i.e. $c_{k} \in V \cup M$, $q_{k} \in V \cup M$ and $a_{k} \in V \cup M$.
	\end{tabular}
\end{table}

\subsubsection{Type of Corpus}
According to whether or not the corpus contains information other than text, such as pictures, the MRC tasks can be divided into two categories: multi-modal (the combination of graphics and text) and textual.

\begin{itemize}[leftmargin=*,labelsep=4.9mm] 	
	\item	Multi-modal
\end{itemize} 

In multi-modal MRC Corpus, multi-modal information includes context, questions, or answers. It can be defined as:

\begin{table}[H]
	\centering
	\begin{tabular}{p{13cm}l}
		\textbf{Definition 2.} In a MRC task with multi-modal corpus, the corpus $P$ can be formalized as a collection of training examples, that is, $P=\big\{C_{i}, Q_{i}, A_{i}\big\}_{i=1}^{n}$, where $C_{i}$ is the context, $Q_{i}$ is a question, and $A_{i}$ is the answer to question $Q_{i}$ according to context $C_{i}$. In the multi-modal corpus $P$, the entities in the corpus consists of text and images at the same time, therefore, $\mathrm{P} \cap \mathrm{V} \neq \emptyset$ and $\mathrm{P} \cap \mathrm{M} \neq \emptyset$. 	
	\end{tabular}
\end{table}

An example of the multi-modal corpus can be seen in Figure \ref{figure: mmtasks } above. There is a certain similarity between multi-modal MRC tasks and Visual Question Answering (VQA) tasks. But multi-modal MRC tasks focus more on natural language understanding, and their context contains more text that needs to be read, and the VQA task usually does not have much context and gives the image directly.

\begin{itemize}[leftmargin=*,labelsep=4.9mm]
 	\item	Textual
\end{itemize}  	
 	
Most MRC tasks belong to this category. Their context, questions and answers are all plain texta. It can be defined as: 

\begin{table}[H]
	\centering
	\begin{tabular}{p{13cm}l}
	    \textbf{Definition 3.} In a MRC task with textual corpus, the corpus $P$ can be formalized as a collection of training examples, that is, $P=\big\{C_{i}, Q_{i}, A_{i}\big\}_{i=1}^{n}$, where $C_{i}$ is the context, $Q_{i}$ is a question, and $A_{i}$ is the answer to question $Q_{i}$ according to context $C_{i}$. In the textual corpus $P$, all the entities in the context, questions and answers are in pure text, therefore, $\mathrm{P} \cap \mathrm{V} \neq \emptyset$ and $\mathrm{P} \cap \mathrm{M} = \emptyset$. 
		\\
	\end{tabular}
\end{table}

Example of textual corpus can be seen in Figure \ref{figure: textual_MRC_task} below:

     
\begin{figure}[H]
	\centering
	\begin{tabular}{p{13cm}l}
		\toprule
		\textbf{Passage:}\quad In meteorology, precipitation is any product of the condensation of atmospheric water vapor that falls under \textit{\textbf{gravity}}. The main forms of precipitation include drizzle, rain, sleet, snow, graupel and hail... Precipitation forms as smaller droplets coalesce via collision with other rain drops or ice crystals within a cloud. Short, in- tense periods of rain in scattered locations are called "showers".\\
		\midrule
		\textbf{Question:}\quad What causes precipitation to fall?\\
		\midrule
		\textbf{Answer:} \quad \textbf{gravity}\\
		\bottomrule
	\end{tabular}
	\caption{An example of textual MRC task.}
	\label{figure: textual_MRC_task}	
\end{figure}
 
\subsubsection{Type of Questions}
According to the type of question, a MRC task can be classified into three categories: cloze style, natural form, and synthetic form:

\begin{itemize}[leftmargin=*,labelsep=4.9mm]	
\item	Cloze style
\end{itemize}

The cloze question is usually a sentence with a placeholder. Its sentence pattern may be a declarative sentence, an imperative sentence, etc., and is not necessarily an interrogative sentence. In addition, the sentence may also contain image information. The system is required to find a correct word, phrase or image that is suitable to be filled in the placeholder so that the sentence is complete. The cloze question can be defined as:

\begin{table}[H]
	\centering
	\begin{tabular}{p{13cm}l}
		\textbf{Definition 4.} Given the context $C=\big\{c_{0}, c_{1}, \ldots c_{j}, \ldots c_{j+n} \ldots,c_{l c}\big\}$ $\big(0 \leq j \leq l c, 0 \leq n \leq l c-1, c_{j} \in V \cup M \big)$, where $l_{c}$ denotes the length of this context $C$. $A=\big\{c_{j}, \ldots c_{j+n}\big\}$ is a short span in context $C$. After replaced $A$ with a placeholder $X$, a cloze style question $Q$ for context $C$ is formed, it can be formulated as $Q=\big\{c_{0}, c_{1}, \ldots  X \ldots, c_{l c}\big\}$, in which the $X$ is a placeholder. The answer to question $Q$ is the $A=\big\{c_{j}, \ldots c_{j+n}\big\}$. 
	\end{tabular}
\end{table}

According to the type of corpus, cloze questions also can be divided into textual and multi-modal. \\
A textual cloze question is usually a sentence with a placeholder. The MRC system is required to find a correct word or phrase that is suitable to be filled in the placeholder so that the sentence is complete. An example of textual cloze question has been shown in Figure \ref{figure: textual_cloze }.

A multi-modal cloze question is a natural sentence with visual information such as images, but some parts of these images are missing, and the MRC system is required to fill in the missing images. For example, a sample of visual cloze question in the RecipeQA \cite{yagcioglu-etal-2018-recipeqa} dataset is shown in Figure \ref{figure: mm_cloze }:
	
\begin{figure}[H]
	\centering
	\begin{tabular}{p{13cm}l}
		\toprule
		\textbf{Passage}\\
	   Last-Minute Lasagna:\\
	   1. Heat oven to 375 degrees F. Spoon a thin layer of sauce over the bottom of a 9-by-13-inch baking dish.\\
	   2. Cover with a single layer of ravioli.\\
	   3. Top with half the spinach half the mozzarella and a third of the remaining sauce.\\
	   4. Repeat with another layer of ravioli and the remaining spinach mozzarella and half the remaining sauce.\\
	   5. Top with another layer of ravioli and the remaining sauce not all the ravioli may be needed. Sprinkle with the Parmesan.\\
	   6. Cover with foil and bake for 30 minutes. Uncover and bake until bubbly, 5 to 10 minutes.\\
	   7. Let cool 5 minutes before spooning onto individual plates.\\
		\midrule
		\textbf{Question:}\quad Choose the best image for the missing blank to correctly complete the recipe.\\
		\includegraphics[width=13 cm]{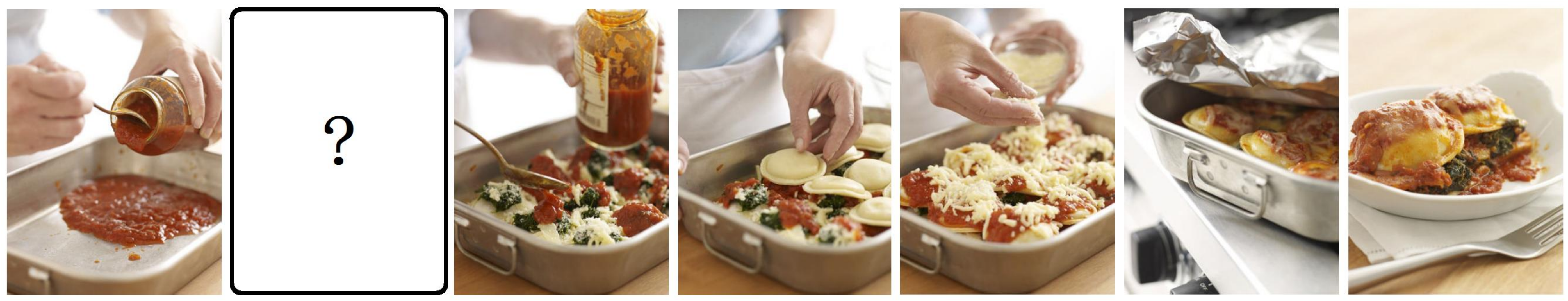}\\
		\midrule
		\textbf{Choices:}\\
		\includegraphics[width=13 cm]{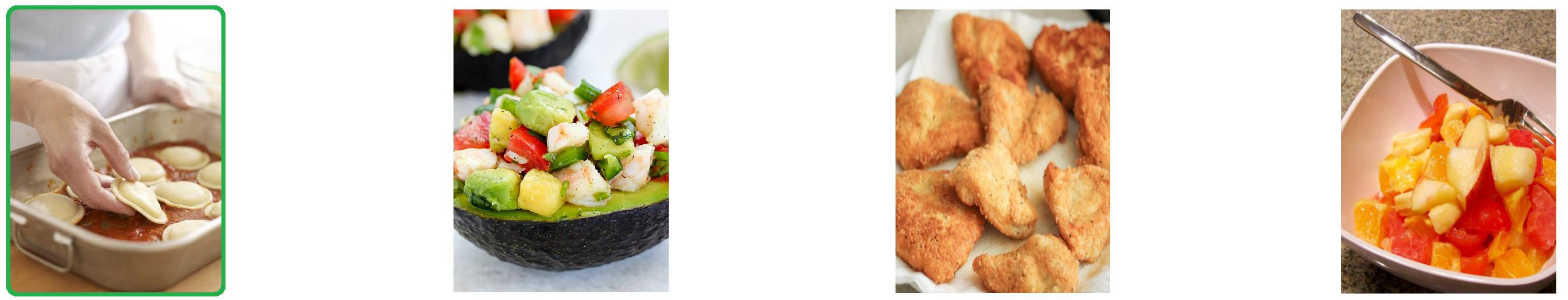}\\
		\quad\quad\  \textbf{(A)} 
		$\surd$\quad\quad\ \quad\qquad\qquad\qquad (B)  \quad\quad\ \qquad\qquad\qquad\qquad (C) \quad\quad\ \qquad\qquad\qquad\qquad (D)\\
		\bottomrule
	\end{tabular}
	\caption{An example of multi-modal cloze style question. The images and questions are taken from the RecipeQA \cite{yagcioglu-etal-2018-recipeqa} dataset.}
    \label{figure: mm_cloze }
\end{figure}

\begin{itemize}[leftmargin=*,labelsep=4.9mm]
	\item	Natural form
\end{itemize}

A question in natural form is a natural question that conforms to the grammar of natural language. Different from the cloze question, which contains placeholder, a natural form question is a complete sentence and a question that conforms to the grammatical rules. It could be defined as:

\begin{table}[H]
	\centering
	\begin{tabular}{p{13cm}l}
		\textbf{Definition 5.} In a MRC task, given a 'Natural' question $Q$, it could be formulated as  $Q_{i}=\big\{q_{0}, q_{1}, \ldots q_{i} \ldots,  q_{l q}\big\}$, where $q_{i} \in V \cup M(0 \leq i \leq l q)$. $Q$ denotes a complete sentence (may also contain images) that conforms to the natural language grammar and $l_{q}$ denotes the length of the question $Q$.
		\\
	\end{tabular}
\end{table}

In most cases, a 'Natural' question $Q$ is an interrogative sentence that asks a direct question and is punctuated at the end with a question mark. However, in some cases, $Q$ may not be an interrogative sentence but an imperative sentence, for example, "please find the correct statement from the following options." 
 
In addition, according to the type of corpus, natural form questions can be divided into textual and multi-modal. Textual natural question is usually a natural question or imperative sentence. With some graphics or video, the multi-modal natural question is also a natural question or imperative. Example of textual natural question is shown in Figure \ref{figure: textual_natural } below, and example of multi-modal natural question has been shown in Figure \ref{figure: mmtasks }.

\begin{figure}[H]
	\centering
	\begin{tabular}{p{13cm}l}
		\toprule
		\textbf{Passage:}\quad In meteorology, precipitation is any product of the condensation of atmospheric water vapor that falls under \textit{\textbf{gravity}}. The main forms of precipitation include drizzle, rain, sleet, snow, graupel and hail... Precipitation forms as smaller droplets coalesce via collision with other rain drops or ice crystals within a cloud. Short, in- tense periods of rain in scattered locations are called "showers".\\
		\midrule
		\textbf{Question:}\quad What causes precipitation to fall?\\
		\midrule
		\textbf{Answer:} \quad \textbf{gravity}\\
		\bottomrule
	\end{tabular}
	\caption{An example of textual natural question.}
	\label{figure: textual_natural }	
\end{figure}

\begin{itemize}[leftmargin=*,labelsep=4.9mm]
	\item	Synthetic style 
\end{itemize}
	
The synthetic form of the question is just a list of words and do not necessarily conform to normal grammatical rules. Common datasets with synthetic form questions are Qangaroo, WikiReading, and so on. Take Qangaroo as an example, in the Qangaroo dataset, the question is replaced by a collection of attribute words. The 'question' here is not a complete sentence that fully conforms to the natural language grammar, but a combination of words. The synthetic form of the question can be defined as: 

\begin{table}[H]
	\centering
	\begin{tabular}{p{13cm}l}
		\textbf{Definition 6.} In a MRC task, given a 'Synthetic style' question $Q$, it could be formulated as  $Q_{i}=\big\{q_{0}, q_{1}, \ldots q_{j} \ldots, q_{l q}\big\}$ , where $q_{i} \in V \cup M(0 \leq i \leq l q)$. $Q$ denotes a series of words (may also contain images) that do not conforms to the natural language grammar and $l_{q}$ denotes the length of the $Q$.
		\\
	\end{tabular}
\end{table}

The example of synthetic style question is in shown in the following:
	
\begin{figure}[H]
	\centering
	\begin{tabular}{p{13cm}l}
		\toprule
		\textbf{Passage:}
		The hanging Gardens, in [Mumbai], also known as Pherozeshah Mehta Gardens, are terraced gardens … They provide sunset views over the [Arabian Sea] …\\
		Mumbai (also known as Bombay, the ofﬁcial name until 1995) is the capital city of the Indian state of Maharashtra. It is the most populous city in \textbf{\textit{India}} …\\
		The Arabian Sea is a region of the northern Indian Ocean bounded on the north by Pakistan and Iran, on the west by northeastern Somalia and the Arabian Peninsula, and on the east by India …\\
		\midrule
		\textbf{Synthetic Question:}\quad (Hanging gardens of Mumbai, country, ?) \\
		\midrule
		\textbf{Choices: } \quad (A)Iran, \quad (B)\textbf{India$\surd$}, \quad (C)Pakistan, \quad (D) Somalia\\
		\bottomrule
	\end{tabular}
	\caption{An example of synthetic style question. The passage and question are taken from the Qangaroo \cite{welbl2018qangaroo} dataset.} 
    \label{figure: synthetic }
\end{figure}

\subsubsection{Type of Answers}
According to the type of answers, MRC tasks can be divided into two categories: multiple-choice forms, natural forms.

\begin{itemize}[leftmargin=*,labelsep=4.9mm]	
	\item	 Multiple-choice answer
\end{itemize}	

In a MRC task, when the type of answers is 'Multi-choice', there is a series of candidate answers for each question. and it can be defined as:

\begin{table}[H]
	\centering
	\begin{tabular}{p{13cm}l}
	    \textbf{Definition 7.} Given the candidate answers $A=\big\{A_{1}, \ldots A_{i} \ldots, A_{n}\big\}$, where $n$ denotes the number of candidate answers for each question, and $A_{i}(0 \leq i \leq n)$ denotes an optional answer. The goal of the task is to find the right answer $A_{j}(0 \leq j \leq n)$ from $A$, and one or more answer options in $A$ is correct.
		\\
	\end{tabular}
\end{table}

Examples of textual multiple-choices form of answers have been shown in Figure \ref{figure: textual_cloze } and Figure \ref{figure: synthetic }, and multi-modal example has been shown in Figure \ref{figure: mm_cloze } above.

\begin{itemize}[leftmargin=*,labelsep=4.9mm]
	\item	Natural form of answers
\end{itemize}

The answer is a natural word, phrase, sentence or image but it doesn't have to be in the form the multiple options. It could be defined as follows:

\begin{table}[H]
	\centering
	\begin{tabular}{p{13cm}l}
		\textbf{Definition 8.} In a MRC task, when the type of answers is 'Natural', it means the answer $A$ can be a word, a phrase or a natural sentence, or even images. The answer $A$ could be formulated as: $A=\big\{a_{1}, \ldots a_{k} \ldots, a_{l}\big\}$ , where $l$ denotes the length of answer $A$. $a_{k} \in V \cup M(0 \leq k \leq l)$.
	\end{tabular}
\end{table}

The example of natural textual answers has been shown in Table \ref{figure: textual_natural } above, and the example of natural multi-modal answer has not been found by us, i.e., all the multi-modal MRC datasets we collected in this survey contain only multiple-choice answers.

\subsubsection{Source of Answers}
According to different sources of answers, we divide the MRC tasks into two categories: span and free-form.

\begin{itemize}[leftmargin=*,labelsep=4.9mm]
	\item	Span answer
\end{itemize}
	
In a MRC task, when the source of answer is 'Spans', it means that the answers come from context and are spans of context, and it can be defined as:

\begin{table}[H]
	\centering
	\begin{tabular}{p{13cm}l}
		\textbf{Definition 9.} Given the context $C=\big\{c_{0}, \ldots c_{k} \ldots, c_{l}\big\}$, where $l$ denotes the length of the context. $c_{k} \in V \cup M(0 \leq k \leq l)$. The 'Span' answer $A$ could be formulated as $A=\big\{c_{m}, \ldots, c_{n}\big\} (0\leq m \leq n \leq l)$.  
	\end{tabular}
\end{table}

The example of textual span answer is shown in Figure \ref{figure: textual_cloze } above. It should be noted that, in this paper, we do not provide example for multi-modal span answers, because such tasks already exist in the field of computer vision, such as semantic segmentation, object detection, or instance segmentation.

\begin{itemize}[leftmargin=*,labelsep=4.9mm]	
	\item	Free-form answer
\end{itemize}

A free-form answer may be any phrase, word, or even image (not necessarily from the context). In a MRC task, when the source of answer is 'Free-form', it means that the answers can be any free-text or images, and there is no limit to where the answer comes from. It could be defined as follows:

\begin{table}[H]
	\centering
	\begin{tabular}{p{13cm}l}
		\textbf{Definition 10.} Given the context $C$, the 'Free-form' answer $A$ may or may not come from context $C$, that is, either $A \subseteq C$ or not. The 'Free-form' answer $A$ could be formulated as $A=\big\{w_{0}, w_{1}, \ldots, w_{l-1}, w_{l}\big\}$ where $l$ denotes the length of the context. $w_{k} \in V \cup M(0 \leq k \leq l)$.
		\\
	\end{tabular}
\end{table}

Example of multi-modal free-form answer are shown in Figure \ref{figure: mm_cloze } and example of textual free-form answer are shown in Figure \ref{figure:free_form_answer_question} below:
	
\begin{figure}[H]
	\centering
	\begin{tabular}{p{13cm}l}
		\toprule
		\textbf{Passage:}
		 That year, his Untitled (1981), a painting of a haloed, black-headed man with a bright red skeletal body, depicted amid the artist’s signature scrawls, was sold by Robert Lehrman for \$16.3 million, well above its \$12 million high estimate.\\
		\midrule
		\textbf{Question:}\quad How many more dollars was the Untitled (1981) painting sold for than the 12 million dollar estimation? \\
		\midrule
		\textbf{Answer: } \quad 4300000\\
		\bottomrule
	\end{tabular}
	\caption{An example of textual free-form answer. The question-answer pair and passage are taken from the DROP \cite{Dua2019DROP} dataset.}
	\label{figure:free_form_answer_question}
\end{figure}

\subsection{Statistics of MRC Tasks}
\begin{figure}[H]
	\centering
	\includegraphics[width=15 cm]{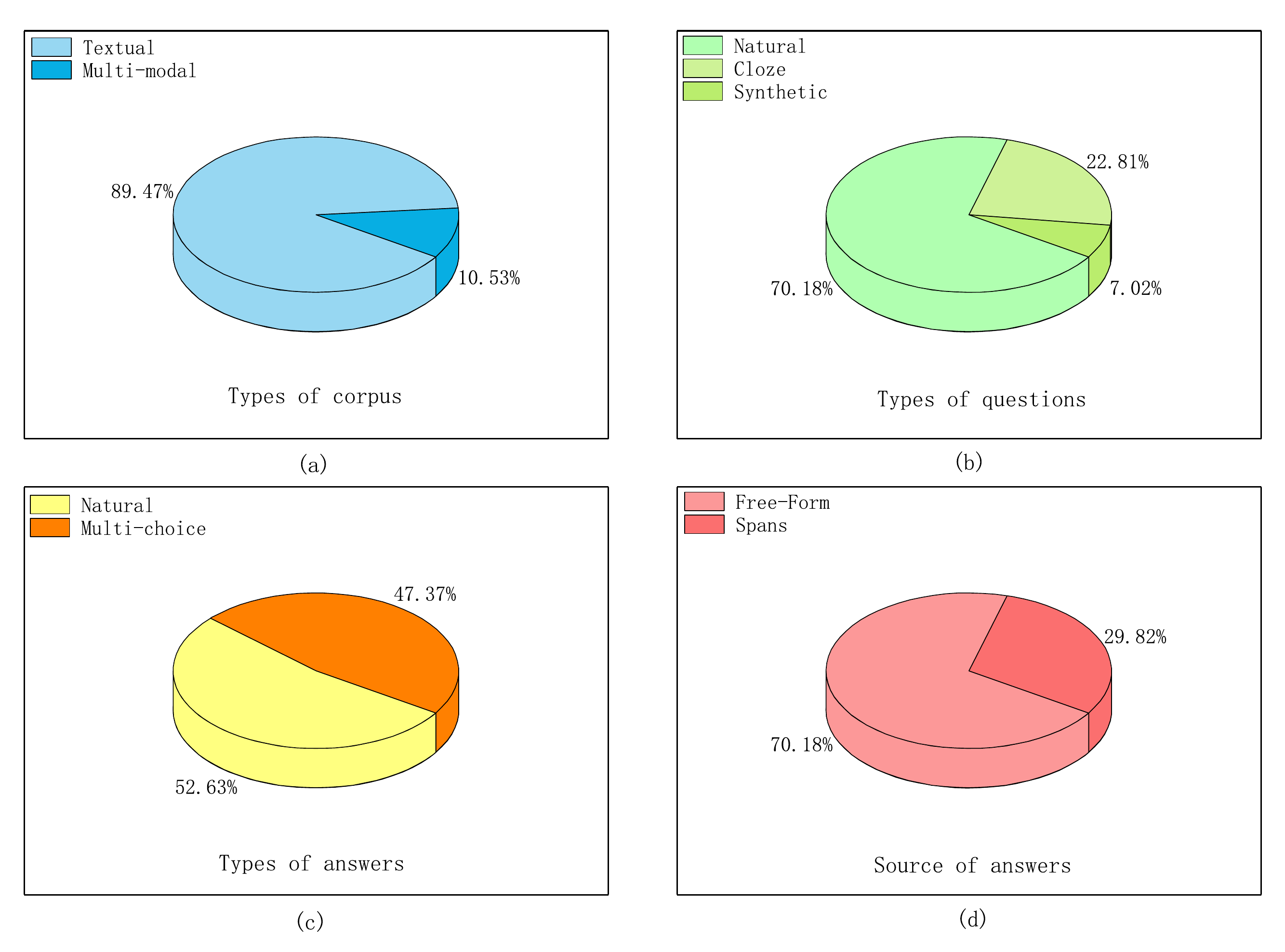}
	\caption{A pie chart on the proportion of different types of machine reading comprehension tasks: (a) Type of corpus. (b) Type of questions. (c) Type of answers. (d) Source of answers.
	} 
	\label{figure: piechart_tasks }   
\end{figure}

In this section, we collected 57 different MRC tasks and made a statistical chart of MRC task classification according to four attributes, as shown in Figure \ref{figure: piechart_tasks }. We can see that for the type of corpus, the textual task still accounts for a large proportion which is 89.47\%. At present, the proportion of multi-modal reading comprehension tasks is still small, about 10.53\%, which shows that the field of multi-modal reading comprehension still has many challenge problems for future research. In terms of question types, the most common type is the natural form of questions, followed by cloze type and synthetic type. In terms of answer types, the proportion of natural type and multiple-choice type are 52.63\% and 47.37\% respectively. In terms of answer source, 29.82\% of the answers are of spans type, and 70.18\% of the answers are of free-form.

As shown in Table \ref{table: Statistics_of_Tasks }. The tasks in the table are ordered by the year the dataset was published. It should be noted that note that the names of many specific MRC tasks are often the same as the names of the datasets they may utilize. And the name of a certain category of MRC task and the name of a specific MRC task are two different concepts. For example, the RecipeQA \cite{yagcioglu-etal-2018-recipeqa} dataset contains two different tasks which are RecipeQA-Coherence and RecipeQA-Cloze.

\begin{center}
    \renewcommand\arraystretch{1.2} 
	\topcaption{ Different machine reading comprehension tasks.}
	\label{table: Statistics_of_Tasks }
	\tablehead{
		      \toprule{\textbf{Year}} & {\textbf{MRC Tasks}}& {\textbf{Corpus Type}} & {\textbf{Question Type}} & {\textbf{Answer Source}} & {\textbf{Answer Type}}\\
	         }
	\tabletail{	
		\hline
		\multicolumn{6}{r}{\small\sl continued on next page}\\
		\hline 
		}
	
	\tablefirsthead{\toprule{\textbf{Year}} & {\textbf{MRC Tasks}}& {\textbf{Corpus Type}} & {\textbf{Question Type}} & {\textbf{Answer Source}} & {\textbf{Answer Type}}\\ \midrule}
	
	\tablelasttail{\bottomrule}
	
	\begin{supertabular}{p{1cm}l p{2cm}l p{1.5cm}l p{1.5cm}l p{1.5cm}l p{1.5cm}l}		
{2013}          & {MCTest \cite{richardson2013mctest}}                               & {Textual}              & {Natural}                & {Free-Form}              & {Multi-choice}    \\
\hline  {2015}          & {CNN/Daily Mail \cite{hermann2015cnn}}                       & {Textual}              & {Cloze}                  & {Spans}                  & {Natural}              \\
\hline  {2015}          & {CuratedTREC \cite{baudivs2015CuratedTREC}}                          & {Textual}              & {Natural}                & {Free-Form}              & {Natural}              \\
\hline  {2015}          & {WikiQA \cite{yang2015wikiqa}}                               & {Textual}              & {Natural}                & {Free-Form}              & {Natural}              \\
\hline  {2016}          & {WikiMovies \cite{miller2016WikiMovies}}                           & {Textual}              & {Natural}                & {Free-Form}              & {Natural}              \\
\hline  {2016}          & {SQuAD 1.1 \cite{rajpurkar2016squad}}                             & {Textual}              & {Natural}                & {Spans}                  & {Natural}              \\
\hline  {2016}          & {Who-did-What \cite{onishi2016whodidwhat}}                         & {Textual}              & {Cloze}                  & {Spans}                  & {Natural}              \\
\hline  {2016}          & {MS MARCO \cite{tri2016MARCO}}                             & {Textual}              & {Natural}                & {Free-Form}              & {Natural}              \\
\hline  {2016}          & {NewsQA \cite{trischler-etal-2017-newsqa}}                               & {Textual}              & {Natural}                & {Spans}                  & {Natural}              \\
\hline  {2016}          & {LAMBADA \cite{boleda2016lambada}}                              & {Textual}              & {Cloze}                  & {Free-Form}              & {Natural}              \\
\hline {2016}          & {WikiReading \cite{hewlett-etal-2016-wikireading}}                          & {Textual}              & {Synthetic}              & {Free-Form}              & {Natural}              \\
\hline {2016}          & {Facebook CBT \cite{hill2016cbt}}                         & {Textual}              & {Cloze}                  & {Free-Form}              & {Multi-choice}         \\
\hline {2016}          & {BookTest \cite{bajgar2016embracing}}                             & {Textual}              & {Cloze}                  & {Free-Form}              & {Multi-choice}         \\
\hline {2016}          & {Google MC-AFP \cite{soricut2016building}}                        & {Textual}              & {Synthetic}              & {Free-Form}              & {Multi-choice}         \\
\hline {2016}          & {MovieQA \cite{tapaswi2016movieqa}}                              & {Multi-modal}          & {Natural}                & {Free-Form}              & {Multi-choice}         \\
\hline {2017}          & {TriviaQA-Web \cite{joshi-etal-2017-triviaqa}}                        & {Textual}              & {Natural}                & {Free-Form}              & {Natural}              \\
\hline {2017}          & {TriviaQA-Wiki \cite{joshi-etal-2017-triviaqa}}                      & {Textual}              & {Natural}                & {Free-Form}              & {Natural}              \\
\hline {2017}          & {RACE \cite{lai-etal-2017-race}}                                 & {Textual}              & {Natural}                & {Free-Form}              & {Multi-choice}         \\
\hline {2017}          & {Quasar-S \cite{Dhingra2017Quasar}}                             & {Textual}              & {Cloze}                  & {Spans}                   & {Multi-choice}         \\
\hline {2017}          & {Quasar-T \cite{Dhingra2017Quasar}}                             & {Textual}              & {Natural}                & {Spans}                  & {Natural}              \\
\hline {2017}          & {SearchQA \cite{dunn2017searchqa}}                             & {Textual}              & {Natural}                & {Free-Form}              & {Natural}              \\
\hline {2017}          & {NarrativeQA \cite{kovcisky2018narrativeqa}}                          & {Textual}              & {Natural}                & {Free-Form}              & {Natural}              \\
\hline {2017}          & {SciQ \cite{welbl2017SciQ}}                                 & {Textual}              & {Natural}                & {Spans}                  & {Multi-choice}         \\
\hline {2017}          & {Qangaroo-MedHop \cite{welbl2018qangaroo}}                      & {Textual}              & {Synthetic}              & {Spans}                   & {Multi-choice}         \\
\hline {2017}          & {Qangaroo-WikiHop \cite{welbl2018qangaroo}}                     & {Textual}              & {Synthetic}              & {Spans}                   & {Multi-choice}         \\
\hline {2017}          & {TQA \cite{kembhavi2017tqa}}                                  & {Multi-modal}          & {Natural}                & {Free-Form}              & {Multi-choice}         \\
\hline {2017}          & {COMICS-Coherence \cite{iyyer2017COMICS}}                    & {Multi-modal}          & {Natural}                & {Free-Form}              & {Multi-choice}         \\
\hline {2017}          & {COMICS-Cloze \cite{iyyer2017COMICS}}                        & {Multi-modal}          & {Cloze}                  & {Free-Form}              & {Multi-choice}         \\
\hline {2018}          & {QuAC \cite{choi-etal-2018-quac}}                                 & {Textual}              & {Natural}                & {Free-Form}              & {Natural}              \\
\hline {2018}          & {CoQA \cite{reddy2019coqa}}                                 & {Textual}              & {Natural}                & {Free-Form}              & {Natural}              \\
\hline {2018}          & {SQuAD 2.0 \cite{rajpurkar-etal-2018-know}}                             & {Textual}              & {Natural}                & {Spans}                  & {Natural}              \\
\hline {2018}          & {HotpotQA-Distractor \cite{yang-etal-2018-hotpotqa}}         & {Textual}              & {Natural}                & {Spans}                  & {Natural}              \\
\hline {2018}          & {HotpotQA-Fullwiki \cite{yang-etal-2018-hotpotqa}}           & {Textual}              & {Natural}                & {Spans}                  & {Natural}              \\
\hline {2018}          & {DuoRC-Self \cite{saha-etal-2018-duorc}}                           & {Textual}              & {Natural}                & {Free-Form}              & {Natural}              \\
\hline {2018}          & {DuoRC-Paraphrase \cite{saha-etal-2018-duorc}}                     & {Textual}              & {Natural}                & {Free-Form}              & {Natural}              \\
\hline {2018}          & {CLOTH \cite{xie2018cloth}}                                & {Textual}              & {Cloze}                  & {Free-Form}              & {Natural}              \\
\hline {2018}          & {ReCoRD \cite{zhang2018record}}                               & {Textual}              & {Cloze}                  & {Free-Form}              & {Natural}              \\
\hline {2018}          & {CliCR \cite{suster-daelemans-2018-clicr}}                                & {Textual}              & {Cloze}                  & {Free-Form}              & {Natural}              \\
\hline {2018}          & {ReviewQA \cite{grail2018reviewqa}}                             & {Textual}              & {Natural}                & {Spans}                   & {Multi-choice}         \\
\hline {2018}          & {ARC-Challenge Set \cite{Clark2018ARC}}                    & {Textual}              & {Natural}                & {Free-Form}              & {Multi-choice}         \\
\hline {2018}          & {ARC-Easy Set \cite{Clark2018ARC}}                         & {Textual}              & {Natural}                & {Free-Form}              & {Multi-choice}         \\
\hline {2018}          & {OpenBookQA \cite{OpenBookQA2018}}                           & {Textual}              & {Natural}                & {Free-Form}              & {Multi-choice}         \\
\hline {2018}          & {SciTail \cite{khot2018scitail}}                              & {Textual}              & {Natural}                & {Free-Form}              & {Multi-choice}         \\
\hline {2018}          & {MultiRC \cite{khashabi-etal-2018MultiRC}}                              & {Textual}              & {Natural}                & {Free-Form}              & {Multi-choice}         \\
\hline {2018}          & {RecipeQA-Cloze \cite{yagcioglu-etal-2018-recipeqa}}                      & {Multi-modal}          & {Cloze}                  & {Free-Form}              & {Multi-choice}         \\
\hline {2018}          & {RecipeQA-Coherence \cite{yagcioglu-etal-2018-recipeqa}}                  & {Multi-modal}          & {Natural}                & {Free-Form}              & {Multi-choice}         \\
\hline {2018}          & {PaperQA-Title \cite{park2019can}}         & {Textual}              & {Cloze}                  & {Free-Form}              & {Multi-choice}         \\
\hline {2018}          & {PaperQA-Last \cite{park2019can}} & {Textual}              & {Cloze}                  & {Free-Form}              & {Multi-choice}         \\
\hline {2018}          & {PaperQA(Hong et al.) \cite{hong2018learning}}        & {Textual}              & {Natural}                & {Spans}                  & {Natural}              \\
\hline {2018}          & {MCScript \cite{ostermann-etal-2018-mcscript}}                             & {Textual}              & {Natural}                & {Free-Form}              & {Multi-choice}         \\
\hline {2018}          & {ProPara \cite{dalvi-etal-2018-ProPara}}                              & {Textual}              & {Natural}                & {Spans}                  & {Natural}              \\
\hline {2019}          & {Natural Questions-Short \cite{kwiatkowski2019natural}}       & {Textual}              & {Natural}                & {Spans}                  & {Natural}              \\
\hline {2019}          & {Natural Questions-Long \cite{kwiatkowski2019natural}}        & {Textual}              & {Natural}                & {Spans}                  & {Natural}              \\
\hline {2019}          & {DREAM \cite{sun2019dream}}                                & {Textual}              & {Natural}                & {Free-Form}              & {Multi-choice}         \\
\hline {2019}          & {ShARC \cite{saeidi2018sharc}}                                & {Textual}              & {Natural}                & {Free-Form}              & {Multi-choice}         \\
\hline {2019}          & {CommonSenseQA \cite{talmor-etal-2019-commonsenseqa}}                        & {Textual}              & {Natural}                & {Free-Form}              & {Multi-choice}         \\
\hline {2019}          & {DROP \cite{Dua2019DROP}}                                 & {Textual}              & {Natural}                & {Free-Form}              & {Natural}              \\
		
	\end{supertabular}
\end{center}

\subsubsection{Form of Task vs. Content of Task}
The discussion above is mainly about the form of MRC tasks. However,it should be noted that, besides the form of the MRC task, the content of the context/passage and the question also determine the type of a task. As shown in Figure \ref{figure: babi_tasks }, in the FaceBook BAbi dataset \cite{weston2016babi}, there are many different types of MRC tasks depending on the content of the passages and questions. But because classifying tasks based on the content is a very subjective matter without established standards, herein, we mainly analyze the forms of tasks rather than the content.

\begin{figure}[H]
	\centering
	\begin{tabular}{p{6cm}l p{6cm}l}
		\toprule
		\textbf{Task 1: Yes/No Questions }& \textbf{Task 2: Counting}\\
	\midrule
	John moved to the playground.               & Daniel picked up the football.\\
	Daniel went to the bathroom.                  &Daniel dropped the football.\\
	John went back to the hallway.                 &Daniel got the milk.\\
		& Daniel took the apple.\\
	\midrule
	Is John in the playground? Answer:{\color{red}no}    &How many objects is Daniel holding?\\
	Is Daniel in the bathroom? Answer:{\color{red}yes}	&Answer:{\color{red}two}\\
	\midrule
	\textbf{Task 3: Lists/Sets }& \textbf{Task 4: Indeﬁnite Knowledge}\\
	\midrule
	Daniel picks up the football.     &John is either in the classroom or the playground. \\
	Daniel drops the newspaper.     &Sandra is in the garden.\\
	Daniel picks up the milk.               &     \\
	John took the apple.                   &      \\
	\midrule
	What is Daniel holding?            &Is John in the classroom? Answer:{\color{red} maybe} \\
	Answer:{\color{red}milk, football}	  &  Is John in the ofﬁce? Answer:{\color{red} no}  \\
	\bottomrule
	\end{tabular}
	\caption{Task examples in the Facebook BAbi dataset \cite{weston2016babi}, the types of these tasks are determined by the the content of passages and questions.}
	\label{figure: babi_tasks }
\end{figure}	 

\section{Evaluation Metrics}
\label{sec:Metrics}
\subsection{Overview of Evaluation Metrics}
The most commonly used evaluation metric for MRC models is accuracy. However, in order to more comprehensively compare the performances of MRC models, the models should be evaluated by various evaluation metrics.
In this section, we introduce the calculation methods of commonly used evaluation metrics in machine reading comprehension, which include: Accuracy, Exact Match, Precision, Recall, F1, ROUGE, BLEU, HEQ and Meteor.
For multiple-choice or cloze style tasks, Accuracy is usually used to evaluate MRC models. For span prediction tasks, Exact Match, Precision, Recall, and F1 are usually used as evaluation metrics. 
Currently, many of the evaluation metrics for MRC tasks are derived from other research areas in NLP (natural language processing) such as machine translation and text summaries.
Similar to machine translation tasks, the goal of a MRC task is also to generate some text and compare it with the correct answer. So the evaluation metrics of machine translation tasks can also be used for MRC tasks. In the following sections, we will give detailed calculation methods of these evaluation metrics.

\subsection{Accuracy}
Accuracy represents the percentage of the questions that a MRC system accurately answers. For example, suppose a MRC task contains $N$ questions, each question corresponds to one correct answer, the answers can be a word, a phrases, or a sentence, and the number of questions that the system answers correctly is $M$. The equation for the accuracy is as follows:

\begin{equation}\text {Accuracy}=\frac{M}{N}\end{equation}

\subsection{Exact Match}
If the correct answer to the question is a sentence or a phrase, it is possible that some of the words in the system-generated answer are correct answers, and the other words are not correct answers. In this case, Exact Match represents the percentage of questions that the system-generated answer exactly matches the correct answer, which means every word is the same. Exact Match is often abbreviated as EM.\\
For example, if a MRC task contains $N$ questions, each question corresponds to one right answer, the answers can be a word, a phrases or a sentence, and the number of questions that the system answers correctly is $M$. Among the remaining $N-M$ answers, some of the answers may contain some ground truth answer words, but not exactly match the ground truth answer. The Exact Match can then be calculated as follows:

\begin{equation}\text {Exact Match}=\frac{M}{N}\end{equation}

Therefore, for the span prediction task, Exact Match and Accuracy are exactly the same. But for a multi-choice task, Exact Match is usually not used because there is no situation where the answer includes a portion of the correct answer. In addition, to make the evaluation more reliable, it is also common to collect multiple correct answers for each question. Therefore, the exact match score is only required to match any of the correct answers \cite{chen2018neural}.

\subsection{Precision}
 \subsubsection{Token-level Precision}
 The token-level precision represents the percentage of token overlap between the tokens in the correct answer and the tokens in the predicted answer. Following the evaluation method in SQuAD \cite{rajpurkar2016squad,rajpurkar-etal-2018-know}, we treat the predicted answer and correct answer as bags of tokens, while ignoring all punctuation marks and the article words such as "a" and "an" or "the". In order to get the token-level Precision, we first need to understand the token-level true positive (TP), false positive (FP), true negative (TN), and false negative (FN), as shown in Figure \ref{figure: token_precison }:
 
 \begin{figure}[H]
 	\centering
 	\includegraphics[width=10cm]{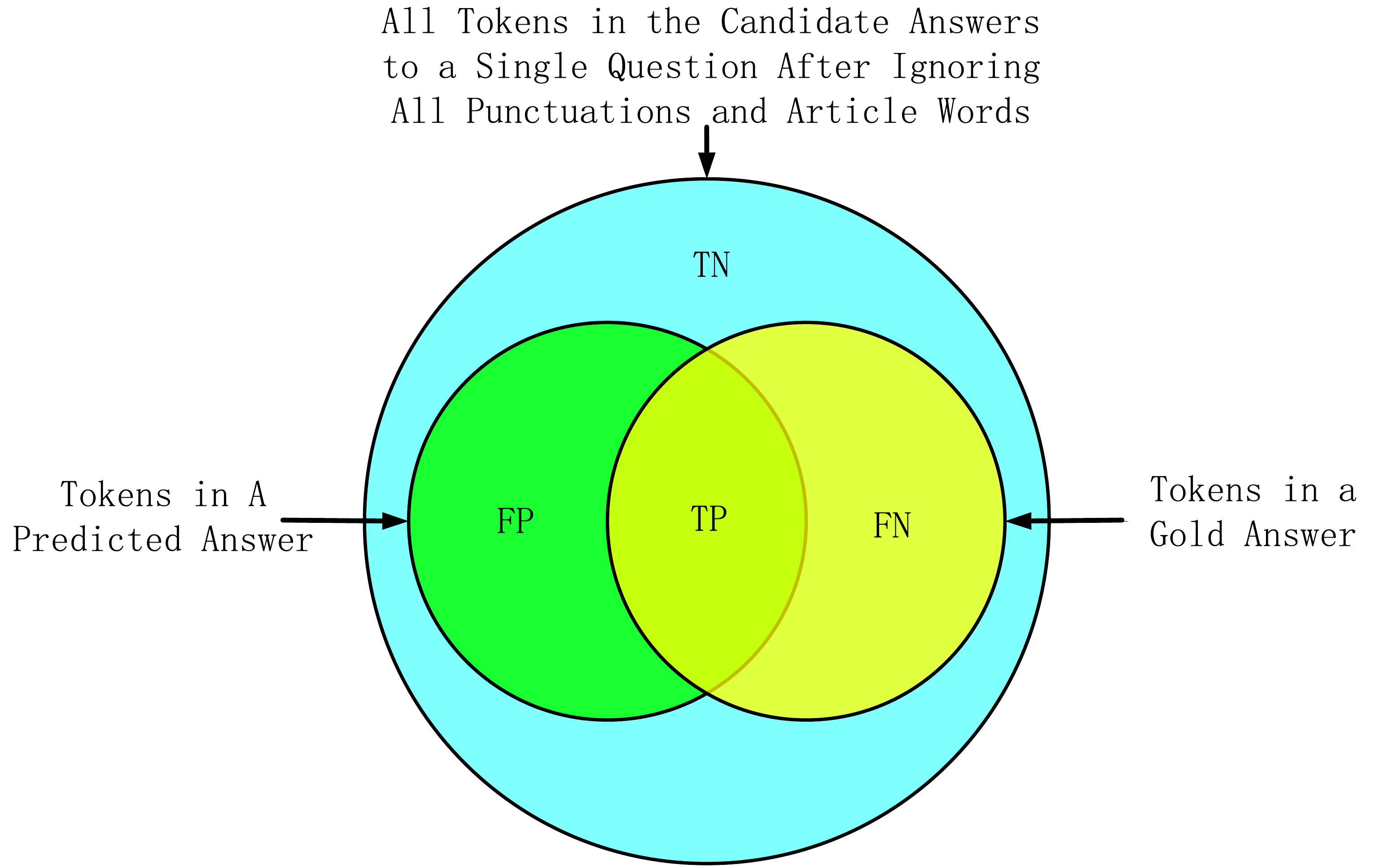}
 	\caption{The token-level true positive (TP), false positive (FP), true negative (TN), and false negative (FN).}
 	\label{figure: token_precison }
 \end{figure}   
 
 As seen in Figure \ref{figure: token_precison }, for a single question, the token-level true positive (TP) denotes the same tokens between the predicted answer and the correct answer. The token-level false positive (FP) denotes the tokens which are not in the correct answer but the predicted answer, while the false negative (FN) denotes the tokens which are not in the predicted answer but the correct answer. A token-level Precision for a single question is computed as follows:

\begin{equation} {Precision}_{T S}=\frac{N u m\big(T P_{T}\big)}{{Num}\big(T P_{T}\big)+{Num}\big(F P_{T}\big)}\end{equation}

Where ${Precision}_{T S}$ denotes the token-level Precision for a single question, and $Num\big(TP_{T}\big)$ denotes the number of token-level true positive (TP) tokens and  $Num\big(FP_{T}\big)$ denotes the number of token-level false positive (FP) tokens. \\
For example, if a correct answer is "a cat in the garden" and the predicted answer is "a dog in the garden". We can see, after ignoring the article word "a" and "the", the number of the shared tokens between the predicted answer and the correct answer is 2, which is also the $Num\big(TP_{T}\big)$ , and $Num\big(FP_{T}\big)$ is 1, so the token-level Precision for this answer is 2/3. 
 
\subsubsection{Question-level Precision}
The question-level precision represents the average percentage of answer overlaps (not token overlap) between all the correct answers and all the predicted answers in a task \cite{yang2015wikiqa}. The question-level true positive (TP), false positive (FP), true negative (TN), and false negative (FN) are shown in Figure \ref{figure: question_precison }:

 \begin{figure}[H]
	\centering
	\includegraphics[width=10cm]{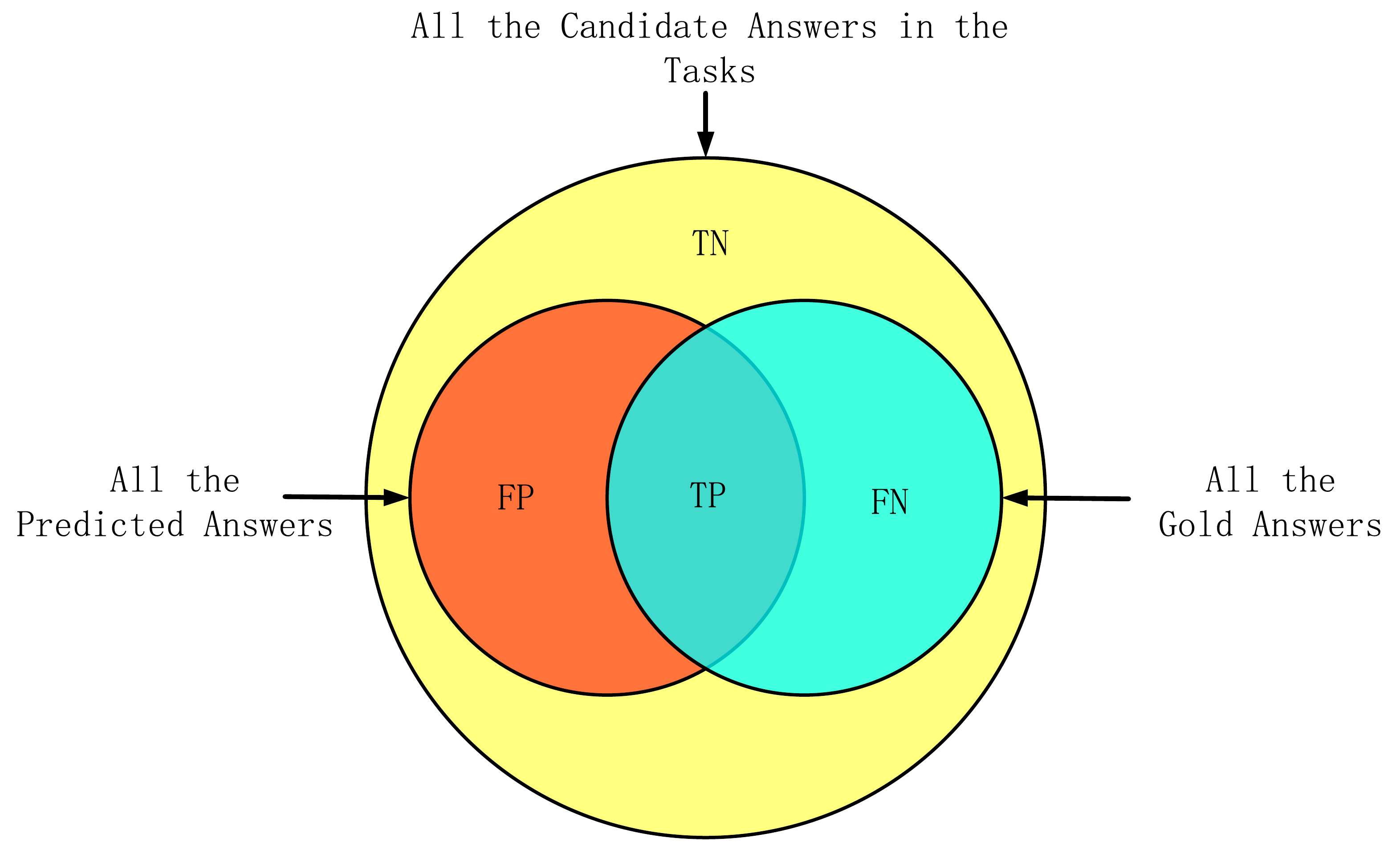}
	\caption{The question-level true positive (TP), false positive (FP), true negative (TN), and false negative (FN).}
	\label{figure: question_precison }
\end{figure}   

As seen in Figure \ref{figure: question_precison }, the question-level true positive (TP) denotes the shared answers between all predicted answers and all correct answers, in which one answer is treated as one entity, no matter how many words it consists of. And the question-level false positive (FP) denotes these predicted answers which do not belong to the set of correct answers, while the question-level false negative (FN) denotes those correct answers which do not belong to the set of predicted answers. A question-level Precision for a task is computed as follows:

\begin{equation} {Precision}_{Q}=\frac{{Num}\big(T P_{Q}\big)}{{Num}\big(T P_{Q}\big)+{Num}\big(F P_{Q}\big)}\end{equation}

Where ${Precision}_{Q}$ denotes the question-level Precision for a task, ${Num}\big(T P_{Q}\big)$ denotes the number of question-level true positive (TP) answers and ${Num}\big(F P_{Q}\big)$ denotes the number of question-level false positive (FP) answers. 

\subsection{Recall}
\subsubsection{Token-level Recall}
The Recall represents the percentage of tokens in a correct answer that have been correctly predicted in a question. Following the definitions of the token-level true positive (TP), false positive (FP), and false negative (FN) above, A token-level Recall for a single answer is computed as follows:

\begin{equation} {Recall}_{T S}=\frac{Num\big(T P_{T}\big)}{{Num}\big(T P_{T}\big)+{Num}\big(F N_{T}\big)}\end{equation}

Where ${Recall}_{TS}$ denotes the token-level Recall for a single question, ${Num}\big(T P_{T}\big)$ denotes the number of token-level true positive (TP) tokens and ${Num}\big(FN_{T}\big)$ denotes the number of token-level false negative (FN) tokens. 

\subsubsection{Question-level Recall}
The question-level Recall represents the percentage of the correct answers that have been correctly predicted in a task \cite{yang2015wikiqa}. Following the definitions of the token-level true positive (TP), false positive (FP), and false negative (FN), A token-level Recall for a single answer is computed as follows:

\begin{equation} {Recall}_{Q}=\frac{Num\big(T P_{Q}\big)}{{Num}\big(T P_{Q}\big)+{Num}\big(F N_{Q}\big)}\end{equation}

Where ${Recall}_{Q}$ denotes the question-level Recall for a task, ${Num}\big(T P_{Q}\big)$ denotes the number of question-level true positive (TP) answers and ${Num}\big(FN_{Q}\big)$ denotes the number of question-level false negative (FN) answers. 

\subsection{F1}
\subsubsection{Token-level F1}
Token-level F1 is a commonly used MRC task evaluation metrics. The equation of token-level F1 for a single question is:

\begin{equation}F 1_{T S}=\frac{2 \times {Precision}_{T S} \times  {Recall}_{T S}}{ {Precision}_{T S}+ {Recall}_{T S}}\end{equation}

Where $F1_{T S}$ denotes the token-level F1 for a single question, ${Precision}_{T S}$ denotes the token-level Precision for a single question and ${Recall}_{T S}$ denotes the token-level Recall for a single question.\\

To make the evaluation more reliable, it is also common to collect multiple correct answers to each question \cite{chen2018neural}. Therefore, to get the average token-level F1, we first have to compute the maximum token-level F1 of all the correct answers of a question, and then average these maximum token-level F1 over all of the questions \cite{chen2018neural}. The equation of average token-level F1 for a task is:

\begin{equation}F 1_{T}=\frac{\sum Max\big( {Precision}_{T S}\big)}{ {Num}( {Questions})}\end{equation}

Where $F 1_{T}$ denotes the average token-level F1 for a task, and ${Max\big({Precision}_{T S}\big)}$ denotes the maximum token-level F1 of all the correct answers for a single question, ${\sum Max\big({Precision}_{T S}\big)}$ denotes the sum of for every question in the task. $ {Num}( {Questions})$ denotes the number of questions in the task.

\subsubsection{Question-level F1}
The equation of question-level F1 for a task is:

\begin{equation}F 1_{Q}=\frac{2 \times  {Precision}_{Q} \times  {Recall}_{Q}}{ {Precision}_{Q}+ {Recall}_{Q}}\end{equation}

Where $F 1_{Q}$ denotes the question-level F1, ${Precision}_{Q}$ denotes the question-level Precision for a task and ${Recall}_{Q}$ denotes the question-level Recall for a task. 
 
\subsection{ROUGE}
ROUGE stands for Recall-Oriented Understudy for Gisting Evaluation, which was first proposed by Chin-Yew Lin \cite{lin-2004-rouge}. In this paper, ROUGE was used to evaluate the performance of text summary systems. Currently, ROUGE is also used in the evaluation of MRC systems.\\

ROUGE-N is a $n$-gram Recall between a candidate summary and a set of reference summaries \cite{lin-2004-rouge}.
According to the value of $n$, ROUGE is specifically divided into ROUGE-1, ROUGE-2, ROUGE-3, and so on. The ROUGE-N is computed as follows:

\begin{equation}\text {ROUGE-N}=\frac{\sum_{S \in\{R S\}} \sum_{g r a m_{n} \in S}  { Count }_{ {match}}\big( {gram}_{n}\big)}{\sum_{S \in\{R S\}} \sum_{ {gram}_{n} \in S}  { Count }\big( {gram}_{n}\big)}\end{equation}

Where $n$ is the length of the $n$-gram, ${Count }\big( {gram}_{n}\big)$ is the maximum number of times the $n$-gram appears in the candidate text and predicted text generated by the algorithm, and $RS$ is an abbreviation of $ReferenceSummaries$.

\subsection{BLEU}
BLEU (Bilingual Evaluation Understudy) was proposed by Papineni et al. \cite{papineni-etal-2002-bleu}. In the original paper, BLEU was used to evaluate the performance of machine translation systems. Currently, BLEU is also used in the performance evaluation of MRC.\\
The computation method of BLEU is to take the geometric mean of the modified Precision and then multiply the result by an exponential brevity penalty factor. Currently, case folding is the only text normalization performed before computing the precision. First, we compute the geometric average of the modified $n$-gram precision, $P_{n}$, using $n$-grams up to length $N$ and positive weights $w_{n}$ summing to one \cite{papineni-etal-2002-bleu}.\\
Next, let $C$ be the length of the candidate sentence and $r$ be the length of the effective reference corpus. The brevity penalty $BP$ is computed as follows[bib BLEU]:\\

\begin{equation}B P=\left\{\begin{array}{ll}
1 & \text { if } c>r \\
e^{(1-r / c)} & \text { if } c \leq r
\end{array}\right.\end{equation}

Then:

\begin{equation}B L E U=B P \cdot \exp \left(\sum_{n=1}^{N} w_{n} \log p_{n}\right)\end{equation}

\subsection{Meteor}
Meteor was first proposed by Banerjee and Lavie \cite{banerjee-lavie-2005-meteor} in order to evaluate the machine translation system. Unlike the BLEU using only Precision, the Meteor indicator uses a combination of Recall and Accuracy to evaluate the system. In addition, Meteors also include features such as synonym matching. \\
Besides Meteor, Denkowski and Lavie also proposed Meteor-next \cite{denkowski-lavie-2010-meteor} and Meteor 1.3 \cite{denkowski-lavie-2011-meteor}, the new metric features include improved text normalization, higher-precision paraphrase matching, and discrimination between content and function words. Currently, some MRC datasets use Meteor as one of their evaluation metrics, such as the NarrativeQA \cite{kovcisky2018narrativeqa} dataset. The Meteor score for the given alignment is computed as follows:

\begin{equation}{Meteor}=F_{ {mean}} \times\big(1- {Penalty}\big)\end{equation}

Where $F_{mean}$ is combined by the $Precision$ and $Recall$ via a harmonic-mean \cite{van1979information} that places most of the weight on $Recall$, and the formula of $F_{mean}$ is:
		
\begin{equation}F_{mean}=\frac{{Precision} \times {Recall}}{\alpha \times {Precision}+(1-\alpha) \times {Recall}}\end{equation}

And $Penalty$ is a fragmentation penalty to account for differences and gaps in word order, which is calculated using the total number of matched words ($m$, average over hypothesis and reference) and number of chunks ($ch$): 

\begin{equation}{ Penalty }=\gamma \times\left(\frac{c h}{m}\right)^{\beta}\end{equation}

Where the parameters $\alpha$, $\beta$, and $\gamma$ are tuned to maximize correlation with human judgments \cite{denkowski-lavie-2011-meteor}.
It should be noted that the $Precision$ and $Recall$ in Meteor 1.3 is improved by text normalization, we can see the original paper of Denkowski and Lavie for the detailed calculation method of $Precision$ and $Recall$ in Meteor 1.3 \cite{denkowski-lavie-2011-meteor}.

\subsection{HEQ}
The HEQ stands for Human Equivalence Score, which is a new MRC evaluation metric that can be used in conversational reading comprehension datasets, such as QuAC \cite{choi-etal-2018-quac}. For these datasets in which questions with multiple valid answers, the F1 may be misleading. Therefore, HEQ was introduced. The HEQ is an evaluation metric for judging whether the output of the system is as good as the output of an ordinary person. For example, suppose a MRC task contains $N$ questions, and the number of questions for which the token-level $F1$ performance of algorithm exceeds or reaches the token-level $F1$ of humans is $M$. The HEQ score is computed as follows \cite{choi-etal-2018-quac}:

\begin{equation}HEQ=\frac{M}{N}\end{equation}

\subsection{Statistics of Evaluation Metrics}
In this section, we collated the evaluation metrics of 57 MRC tasks. As seen in Table \ref{table: metrics }, the typical MRC dataset evaluation metrics are Accuracy, Exact Match, F1 score, ROUGE, BLEU, HEQ, and Meteor. Many datasets use more than one evaluation metric. Moreover, some datasets adopt detailed evaluation metrics according to their own characteristics. For example, the HotpotQA \cite{yang-etal-2018-hotpotqa} dataset adopts evaluation metrics such as Exact Match of Supportings, F1 of Supportings, Exact Match of Answer, F1 of Answer, etc. And the Facebook CBT \cite{hill2016cbt} dataset adopts Accuracy on Named Entities, Accuracy on Common Nouns, Accuracy on Verbs, Accuracy on Prepositions.

\begin{center}
	\renewcommand\arraystretch{1.1} 
	\topcaption{Evaluation metrics of different machine reading comprehension tasks.}
	\label{table: metrics }
	\tablehead{
		\toprule{\textbf{Year}} & {\textbf{MRC Tasks}}& {\textbf{Metric 1}} & {\textbf{Metric 2}} & {\textbf{Metric 3}} & {\textbf{Metric 4}}\\
	}
	\tabletail{	
		\hline
		\multicolumn{6}{r}{\small\sl continued on next page}\\
		\hline 
	}
	
	\tablefirsthead{
		\toprule{\textbf{Year}} & {\textbf{MRC Tasks}}& {\textbf{Metric 1}} & {\textbf{Metric 2}} & {\textbf{Metric 3}} & {\textbf{Metric 4}}\\ \midrule}
	
	\tablelasttail{\bottomrule}
	
	\begin{supertabular}{p{0.8cm} p{3cm} p{2cm} p{2cm} p{2cm}  p{2cm}}	
     {2013} & {MCTest \cite{richardson2013mctest}} & {Accuracy} & {N/A} & {N/A} & {N/A} \\
	\hline {2015} & {CNN/Daily Mail \cite{hermann2015cnn}} & {Accuracy} & {N/A} & {N/A} & {N/A} \\
	\hline {2015} & {CuratedTREC \cite{baudivs2015CuratedTREC}} & {Exact Match} & {N/A} & {N/A} & {N/A} \\
	\hline {2015} & {WikiQA \cite{yang2015wikiqa}} & {Question-level Precision} & {Question-level Recall} & {Question-level F1} & {N/A} \\
	\hline {2016} & {BookTest \cite{bajgar2016embracing}} & {Accuracy} & {N/A} & {N/A} & {N/A} \\
	\hline {2016} & {Facebook CBT \cite{hill2016cbt}} & {Accuracy on Named Entities} & {Accuracy on Common Nouns} & {Accuracy on Verbs} & {Accuracy on Prepositions} \\
	\hline {2016} & {Google MC-AFP \cite{soricut2016building}} & {Accuracy} & {N/A} & {N/A} & {N/A} \\
	\hline {2016} & {LAMBADA \cite{boleda2016lambada}} & {Accuracy} & {N/A} & {N/A} & {N/A} \\
	\hline {2016} & {MovieQA \cite{tapaswi2016movieqa}} & {Accuracy of Video Clips} & {Accuracy of Plots and Subtitles} & {N/A} & {N/A} \\
	\hline {2016} & {MS MARCO \cite{tri2016MARCO}} & {Rouge-L} & {BLEU-1} & {N/A} & {N/A} \\
	\hline {2016} & {NewsQA \cite{trischler-etal-2017-newsqa}} & {Exact Match} & {Token-level F1} & {N/A} & {N/A} \\
	\hline {2016} & {SQuAD 1.1 \cite{rajpurkar2016squad}} & {Token-level F1} & {Exact Match} & {N/A} & {N/A} \\
	\hline {2016} & {Who-did-What \cite{onishi2016whodidwhat}} & {Accuracy} & {N/A} & {N/A} & {N/A} \\
	\hline {2016} & {WikiMovies \cite{miller2016WikiMovies}} & {Accuracy} & {N/A} & {N/A} & {N/A} \\
	\hline {2016} & {WikiReading \cite{hewlett-etal-2016-wikireading}} & {Question level F1} & {N/A} & {N/A} & {N/A} \\
	\hline {2017} & {COMICS-Cl \cite{iyyer2017COMICS}} & {Accuracy of Text Cloze} & {Accuracy of Visual Cloze} & {N/A} & {N/A} \\
	\hline {2017} & {COMICS-Co \cite{iyyer2017COMICS}} & {Accuracy of Coherence} & {N/A} & {N/A} & {N/A} \\
	\hline {2017} & {NarrativeQA \cite{kovcisky2018narrativeqa}} & {ROUGE-L} & {BLEU-1} & {BLEU-4} & {Meteor} \\
	\hline {2017} & {Qangaroo-M \cite{welbl2018qangaroo}} & {Accuracy} & {N/A} & {N/A} & {N/A} \\
	\hline {2017} & {Qangaroo-W \cite{welbl2018qangaroo}} & {Accuracy} & {N/A} & {N/A} & {N/A} \\
	\hline {2017} & {Quasar-S \cite{Dhingra2017Quasar}} & {Accuracy} & {N/A} & {N/A} & {N/A} \\
	\hline {2017} & {Quasar-T \cite{Dhingra2017Quasar}} & {Exact Match} & {Token-level F1} & {N/A} & {N/A} \\
	\hline {2017} & {RACE \cite{lai-etal-2017-race}} & {Accuracy} & {N/A} & {N/A} & {N/A} \\
	\hline {2017} & {SciQ \cite{welbl2017SciQ}} & {Accuracy} & {N/A} & {N/A} & {N/A} \\
	\hline {2017} & {SearchQA \cite{dunn2017searchqa}} & {F1 score (for n-gram)} & {Accuracy} & {N/A} & {N/A} \\
	\hline {2017} & {TQA \cite{kembhavi2017tqa}} & {Accuracy of All} & {Accuracy of Diagram} & {N/A} & {N/A} \\
	\hline {2017} & {TriviaQA-Wiki \cite{joshi-etal-2017-triviaqa}} & {Exact Match} & {Question-level F1} & {Verified-EM} & {Verified-F1} \\
	\hline {2017} & {TriviaQA-Web \cite{joshi-etal-2017-triviaqa}} & {Exact Match} & {Document-level F1} & {Verified-EM} & {Verified-F1} \\
	\hline {2018} & {ARC-C \cite{Clark2018ARC}} & {Accuracy} & {N/A} & {N/A} & {N/A} \\
	\hline {2018} & {ARC-E \cite{Clark2018ARC}} & {Accuracy} & {N/A} & {N/A} & {N/A} \\
	\hline {2018} & {CliCR \cite{suster-daelemans-2018-clicr}} & {Exact Match} & {Token-level F1} & {BLEU-2} & {BLEU-4} \\
	\hline {2018} & {CLOTH \cite{xie2018cloth}} & {Accuracy} & {N/A} & {N/A} & {N/A} \\
	\hline {2018} & {CoQA \cite{reddy2019coqa}} & {Token-level F1} & {F1 out of domain} & {F1 in domain} & {N/A} \\
	\hline {2018} & {DuoRC-P \cite{saha-etal-2018-duorc}} & {Accuracy} & {Token-level F1} & {N/A} & {N/A} \\
	\hline {2018} & {DuoRC-S \cite{saha-etal-2018-duorc}} & {Accuracy} & {Token-level F1} & {N/A} & {N/A} \\
	\hline {2018} & {HotpotQA-D \cite{yang-etal-2018-hotpotqa}} & {EM of Answer} & {F1 of Answer (Token-level)} & {EM of Supportings} & {F1 of Supportings} \\
	\hline {2018} & {HotpotQA-F \cite{yang-etal-2018-hotpotqa}} & {EM of Answer} & {F1 of Answer (Token-level)} & {EM of Supportings} & {F1 of Supportings} \\
	\hline {2018} & {MCScript \cite{ostermann-etal-2018-mcscript}} & {Accuracy} & {N/A} & {N/A} & {N/A} \\
	\hline {2018} & {MultiRC \cite{khashabi-etal-2018MultiRC}} & {F1m} & {Exact Match} & {N/A} & {N/A} \\
	\hline {2018} & {OpenBookQA \cite{OpenBookQA2018}} & {Accuracy} & {N/A} & {N/A} & {N/A} \\
	\hline {2018} & {PaperQA(Hong et al.) \cite{hong2018learning}} & {F1} & {N/A} & {N/A} & {N/A} \\
	\hline {2018} & {PaperQA-LS \cite{park2019can}} & {Accuracy} & {N/A} & {N/A} & {N/A} \\
	\hline {2018} & {PaperQA-T \cite{park2019can}} & {Accuracy} & {N/A} & {N/A} & {N/A} \\
	\hline {2018} & {ProPara \cite{dalvi-etal-2018-ProPara}} & {Accuracy} & {N/A} & {N/A} & {N/A} \\
	\hline {2018} & {QuAC \cite{choi-etal-2018-quac}} & {Token-level F1} & {HEQ-Q} & {HEQ-D} & {N/A} \\
	\hline {2018} & {RecipeQA-Cl \cite{yagcioglu-etal-2018-recipeqa}} & {Accuracy of Textual Cloze} & {Accuracy of Visual Cloze} & {N/A} & {N/A} \\
	\hline {2018} & {RecipeQA-Co \cite{yagcioglu-etal-2018-recipeqa}} & {Accuracy-VO} & {Accuracy-VC} & {N/A} & {N/A} \\
	\hline {2018} & {ReCoRD \cite{zhang2018record}} & {Exact Match} & {Token-level F1} & {N/A} & {N/A} \\
	\hline {2018} & {ReviewQA \cite{grail2018reviewqa}} & {Accuracy} & {N/A} & {N/A} & {N/A} \\
	\hline {2018} & {SciTail \cite{khot2018scitail}} & {Accuracy} & {N/A} & {N/A} & {N/A} \\
	\hline {2018} & {SQuAD 2.0 \cite{rajpurkar-etal-2018-know}} & {Token-level F1} & {EM} & {N/A} & {N/A} \\
	\hline {2019} & {CommonSenseQA \cite{talmor-etal-2019-commonsenseqa}} & {Accuracy} & {N/A} & {N/A} & {N/A} \\
	\hline {2019} & {DREAM \cite{sun2019dream}} & {Accuracy} & {N/A} & {N/A} & {N/A} \\
	\hline {2019} & {DROP \cite{Dua2019DROP}} & {EM} & {Token-level F1} & {N/A} & {N/A} \\
	\hline {2019} & {Natural Questions-Long \cite{kwiatkowski2019natural}} & {Precision} & {Recall} & {N/A} & {N/A} \\
	\hline {2019} & {Natural Questions-Short \cite{kwiatkowski2019natural}} & {Precision} & {Recall} & {F1} & {N/A} \\
	\hline {2019} & {ShARC \cite{saeidi2018sharc}} & {Micro Accuracy} & {Macro Accuracy} & {BLEU-1} & {BLEU-4} \\ 
	\end{supertabular}
\end{center}

Table \ref{table: metrics_usage } shows the statistics on the usage of different evaluation metrics in the 57 MRC tasks collected in this paper. Among them, Accuracy is the most widely used evaluation metric, and 61.40\% of MRC tasks collected in this paper used it. It is followed by F1 (36.84\%) and Exact Match (22.81\%). The rest of these evaluation metrics are less used, as shown in Table \ref{table: metrics_usage }:

\begin{table}[H]
	\caption{Statistics on the usage of different evaluation metrics in 57 machine reading comprehension tasks.}
	\centering
	\label{table: metrics_usage }
	\begin{tabular}{p{1.1cm}| p{1.1cm} p{1.1cm} p{1.1cm} p{1.1cm} p{1.1cm}  p{1.1cm}  p{1.1cm} p{1.1cm} p{1.1cm}}
		\toprule
	    Metrics & Accuracy & F1 & EM & BLEU & Recall & Precision & ROUGE-L & HEQ-D & Meteor \\ 
		\midrule
		Usage & 61.40\% & 36.84\% & 22.81\% & 7.02\% & 5.26\% & 5.26\% & 3.51\% & 1.75\% & 1.75\% \\ 
		\bottomrule
	\end{tabular}
\end{table}

\begin{figure}[H]
	\centering
	\includegraphics[width=15cm]{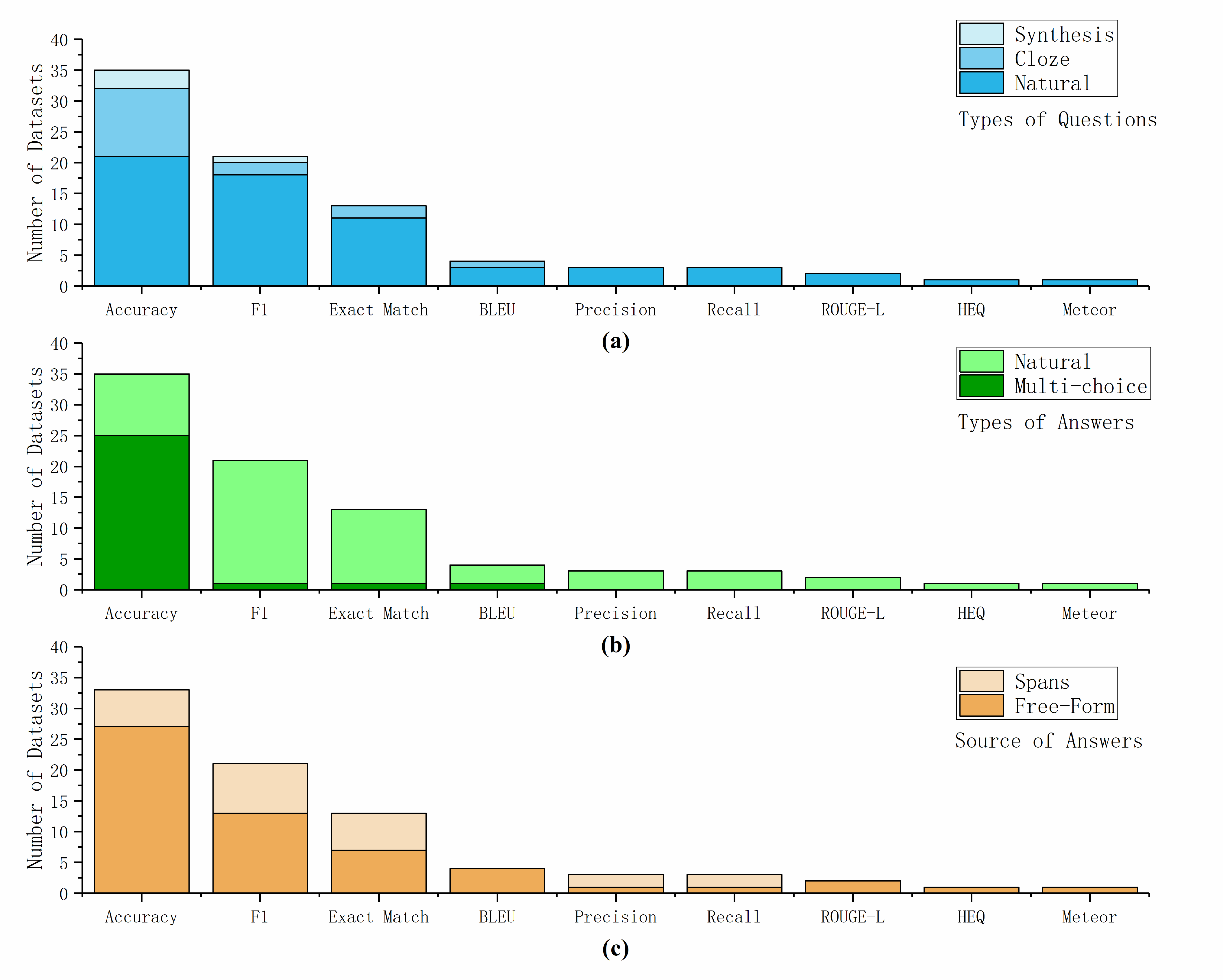}
	\caption{The usage of evaluation metrics with different types of tasks. Different colors represent different types of tasks. }
	\label{figure: metrics_task_usage }
\end{figure}

We also analyzed the relationship between the evaluation metrics and task types. Figure \ref{figure: metrics_task_usage } shows the usage of evaluation metrics with different types of tasks. Taking the "Accuracy" in Figure \ref{figure: metrics_task_usage } (b) as an example, a total of 35 MRC tasks use the "Accuracy" as the evaluation metric. Among them, 25 tasks have the "Multi-choice" type of answers, and the remaining 10 tasks have the "Natural" type of answers. It can be seen from Figure \ref{figure: metrics_task_usage } (b) that tasks with the "Multi-choice" type of answers prefer to use the "Accuracy" evaluation metric rather than other evaluation metrics. This is because it is impossible to calculate the EM, Precision, BLEU or F1 score of a typical "Multi-choice" question which has only one correct answer in the candidates. Among the "Multi-choice" tasks we collected, only the MultiRC \cite{khashabi-etal-2018MultiRC} task does not use Accuracy, but F1 and Exact Match as the evaluation metric. That is because there are multiple correct answers in the candidates of the MultiRC task. As can be seen from Figure \ref{figure: metrics_task_usage } (a), tasks with "Cloze" questions prefer to use the "Accuracy" as evaluation metrics rather than other evaluation metrics, which is because "Cloze" tasks tend to have "Multi-choice" answers. From Figure \ref{figure: metrics_task_usage } (c), we can see that tasks with "Spans" answers and tasks with "Free-form" answers have no special preference in selecting evaluation metrics.

\section{Benchmark Dataset}
\label{sec:Dataset}
\label{sec: Datasets}
\begin{figure}[H]
	\centering
	\includegraphics[width=15cm]{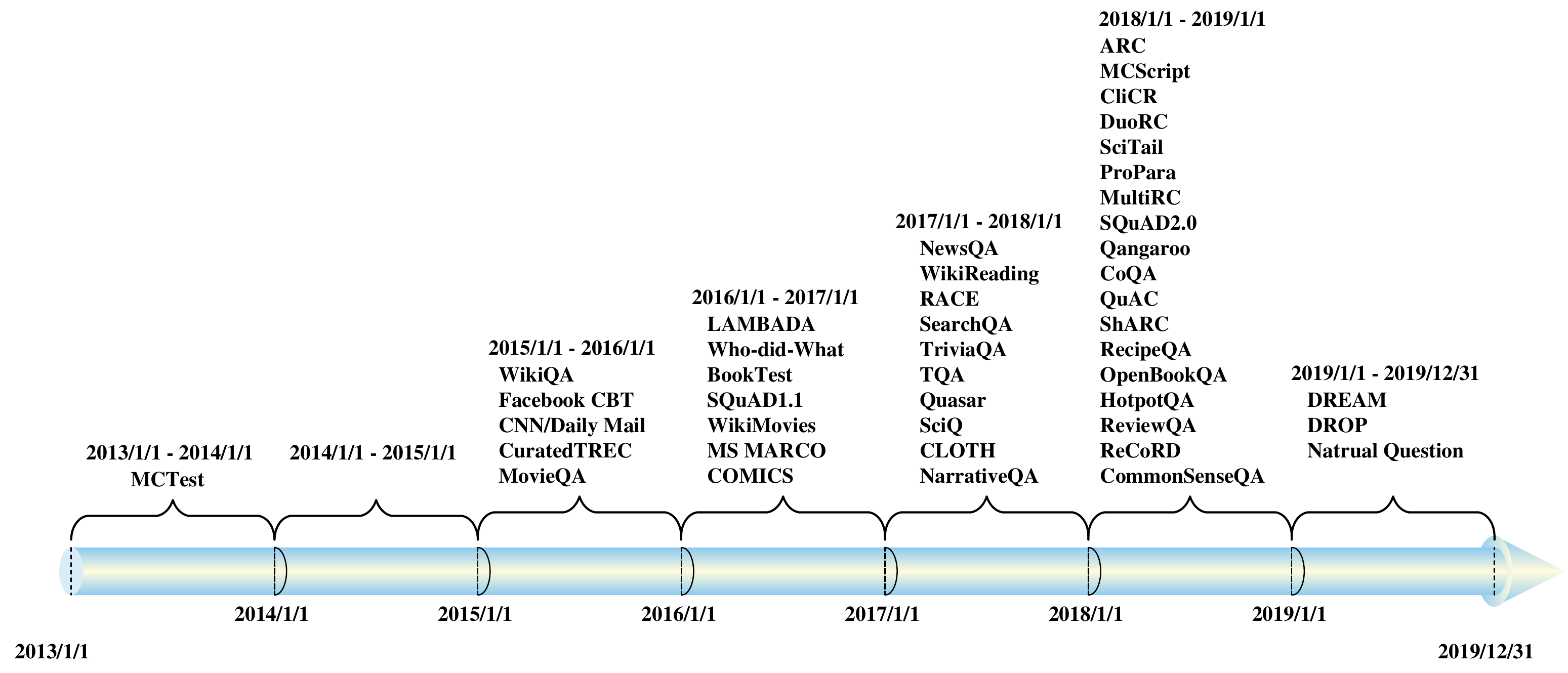}
	\caption{The timeline of MRC datasets discussed in this survey.}
	\label{figure: Timeline}
\end{figure}

In this section, we analyze various attributes of 57 MRC benchmark datasets, including dataset size, generation method, source of corpus, context type, availability of leaderboards and baselines, prerequisite skills, and citations of related papers. We have provided the timeline figure of the MRC datasets, as seen in Figure \ref{figure: Timeline}.

\subsection{The Size of Datasets}
The recent success of machine reading comprehension is driven largely by both large-scale datasets and neural models \cite{chen2018neural}. The size of a dataset affects the generalization ability of the MRC model and determines whether the model is useful in the real world. Early MRC datasets tend to of small sizes. With the continuous development of MRC datasets in recent years, the question set sizes of newly created MRC datasets are generally more than 10K. Here, we have counted the total number of questions in each MRC dataset along with the sizes of its training set, development set, and testing set, as well as the proportion of training set to the total number of questions. The data is shown in Table \ref{table: question_size } which is sorted by the question set size of the datasets. 

\begin{center}
	\renewcommand\arraystretch{1.2} 
	\topcaption{The question set size of machine reading comprehension datasets.}
	\label{table: question_size }
	\tablehead{
		\toprule{\textbf{Year}} & {\textbf{Datasets}}& {\textbf{Question size}} & {\textbf{\#Training questions}} & {\textbf{\#Dev questions}} & {\textbf{\#Test  questions}}     & {\textbf{Percentage  of Training set}}  \\
	}
	\tabletail{	
		\hline
		\multicolumn{7}{r}{\small\sl continued on next page}\\
		\hline 
	}
	
	\tablefirsthead{
		\toprule{\textbf{Year}} & {\textbf{Datasets}}& {\textbf{Question size}} & {\textbf{\#Training questions}} & {\textbf{\#Dev questions}} & {\textbf{\#Test  questions}}     & {\textbf{Percentage  of Training set}}  \\
		\midrule}
	
	\tablelasttail{\bottomrule}
	
	\begin{supertabular}{p{0.8cm} p{2.4cm} p{1.5cm} p{1.5cm} p{1.5cm}  p{1.5cm} p{2cm}}	
		
		{2016}          & {WikiReading \cite{hewlett-etal-2016-wikireading}}                          & {18.87M}                   & {16.03M}                          & {1.89M}                      & {0.95M}                       & {84.95\%}                      \\
		\hline {2016}          & {BookTest \cite{bajgar2016embracing}}                             & {14,160,825}               & {14,140,825}                      & {10,000}                     & {10,000}                      & {99.86\%}                      \\
		\hline {2016}          & {Google MC-AFP \cite{soricut2016building}}                        & {1,742,618}                & {1,727,423}                       & {7,602}                      & {7,593}                       & {99.13\%}                      \\
		\hline {2015}          & {Daily Mail\cite{hermann2015cnn}}                           & {997,467}                  & {879,450}                         & {64,835}                     & {53,182}                      & {88.17\%}                      \\
		\hline {2016}          & {Facebook CBT \cite{hill2016cbt}}                         & {687K}                     & {669,343}                         & {8,000}                      & {10,000}                      & {97.38\%}                      \\
		\hline {2018}          & {ReviewQA \cite{grail2018reviewqa}}                             & {587,492}                  & {528,665}                         & {N/A}                        & {58,827}                      & {89.99\%}                      \\
		\hline {2015}          & {CNN \cite{hermann2015cnn}}                                  & {387,420}                  & {380,298}                         & {3,924}                      & {3,198}                       & {98.16\%}                      \\
		\hline {2019}          & {Natural Questions \cite{kwiatkowski2019natural}}                  & {323,045}                  & {307,373}                         & {7,830}                      & {7,842}                       & {95.15\%}                      \\
		\hline {2016}          & {Who-did-What \cite{onishi2016whodidwhat}}                         & {147,786}                  & {127,786}                         & {10,000}                     & {10,000}                      & {86.47\%}                      \\
		\hline {2018}          & {SQuAD 2.0 \cite{rajpurkar-etal-2018-know}}                             & {151,054}                  & {130,319}                         & {11,873}                     & {8,862}                       & {86.27\%}                      \\
		\hline {2017}          & {SearchQA \cite{dunn2017searchqa}}                             & {140,461}                  & {99,820}                          & {13,393}                     & {27,248}                      & {71.07\%}                      \\
		\hline {2018}          & {CoQA \cite{reddy2019coqa}}                                 & {127K}                     & {110K}                            & {7K}                         & {10K}                         & {86.61\%}                      \\
		\hline {2018}          & {ReCoRD \cite{zhang2018record}}                               & {120,730}                  & {100,730}                         & {10,000}                     & {10,000}                      & {83.43\%}                      \\
		\hline {2016}          & {NewsQA \cite{trischler-etal-2017-newsqa}}                               & {119K}                     & {107K}                            & {6K}                         & {6K}                          & {89.92\%}                      \\
		
		\hline {2018}          & {HotpotQA \cite{yang-etal-2018-hotpotqa}}                          & {105,374}                  & {90,564}                          & {7,405}                      & {7,405}                       & {85.95\%}                      \\
		\hline {2018}          & {CliCR \cite{suster-daelemans-2018-clicr}}                                & {104,919}                  & {91,344}                          & {6,391}                      & {7,184}                       & {87.06\%}                      \\
		\hline {2018}          & {DuoRC-P \cite{saha-etal-2018-duorc}}                     & {100,316}                  & {70K}                             & {15K}                        & {15K}                         & {70.00\%}                      \\
		\hline {2016}          & {SQuAD 1.1 \cite{rajpurkar2016squad}}                             & {107,702}                  & {87,599}                          & {10,570}                     & {9,533}                       & {81.33\%}                      \\
		\hline {2016}          & {WikiMovies \cite{miller2016WikiMovies}}                           & {116K}                     & {96K}                             & {10K}                        & {10K}                         & {82.76\%}                      \\
		\hline {2018}          & {CLOTH \cite{xie2018cloth}}                                & {99,433}                   & {76,850}                          & {11,067}                     & {11,516}                      & {77.29\%}                      \\
		\hline {2018}          & {QuAC \cite{choi-etal-2018-quac}}                                 & {98,275}                   & {83,568}                          & {7,354}                      & {7,353}                       & {85.03\%}                      \\
		\hline {2017}          & {RACE \cite{lai-etal-2017-race}}                                 & {97,687}                   & {87,866}                          & {4,887}                      & {4,934}                       & {89.95\%}                      \\
		\hline {2019}          & {DROP \cite{Dua2019DROP}}                                 & {96,567}                   & {77,409}                          & {9,536}                      & {9,622}                       & {80.16\%}                      \\
		\hline {2017}          & {TriviaQA-Web \cite{joshi-etal-2017-triviaqa}}                        & {95,956}                   & {76,496}                          & {9,951}                      & {9,509}                       & {79.72\%}                      \\
		\hline {2018}          & {PaperQA-T \cite{park2019can}}         & {84,803}                   & {77,298}                          & {3,752}                      & {3,753}                       & {91.15\%}                      \\
		\hline {2018}          & {DuoRC-S \cite{saha-etal-2018-duorc}}                           & {84K}                   & {60K}                             & {12K}                        & {12K}                         & {70.00\%}                      \\
		\hline {2018}          & {PaperQA-L \cite{park2019can}} & {80,118}                   & {71,804}                          & {4,179}                      & {4,135}                       & {89.62\%}                      \\
		\hline {2017}          & {TriviaQA-Wiki \cite{joshi-etal-2017-triviaqa}}                      & {77,582}                   & {61,888}                          & {7,993}                      & {7,701}                       & {79.77\%}                      \\
		\hline {2017}          & {Qangaroo-W \cite{welbl2018qangaroo}}                     & {51,318}                   & {43,738}                          & {5,129}                      & {2,451}                       & {85.23\%}                      \\
		\hline {2017}          & {NarrativeQA \cite{kovcisky2018narrativeqa}}                          & {46,765}                   & {32,747}                          & {3,461}                      & {10,557}                      & {70.02\%}                      \\
		\hline {2017}          & {Quasar-T \cite{Dhingra2017Quasar}}                             & {43,013}                   & {37,012}                          & {3,000}                      & {3,000}                       & {86.05\%}                      \\
		\hline {2017}          & {Quasar-S \cite{Dhingra2017Quasar}}                             & {37,362}                   & {31,049}                          & {3,174}                      & {3,139}                       & {83.10\%}                      \\
		\hline {2018}          & {RecipeQA \cite{yagcioglu-etal-2018-recipeqa}}                             & {36K}                      & {29,657}                          & {3,562}                      & {3,567}                       & {80.62\%}                      \\
		\hline {2017}          & {TQA \cite{kembhavi2017tqa}}                                  & {26,260}                   & {15,154}                          & {5,309}                      & {5,797}                       & {57.71\%}                      \\
		\hline {2016}          & {MovieQA \cite{tapaswi2016movieqa}}                              & {21,406}                   & {14,166}                          & {2,844}                      & {4,396}                       & {66.18\%}                      \\
		\hline {2018}          & {MCScript \cite{ostermann-etal-2018-mcscript}}                             & {13,939}                   & {9,731}                           & {1,411}                      & {2,797}                       & {69.81\%}                      \\
		\hline {2017}          & {SciQ \cite{welbl2017SciQ}}                         & {13,679}                   & {11,679}                          & {1,000}                      & {1,000}                       & {85.38\%}                      \\
		\hline {2019}          & {CommonSenseQA \cite{talmor-etal-2019-commonsenseqa}}                        & {12,102}                   & {9741}                            & {1221}                       & {1140}                        & {80.49\%}                      \\
		\hline {2019}          & {DREAM \cite{sun2019dream}}                                & {10,197}                   & {6,116}                           & {2,040}                      & {2,041}                       & {59.98\%}                      \\
		\hline {2018}          & {OpenBookQA \cite{OpenBookQA2018}}                           & {5,957}                    & {4,957}                           & {500}                        & {500}                         & {83.21\%}                      \\
		\hline {2018}          & {ARC-Easy Set \cite{Clark2018ARC}}                         & {5,197}                    & {2,251}                           & {570}                        & {2,376}                       & {43.31\%}                      \\
		\hline {2015}          & {WikiQA \cite{yang2015wikiqa}}                               & {3,047}                    & {2,118}                           & {296}                        & {633}                         & {69.51\%}                      \\
		\hline {2018}          & {ARC-Challenge Set \cite{Clark2018ARC}}                    & {2,590}                    & {1,119}                           & {299}                        & {1,172}                       & {43.20\%}                      \\
		\hline {2017}          & {Qangaroo-M \cite{welbl2018qangaroo}}                      & {2,508}                    & {1,620}                           & {342}                        & {546}                         & {64.59\%}                      \\
		\hline {2013}          & {MCTest-mc500 \cite{richardson2013mctest}}                         & {2,000}                    & {1,200}                           & {200}                        & {600}                         & {60.00\%}                      \\
		\hline {2018}          & {SciTail \cite{khot2018scitail}}                              & {1,834}                    & {1,542}                           & {121}                        & {171}                         & {84.08\%}                      \\
		\hline {2019}          & {ShARC \cite{saeidi2018sharc}}                                & {948}                      & {628}                             & {69}                         & {251}                         & {66.24\%}                      \\
		\hline {2013}          & {MCTest-mc160 \cite{richardson2013mctest}}                         & {640}                      & {280}                             & {120}                        & {240}                         & {43.75\%}                      \\
		\hline {2018}          & {ProPara \cite{dalvi-etal-2018-ProPara}}                              & {488}                      & {391}                             & {54}                         & {43}                          & {80.12\%}                      \\
		
	\end{supertabular}
\end{center}

\begin{figure}[H]
	\centering
	\includegraphics[width=15cm]{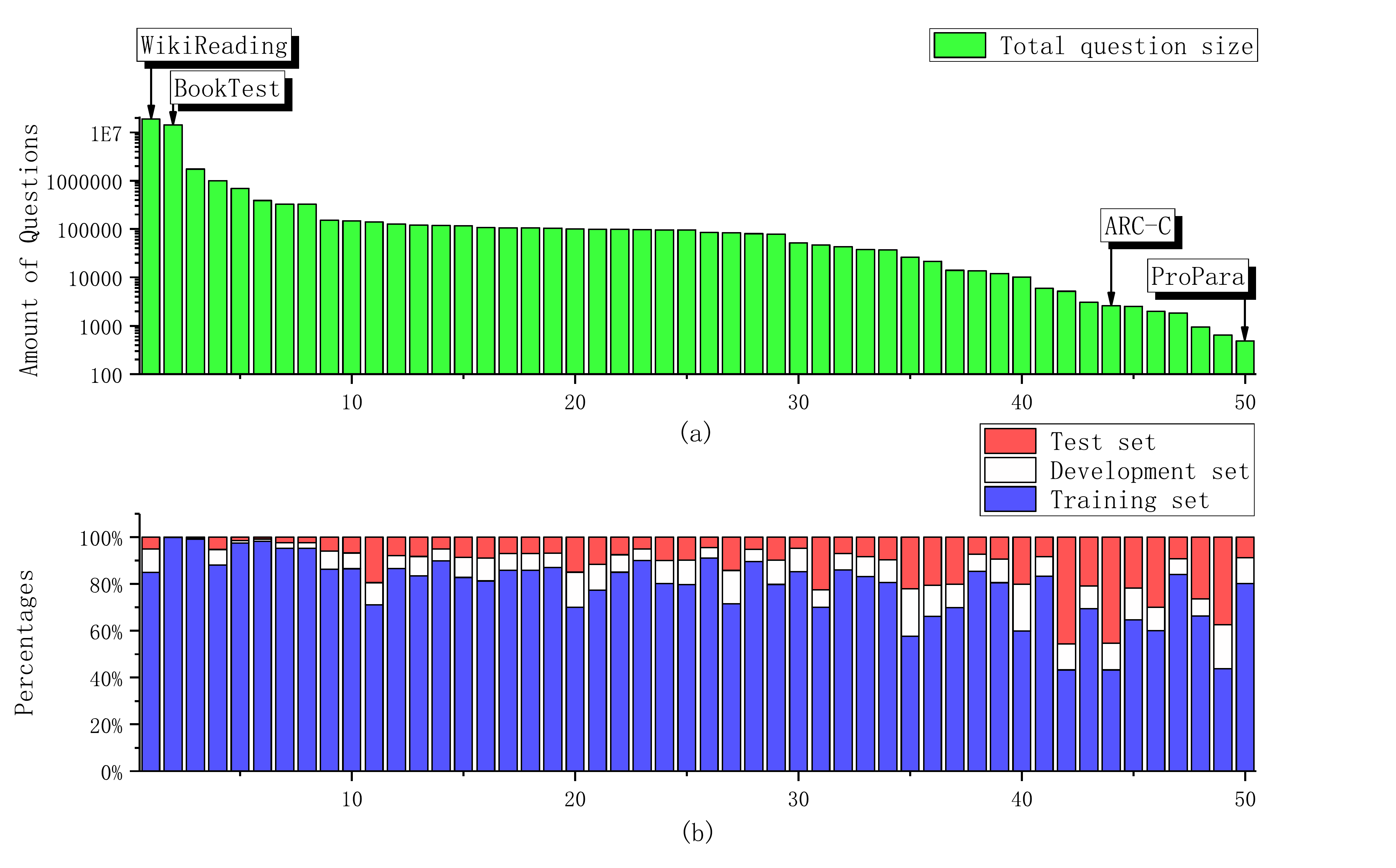}
	\caption{The size of machine Reading Comprehension datasets: (\textbf{a}) Total question size of each dataset. (\textbf{b}) Percentages of training sets, development sets and test sets.}
	\label{figure: question_size }
\end{figure}

We also use the data in Table \ref{table: question_size } to make a statistical chart where the Y coordinate is logarithmic, as shown in Figure \ref{figure: question_size }, we can see that the WikiReading is the dataset with the largest question size \cite{hewlett-etal-2016-wikireading} of a total of 18.87M questions; BookTest \cite{bajgar2016embracing} is ranked second, and ProPara \cite{dalvi-etal-2018-ProPara} is the smallest which has only 488 questions. When it comes to the proportion of training sets, BookTest has the highest proportion, 99.86\%, while the ARC (challenge set) has the lowest proportion which is 43.20\%. The development set is generally slightly smaller than the testing set.

Because different MRC datasets contain different corpora, we also give details of the corpus used in each MRC dataset, including the size of corpus and the unit of corpus, as well as the size of training set, development set, and testing set. As seen in Table \ref{table: corpus_size }, The units of corpus in MRC datasets are various, such as paragraphs, documents, etc.

\begin{center}
	\renewcommand\arraystretch{1.2} 
	\topcaption{The corpus size of machine reading comprehension datasets.}
	\label{table: corpus_size }
	\tablehead{
		\toprule{\textbf{Year}} &{\textbf{Datasets}}   &{\textbf{Corpus size}} &{\textbf{\#Train Corpus}} &{\textbf{\#Dev Corpus}} &{\textbf{\#Test Corpus}} &{\textbf{Unit of Corpus}}\\
	}
	\tabletail{	
		\hline
		\multicolumn{7}{r}{\small\sl continued on next page}\\
		\hline 
	}
	
	\tablefirsthead{
		\toprule{\textbf{Year}} &{\textbf{Datasets}}   &{\textbf{Corpus size}} &{\textbf{\#Train Corpus}} &{\textbf{\#Dev Corpus}} &{\textbf{\#Test Corpus}} &{\textbf{Unit of Corpus}}\\
		\midrule}
	
	\tablelasttail{\bottomrule}
	
	\begin{supertabular}{p{0.8cm} p{2.4cm} p{1.5cm} p{1.5cm} p{1.5cm}  p{1.5cm} p{1.5cm}l}			
		{2016}  &{WikiReading \cite{hewlett-etal-2016-wikireading}}    &{4.7M}   &{N/A}   &{N/A}   &{N/A}   &{Article}  \\
		\hline{2016}  &{SQuAD 1.1 \cite{rajpurkar2016squad}}    &{536}   &{442}   &{48}   &{46}   &{Article}  \\
		\hline{2018}  &{SQuAD 2.0 \cite{rajpurkar-etal-2018-know}}    &{505}   &{442}   &{35}   &{28}   &{Article}  \\
		\hline{2016}  &{BookTest \cite{bajgar2016embracing}}    &{14062}  &{N/A}   &{N/A}   &{N/A}   &{Book}  \\
		\hline{2017}  &{COMICS \cite{iyyer2017COMICS}}    &{3948}   &{N/A}   &{N/A}   &{N/A}   &{Book}  \\
		\hline{2016}  &{Facebook CBT \cite{hill2016cbt}}   &{108}   &{98}   &{5}   &{5}   &{Book}  \\
		\hline{2019}  &{DREAM \cite{sun2019dream}}    &{6444}   &{3869}   &{1288}   &{1287}   &{Dialogue}  \\
		\hline{2016}  &{NewsQA \cite{trischler-etal-2017-newsqa}}    &{1010916}  &{909824}   &{50546}   &{50546}   &{Document}  \\
		\hline{2017}  &{TriviaQA-Web \cite{joshi-etal-2017-triviaqa}}   &{662659}  &{528979}   &{68621}   &{65059}   &{Document}  \\
		\hline{2015}  &{Daily Mail \cite{hermann2015cnn}}    &{219506}  &{196961}   &{12148}   &{10397}   &{Document}  \\
		\hline{2017}  &{TriviaQA-Wiki \cite{joshi-etal-2017-triviaqa}}   &{138538}  &{110648}   &{14229}   &{13661}   &{Document}  \\
		\hline{2018}  &{ReviewQA \cite{grail2018reviewqa}}    &{100000}  &{90000}   &{N/A}   &{10000}   &{Document}  \\
		\hline{2015}  &{CNN \cite{hermann2015cnn}}     &{92579}  &{90266}   &{1220}   &{1093}   &{Document}  \\
		\hline{2017}  &{NarrativeQA \cite{kovcisky2018narrativeqa}}    &{1572}   &{1102}   &{115}   &{355}   &{Document}  \\
		\hline{2017}  &{TQA \cite{kembhavi2017tqa}} &{1076}   &{666}   & {200}   &{210}   &{Lesson}  \\
		\hline{2016}  &{MovieQA \cite{tapaswi2016movieqa}}    &{548}   &{362}   &{77}   &{109}   &{Movie}  \\
		\hline{2016}  &{Google MC-AFP \cite{soricut2016building}}   &{1742618}  &{1727423}   &{7602}   &{7593}   &{Passage}  \\
		\hline{2016}  &{Who-did-What \cite{onishi2016whodidwhat}}    &{147786}  &{127786}   &{10000}   &{10000}   &{Passage}  \\
		\hline{2017}  &{SearchQA \cite{dunn2017searchqa}}    &{140461}  &{99820}   &{13393}   &{27248}   &{Passage}  \\
		\hline{2018}  &{ReCoRD \cite{zhang2018record}}    &{80121}  &{65709}   &{7133}   &{7279}   &{Passage}  \\
		\hline{2017}  &{Quasar-T \cite{Dhingra2017Quasar}}    &{43012}  &{37012}   &{3000}   &{3000}   &{Passage}  \\
		\hline{2017}  &{Quasar-S \cite{Dhingra2017Quasar}}    &{37362}  &{31049}   &{3174}   &{3139}   &{Passage}  \\
		\hline{2017}  &{RACE \cite{lai-etal-2017-race}}     &{27933}  &{25137}   &{1389}   &{1407}   &{Passage}  \\
		\hline{2018}  &{SciTail \cite{khot2018scitail}}   &{27026}  &{23596}   &{1304}   &{2126}   &{Passage}  \\
		\hline{2016}  &{LAMBADA \cite{boleda2016lambada}}    &{12684}  &{2662}   &{4869}   &{5153}   &{Passage}  \\
		\hline{2018}  &{CoQA \cite{reddy2019coqa}}     &{8399}   &{7199}   &{500}   &{700}   &{Passage}  \\
		\hline{2018}  &{CLOTH \cite{xie2018cloth}}    &{7131}   &{5513}   &{805}   &{813}   &{Passage}  \\
		\hline{2017}  &{Qangaroo-W \cite{welbl2018qangaroo}}   &{51318}  &{43738}   &{5129}   &{2451}   &{Passage}  \\
		\hline{2019}  &{DROP \cite{Dua2019DROP}}     &{6735}   &{5565}   &{582}   &{588}   &{Passage}  \\
		\hline{2017}  &{Qangaroo-M \cite{welbl2018qangaroo}}   &{2508}   &{1620}   &{342}   &{546}   &{Passage}  \\
		\hline{2018}  &{RecipeQA \cite{yagcioglu-etal-2018-recipeqa}}    &{19779}  &{15847}   &{1963}   &{1969}   &{Recipe}  \\
		\hline{2015}  &{WikiQA \cite{yang2015wikiqa}}    &{29258}  & {20360}   &{2733}   &{6165}   &{Sentence}  \\
		\hline{2013}  &{MCTest-mc500 \cite{richardson2013mctest}}   &{500}   &{300}   &{50}   &{150}   &{Story}  \\
		\hline{2013}  &{MCTest-mc160 \cite{richardson2013mctest}}   &{160}   &{70}   &{30}   &{60}   &{Story}  \\
		\hline{2018}  &{QuAC \cite{choi-etal-2018-quac}}     &{8845}   &{6843}   &{1000}   &{1002}   &{Unique section} \\
		\hline{2019}  &{ShARC \cite{saeidi2018sharc}}    &{32436}  &{21890}   &{2270}   &{8276}   &{Utterance} \\
		\hline{2019}  &{Natural Questions \cite{kwiatkowski2019natural}}   &{323045}  &{307373}   &{7830}   &{7842}   &{Wikipedia Page} \\

	\end{supertabular}
\end{center}

\subsection{The Generation Method of Datasets}

The generation method of datasets can be roughly described into several categories: Crowdsourcing, Expert, and Automated. "Crowdsourcing" is evolving as a distributed problem-solving and business production model in recent years \cite{yuen2011survey}. An example of crowdsourcing website is Amazon Mechanical Turk. Today, many MRC datasets are posed by the distributed workforce on such crowdsourcing websites. The "Expert" generation method means that question and answer pairs in the dataset are generated by people with professional knowledge in some fields. For example, in the ARC dataset \cite{Clark2018ARC}, there are 7,787 science questions covered by US elementary and middle schools. The "Automated" generation method means that question and answer pairs are automatically generated based on corpus, such as many cloze datasets.

\subsection{The Source of Corpus}
The source of corpus affects the readability and complexity of machine reading comprehension datasets. According to the source of corpus, the MRC datasets can be described as the following types: Exam Text, Wikipedia, News articles, Abstract of Scientific Paper, Crafted story, Technical documents, Text Book, Movie plots, Recipe, Government Websites, Search engine query logs, Hotel Comments, Narrative text, etc.

\subsection{The Type of Context}
The type of context can affect the training method of machine reading comprehension model, which produces many special models, such as the multi-hop reading comprehension, and multi-document reading comprehension. There are many types of context in MRC datasets, including Paragraph, Multi-paragraph, Document, Multi-document, URL, Paragraphs with diagrams or images. As shown in Table \ref{table: generation_method }, we give details of the generation method, corpus source, and context type of each machine's reading comprehension dataset.

\begin{center}
	\renewcommand\arraystretch{1.2} 
	\topcaption{The generation method of datasets, source of corpus and type of context. }
	\label{table: generation_method }
	\tablehead{
		\toprule { \textbf{Year}} & { \textbf{Datasets}}           & { \textbf{Generation Method}} & { \textbf{Source of Corpus}}  & { \textbf{Type of Context}}       \\
	}
	\tabletail{	
		\hline
		\multicolumn{5}{r}{\small\sl continued on next page}\\
		\hline 
	}
	
	\tablefirsthead{
		\toprule { \textbf{Year}} & { \textbf{Datasets}}           & { \textbf{Generation Method}} & { \textbf{Source of Corpus}}  & { \textbf{Type of Context}}       \\
		\midrule}
	
	\tablelasttail{\bottomrule}
	
	\begin{supertabular}{p{0.8cm} p{3cm} p{3cm} p{3cm}  p{3cm} }

		{2013}     &{MCTest-mc160 \cite{richardson2013mctest} }             &{Crowd-sourcing}       &{Factoid  stories}     &{Paragraph}        \\
		\hline{2013}     &{MCTest-mc500 \cite{richardson2013mctest}}             &{Crowd-sourcing}       &{Factoid  stories}     &{Paragraph}        \\
		\hline{2015}     &{CNN \cite{hermann2015cnn}}                &{Automated}         &{News}            &{Document}         \\
		\hline{2015}     &{CuratedTREC \cite{baudivs2015CuratedTREC}}              &{Crowd-sourcing}       &{Factoid  stories}     &{Paragraph}        \\
		\hline{2015}     &{Daily  Mail\cite{hermann2015cnn}}              &{Automated}         &{News}            &{Document}         \\
		\hline{2015}     &{WikiQA \cite{yang2015wikiqa}}                 &{Crowd-sourcing}       &{Wikipedia}         &{Paragraph}        \\
		\hline{2016}     &{BookTest \cite{bajgar2016embracing}}                &{Automated}         &{Factoid  stories}     &{Paragraph}        \\
		\hline{2016}     &{Facebook  CBT \cite{hill2016cbt}}             &{Automated}         &{Factoid  stories}     &{Paragraph}        \\
		\hline{2016}     &{Google  MC-AFP}            &{Automated}         &{The  Gigaword corpus}   &{Paragraph}        \\
		\hline{2016}     &{LAMBADA \cite{boleda2016lambada}}                &{Crowd-sourcing}       &{Book  Corpus}       &{Paragraph}        \\
		\hline{2016}     &{MovieQA \cite{tapaswi2016movieqa}}                &{Crowd-sourcing}       &{Movie}           &{Paragraph  with Images and Videos}  \\
		\hline{2016}     &{MS  MARCO}               &{Automated}         &{The  Bing}         &{Paragraph}        \\
		\hline{2016}     &{NewsQA \cite{trischler-etal-2017-newsqa}}                 &{Crowd-sourcing}       &{News}            &{Document}         \\
		\hline{2016}     &{SQuAD 1.1 \cite{rajpurkar2016squad}}                &{Crowd-sourcing}       &{Wikipedia}         &{Paragraph}        \\
		\hline{2016}     &{Who-did-What \cite{onishi2016whodidwhat}}              &{Automated}         &{News}            &{Document}         \\
		\hline{2016}     &{WikiMovies \cite{miller2016WikiMovies}}               &{Automated}         &{Movie}           &{Document}         \\
		\hline{2016}     &{WikiReading \cite{hewlett-etal-2016-wikireading}}              &{Automated}         &{Wikipedia}         &{Document}         \\
		\hline{2017}     &{COMICS \cite{iyyer2017COMICS}}                 &{Automated}         &{Comics}           &{Paragraph  with Images}       \\
		\hline{2017}     &{NarrativeQA \cite{kovcisky2018narrativeqa}}              &{Crowd-sourcing}       &{Movie}           &{Document}         \\
		\hline{2017}     &{Qangaroo-M \cite{welbl2018qangaroo}}            &{Crowd-sourcing}       &{Wikipedia}         &{Paragraph}        \\
		\hline{2017}     &{Qangaroo-W \cite{welbl2018qangaroo}}            &{Crowd-sourcing}       &{Scientic  paper}      &{Paragraph}        \\
		\hline{2017}     &{Quasar-S \cite{Dhingra2017Quasar}}                &{Crowd-sourcing}       &{Stack  Overflow}      &{Paragraph}        \\
		\hline{2017}     &{Quasar-T \cite{Dhingra2017Quasar}}                &{Crowd-sourcing}       &{Stack  Overflow}      &{Paragraph}        \\
		\hline{2017}     &{RACE \cite{lai-etal-2017-race}}                  &{Expert}           &{English  Exam}       &{Document}         \\
		\hline{2017}     &{SciQ \cite{welbl2017SciQ}}                  &{Crowd-sourcing}       &{School  science curricula} &{Paragraph}        \\
		\hline{2017}     &{SearchQA \cite{dunn2017searchqa}}                &{Crowd-sourcing}       &{J!  Archive and Google}  &{Paragraph  \& URL}          \\
		\hline{2017}     &{TQA \cite{kembhavi2017tqa}}                  &{Expert}           &{School  science curricula} &{Paragraph  with Images}       \\
		\hline{2017}     &{TriviaQA-Wiki \cite{joshi-etal-2017-triviaqa}}           &{Automated}         &{The  Bing}         &{Paragraph}        \\
		\hline{2017}     &{TriviaQA-Web \cite{joshi-etal-2017-triviaqa}}             &{Automated}         &{The  Bing}         &{Paragraph}        \\
		\hline{2018}     &{ARC-Challenge Set \cite{Clark2018ARC}}          &{Expert}           &{School  science curricula} &{Paragraph}        \\
		\hline{2018}     &{ARC-Easy Set \cite{Clark2018ARC}}             &{Expert}           &{School  science curricula} &{Paragraph}        \\
		\hline{2018}     &{CliCR \cite{suster-daelemans-2018-clicr}}                 &{Automated}         &{BMJ  Case Reports}     &{Paragraph}        \\
		\hline{2018}     &{CLOTH \cite{xie2018cloth}}                 &{Expert}           &{English  Exam}       &{Document}         \\
		\hline{2018}     &{CoQA \cite{reddy2019coqa}}                  &{Crowd-sourcing}       &{Jeopardy}          &{Paragraph}        \\
		\hline{2018}     &{DuoRC-Paraphrase \cite{saha-etal-2018-duorc}}            &{Crowd-sourcing}       &{Movie}           &{Paragraph}        \\
		\hline{2018}     &{DuoRC-Self \cite{saha-etal-2018-duorc}}               &{Crowd-sourcing}       &{Movie}           &{Paragraph}        \\
		\hline{2018}     &{HotpotQA-D \cite{yang-etal-2018-hotpotqa}}     &{Crowd-sourcing}       &{Wikipedia}         &{Multi-paragraph}           \\
		\hline{2018}     &{HotpotQA-F \cite{yang-etal-2018-hotpotqa}}      &{Crowd-sourcing}       &{Wikipedia}         &{Multi-paragraph}           \\
		\hline{2018}     &{MCScript \cite{ostermann-etal-2018-mcscript}}                &{Crowd-sourcing}       &{Narrative  texts}     &{Paragraph}        \\
		\hline{2018}     &{MultiRC \cite{khashabi-etal-2018MultiRC}}                &{Crowd-sourcing}       &{News  and other web pages} &{Multi-sentence}      \\
		\hline{2018}     &{OpenBookQA \cite{OpenBookQA2018}}               &{Crowd-sourcing}       &{School  science curricula} &{Paragraph}        \\
		\hline{2018}     &{PaperQA(Hong et al.) \cite{hong2018learning}}    &{Crowd-sourcing}       &{Scientic  paper}      &{Paragraph}        \\
		\hline{2018}     &{PaperQA-L \cite{park2019can}} &{Automated}         &{Scientic  paper}      &{Paragraph}        \\
		\hline{2018}     &{PaperQA-T \cite{park2019can}}     &{Automated}         &{Scientic  paper}      &{Paragraph}        \\
		\hline{2018}     &{ProPara \cite{dalvi-etal-2018-ProPara}}                &{Crowd-sourcing}       &{Process  Paragraph}    &{Paragraph}        \\
		\hline{2018}     &{QuAC \cite{choi-etal-2018-quac}}                  &{Crowd-sourcing}       &{Wikipedia}         &{Document}         \\
		\hline{2018}     &{RecipeQA \cite{yagcioglu-etal-2018-recipeqa}}                &{Automated}         &{Recipes}          &{Paragraph  with Images}       \\
		\hline{2018}     &{ReCoRD \cite{zhang2018record}}                 &{Crowd-sourcing}       &{News}            &{Paragraph}        \\
		\hline{2018}     &{ReviewQA \cite{grail2018reviewqa}}                &{Crowd-sourcing}       &{Hotel  Comments}      &{Paragraph}        \\
		\hline{2018}     &{SciTail \cite{khot2018scitail}}                &{Crowd-sourcing}       &{School  science curricula} &{Paragraph}        \\
		\hline{2018}     &{SQuAD 2.0 \cite{rajpurkar-etal-2018-know}}                &{Crowd-sourcing}       &{Wikipedia}         &{Paragraph}        \\
		\hline{2019}     &{CommonSenseQA \cite{talmor-etal-2019-commonsenseqa}}             &{Crowd-sourcing}       &{Narrative  texts}     &{Paragraph}        \\
		\hline{2019}     &{DREAM \cite{sun2019dream}}                 &{Crowd-sourcing}       &{English  Exam}       &{Dialogues}        \\
		\hline{2019}     &{DROP \cite{Dua2019DROP}}                  &{Crowd-sourcing}       &{Wikipedia}         &{Paragraph}                \\
		\hline{2019}     &{Natural Questions-L \cite{kwiatkowski2019natural}}    &{Crowd-sourcing}       &{Wikipedia}         &{Paragraph}                \\
		\hline{2019}     &{Natural Questions-S \cite{kwiatkowski2019natural}}    &{Crowd-sourcing}       &{Wikipedia}         &{Paragraph}                \\
		\hline{2019}     &{ShARC \cite{saeidi2018sharc}}                 &{Crowd-sourcing}       &{Government  Websites}   &{Paragraph}            \\   
		
	\end{supertabular}
\end{center}

\subsection{The Availability of Datasets, Leaderboards and Baselines}
The release of MRC baseline projects and leaderboards can help the researchers evaluate the performance of their models. In this section, we try to find all the MRC dataset download links, leaderboards, and baseline projects. As shown in Table \ref{table: availability }, all the download links of MRC datasets are available except PaperQA \cite{park2019paperqa}. Most of the datasets provide leaderboards and baseline projects except only 19.3\% of the datasets. We have published all the download links, leaderboards, and the baseline projects on our \href{https://mrc-datasets.github.io/}{website}.

\begin{center}
	\topcaption{The availability of datasets, leaderboards and baselines. }
	\label{table: availability }
	\tablehead{
		\toprule {\textbf{Year}} & {\textbf{Datasets}}                    & {\textbf{Dataset Availability}} & {\textbf{Leaderboard Availability}} & {\textbf{Baseline Availability}}   \\
	}
	\tabletail{	
		\hline
		\multicolumn{5}{r}{\small\sl continued on next page}\\
		\hline 
	}
	
	\tablefirsthead{
		\toprule {\textbf{Year}} & {\textbf{Datasets}}                    & {\textbf{Dataset Availability}} & {\textbf{Leaderboard Availability}} & {\textbf{Baseline Availability}}   \\
		\midrule}
	
	\tablelasttail{\bottomrule}
	
	\begin{supertabular}{p{1cm} p{3.5cm} p{2.5cm} p{2.5cm}  p{2.5cm} }	
		
		{2019}          & {CommonSenseQA \cite{talmor-etal-2019-commonsenseqa}}                        & {$\surd$}                          & {$\surd$}                              & {$\surd$}                 \\
		\hline {2018}          & {MCScript \cite{ostermann-etal-2018-mcscript}}                             & {$\surd$}                          & {$\surd$}                              & {$\times$}                         \\
		\hline {2018}          & {OpenBookQA \cite{OpenBookQA2018}}                           & {$\surd$}                          & {$\surd$}                              & {$\times$}                         \\
		\hline {2018}          & {ReCoRD \cite{zhang2018record}}                               & {$\surd$}                          & {$\surd$}                              & {$\times$}                         \\
		\hline {2018}          & {ARC-Challenge Set \cite{Clark2018ARC}}                    & {$\surd$}                          & {$\surd$}                              & {$\surd$}                         \\
		\hline {2018}          & {ARC-Easy Set \cite{Clark2018ARC}}                         & {$\surd$}                          & {$\surd$}                              & {$\surd$}                         \\
		\hline {2018}          & {CLOTH \cite{xie2018cloth}}                                & {$\surd$}                          & {$\surd$}                              & {$\surd$}                         \\
		\hline {2016}          & {Facebook CBT \cite{hill2016cbt}}                         & {$\surd$}                          & {$\times$}                              & {$\surd$}                         \\
		\hline {2016}          & {NewsQA \cite{trischler-etal-2017-newsqa}}                               & {$\surd$}                          & {$\times$}                              & {$\times$}                         \\
		\hline {2018}          & {ProPara \cite{dalvi-etal-2018-ProPara}}                              & {$\surd$}                          & {$\surd$}                              & {$\times$}                         \\
		\hline {2017}          & {RACE \cite{lai-etal-2017-race}}                                 & {$\surd$}                          & {$\surd$}                              & {$\surd$}                         \\
		\hline {2016}          & {SQuAD 1.1 \cite{rajpurkar2016squad}}                             & {$\surd$}                          & {$\surd$}                              & {$\surd$}                         \\
		\hline {2017}          & {TriviaQA-Wiki \cite{joshi-etal-2017-triviaqa}}                       & {$\surd$}                          & {$\surd$}                              & {$\surd$}                         \\
		\hline {2017}          & {TriviaQA-Web \cite{joshi-etal-2017-triviaqa}}                        & {$\surd$}                          & {$\surd$}                              & {$\surd$}                         \\
		\hline {2019}          & {DROP \cite{Dua2019DROP}}                                 & {$\surd$}                          & {$\surd$}                              & {$\surd$}                         \\
		\hline {2017}          & {NarrativeQA \cite{kovcisky2018narrativeqa}}                          & {$\surd$}                          & {$\times$}                              & {$\surd$}                         \\
		\hline {2019}          & {ShARC \cite{saeidi2018sharc}}                                & {$\surd$}                          & {$\surd$}                              & {$\times$}                         \\
		\hline {2018}          & {CoQA \cite{reddy2019coqa}}                                 & {$\surd$}                          & {$\surd$}                              & {$\surd$}                         \\
		\hline {2019}          & {DREAM \cite{sun2019dream}}                                & {$\surd$}                          & {$\surd$}                              & {$\surd$}                         \\
		\hline {2018}          & {QuAC \cite{choi-etal-2018-quac}}                                 & {$\surd$}                          & {$\surd$}                              & {$\surd$}                         \\
		\hline {2013}          & {MCTest-mc160 \cite{richardson2013mctest}}                         & {$\surd$}                          & {$\surd$}                              & {$\surd$}                         \\
		\hline {2013}          & {MCTest-mc500 \cite{richardson2013mctest}}                         & {$\surd$}                          & {$\surd$}                              & {$\surd$}                         \\
		\hline {2015}          & {WikiQA \cite{yang2015wikiqa}}                               & {$\surd$}                          & {$\times$}                              & {$\times$}                         \\
		\hline {2018}          & {CliCR \cite{suster-daelemans-2018-clicr}}                                & {$\surd$}                          & {$\times$}                              & {$\surd$}                         \\
		\hline {2018}          & {PaperQA(Hong et al.)\cite{hong2018learning}}             & {$\surd$}                          & {$\times$}                              & {$\times$}                         \\
		\hline {2018}          & {PaperQA-L \cite{park2019can}}    & {$\times$}                          & {$\times$}                              & {$\times$}                         \\
		\hline {2018}          & {PaperQA-T \cite{park2019can}}              & {$\times$}                          & {$\times$}                              & {$\times$}                         \\
		\hline {2018}          & {ReviewQA \cite{grail2018reviewqa}}                             & {$\surd$}                          & {$\times$}                              & {$\times$}                         \\
		\hline {2017}          & {SciQ \cite{welbl2017SciQ}}                                 & {$\surd$}                          & {$\times$}                              & {$\times$}                         \\
		\hline {2016}          & {WikiMovies \cite{miller2016WikiMovies}}                           & {$\surd$}                          & {$\times$}                              & {$\surd$}                         \\
		\hline {2016}          & {BookTest \cite{bajgar2016embracing}}                             & {$\surd$}                          & {$\times$}                              & {$\times$}                         \\
		\hline {2015}          & {CNN \cite{hermann2015cnn}}                                  & {$\surd$}                          & {$\times$}                              & {$\surd$}                         \\
		\hline {2015}          & {Daily Mail\cite{hermann2015cnn}}                           & {$\surd$}                          & {$\times$}                              & {$\surd$}                         \\
		\hline {2016}          & {Who-did-What \cite{onishi2016whodidwhat}}                         & {$\surd$}                          & {$\surd$}                              & {$\surd$}                         \\
		\hline {2016}          & {WikiReading \cite{hewlett-etal-2016-wikireading}}                          & {$\surd$}                          & {$\times$}                              & {$\surd$}                         \\
		\hline {2016}          & {Google MC-AFP \cite{soricut2016building}}                        & {$\surd$}                          & {$\times$}                              & {$\times$}                         \\
		\hline {2016}          & {LAMBADA \cite{boleda2016lambada}}                              & {$\surd$}                          & {$\times$}                              & {$\surd$}                         \\
		\hline {2018}          & {SciTail \cite{khot2018scitail}}                              & {$\surd$}                          & {$\surd$}                              & {$\times$}                         \\
		\hline {2018}          & {DuoRC-Paraphrase \cite{saha-etal-2018-duorc}}                     & {$\surd$}                          & {$\surd$}                              & {$\surd$}                         \\
		\hline {2018}          & {DuoRC-Self \cite{saha-etal-2018-duorc}}                           & {$\surd$}                          & {$\surd$}                              & {$\surd$}                         \\
		\hline {2015}          & {CuratedTREC \cite{baudivs2015CuratedTREC}}                          & {$\surd$}                          & {$\surd$}                              & {$\surd$}                         \\
		\hline {2017}          & {Quasar-S \cite{Dhingra2017Quasar}}                             & {$\surd$}                          & {$\times$}                              & {$\surd$}                         \\
		\hline {2017}          & {Quasar-T \cite{Dhingra2017Quasar}}                             & {$\surd$}                          & {$\times$}                              & {$\surd$}                         \\
		\hline {2017}          & {SearchQA \cite{dunn2017searchqa}}                             & {$\surd$}                          & {$\times$}                              & {$\times$}                         \\
		\hline {2019}          & {Natural Questions-L \cite{kwiatkowski2019natural}}        & {$\surd$}                          & {$\surd$}                              & {$\surd$}                         \\
		\hline {2019}          & {Natural Questions-S \cite{kwiatkowski2019natural}}       & {$\surd$}                          & {$\surd$}                              & {$\surd$}                         \\
		\hline {2018}          & {SQuAD 2.0 \cite{rajpurkar-etal-2018-know}}                             & {$\surd$}                          & {$\surd$}                              & {$\surd$}                         \\
		\hline {2016}          & {MS MARCO \cite{tri2016MARCO}}                             & {$\surd$}                          & {$\surd$}                              & {$\surd$}                         \\
		\hline {2017}          & {Qangaroo-MEDHOP \cite{welbl2018qangaroo}}                      & {$\surd$}                          & {$\surd$}                              & {$\times$}                         \\
		\hline {2017}          & {Qangaroo-WIKIHOP \cite{welbl2018qangaroo}}                     & {$\surd$}                          & {$\surd$}                              & {$\times$}                         \\
		\hline {2018}          & {MultiRC \cite{khashabi-etal-2018MultiRC}}                              & {$\surd$}                          & {$\surd$}                              & {$\surd$}                         \\
		\hline {2018}          & {HotpotQA-Distractor \cite{yang-etal-2018-hotpotqa}}         & {$\surd$}                          & {$\surd$}                              & {$\surd$}                         \\
		\hline {2018}          & {HotpotQA-Fullwiki \cite{yang-etal-2018-hotpotqa}}           & {$\surd$}                          & {$\surd$}                              & {$\surd$}                         \\
		\hline {2017}          & {COMICS \cite{iyyer2017COMICS}}                               & {$\surd$}                          & {$\times$}                              & {$\surd$}                                    \\
		\hline {2016}          & {MovieQA \cite{tapaswi2016movieqa}}                              & {$\surd$}                          & {$\surd$}                              & {$\surd$}                                   \\
		\hline {2018}          & {RecipeQA \cite{yagcioglu-etal-2018-recipeqa}}                             & {$\surd$}                          & {$\surd$}                              & {$\times$}                                      \\
		\hline {2017}          & {TQA \cite{kembhavi2017tqa}} & {$\surd$}                          & {$\surd$}                              & {$\times$}                       \\

	\end{supertabular}
\end{center}

\subsection{Statistical Analysis}
\begin{figure}[H]
	\centering
		\includegraphics[width=15cm]{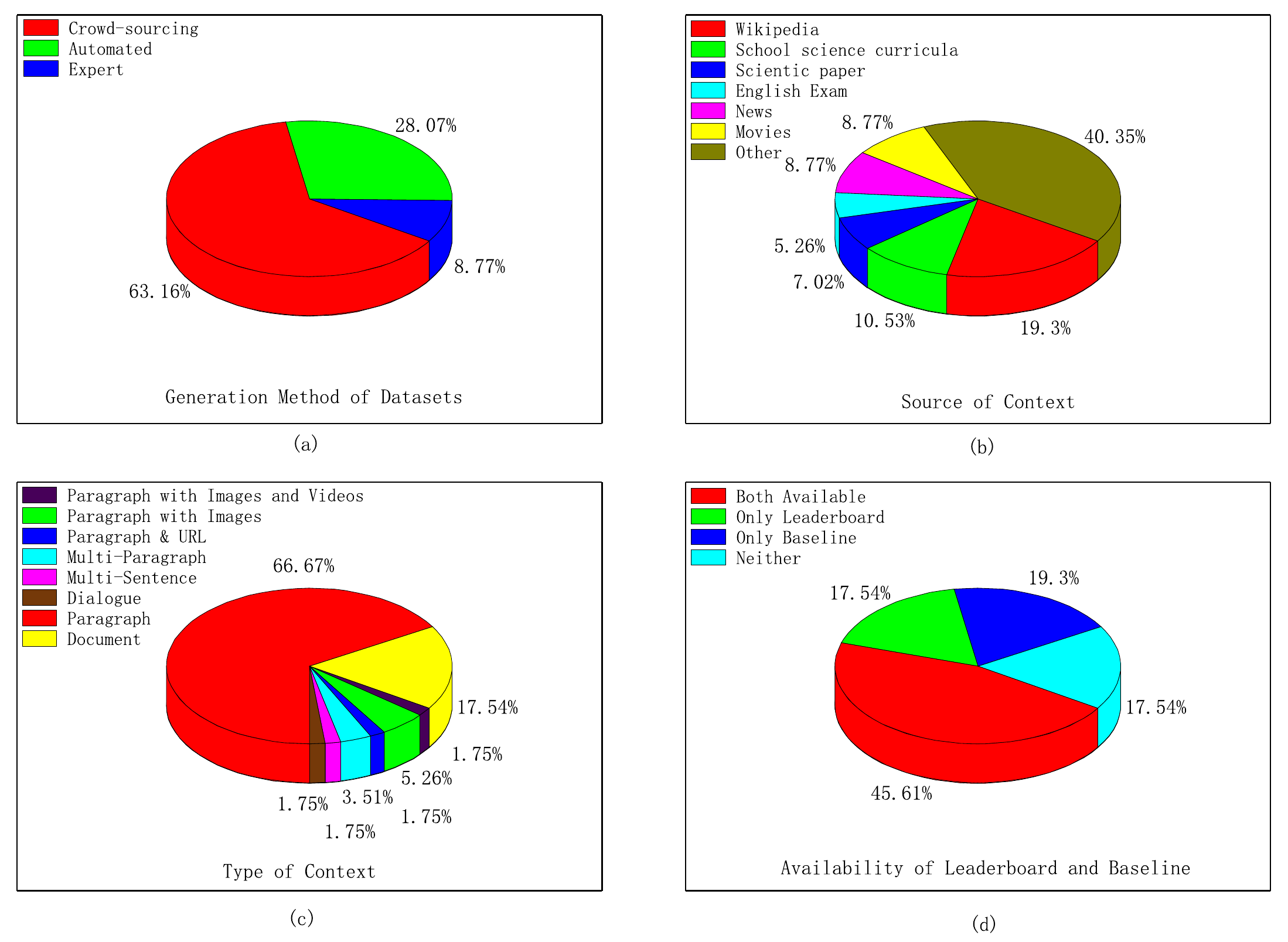}
	\caption{Statistical analysis of the datasets: (\textbf{a}) The generation method of datasets. (\textbf{b}) The source of corpus.(\textbf{c}) The type of context.(\textbf{d}) The availability of leaderboards and baselines.}
	\label{table: Statistical_of_datasets }
	
\end{figure}

Figure \ref{table: Statistical_of_datasets } demonstrates the statistical analysis of the attributes of datasets in Table \ref{table: generation_method }. As seen in Figure \ref{table: Statistical_of_datasets } (a), the most common way to generate datasets is "Crowdsourcing", by which we can generate question and answer pairs that need complex reasoning abilities. The second is the "Automated" method which can help us quickly create large-scale MRC datasets. The "Expert" generate method is the least used because it is usually expensive. When it comes to context type, as seen in Figure \ref{table: Statistical_of_datasets } (b), the main context type is the "Paragraph" type, followed by "Document" type, "Paragraph with images", "Multi-Paragraph" and so on. Figure \ref{table: Statistical_of_datasets } (c) shows the source of corpus which is very diverse. Among them, "Wikipedia" is the most common context source, but only accounts for 19.30\%. Figure \ref{table: Statistical_of_datasets } (d) illustrated the availability of leaderboard and baseline. As can be seen in Figure \ref{table: Statistical_of_datasets } (d), 45.61\% of the datasets provide both leaderboards and baseline project, only 19.3\% of the datasets neither provide leaderboards nor baseline projects. For the availability of dataset,all the download links of MRC datasets are available except PaperQA \cite{park2019paperqa}.

\subsection{Prerequisite Skills}
When humans read passages and answer questions, we need to master various prerequisite skills to answer them correctly. The analysis of these prerequisite skills may help us understand the intrinsic properties of the MRC datasets. In Table \ref{table: prerequisite_skills }, we quote the descriptions and examples of prerequisite skills proposed by Sugawara et al.\cite{sugawara2016analysis}. They defined 10 kinds of prerequisite skills, including List/Enumeration, Mathematical operations, Coreference resolution, Logical reasoning, etc. By manually annotate questions in the MCTest \cite{richardson2013mctest} and SQuAD 1.1 \cite{rajpurkar2016squad}, they got the frequencies of each prerequisite skill in the two MRC datasets. As seen in Table \ref{table: prerequisite_skills }.
However, the definition and classification of these prerequisite skills are often subjective and changeable. Many definitions have been drawn \cite{sugawara2016analysis,sugawara2017evaluation,sugawara2019assessing} , but they are still hard to give a standard mathematical definition of them, which is the same as natural language understanding.

\begin{center}
	\renewcommand\arraystretch{1.2} 
	\topcaption{Prerequisite skills with descriptions or examples \cite{sugawara2016analysis}, and their frequencies (in percentage) in SQuAD 1.1 \cite{rajpurkar2016squad} and MCTest \cite{richardson2013mctest} (MC160 development set). }
	\label{table: prerequisite_skills }
	\tablehead{
		\toprule { \textbf{Prerequisite skills}} & { \textbf{Descriptions or examples}}         & { \textbf{Frequency SQuAD}} & { \textbf{Frequency MCTest}}   \\
	}
	\tabletail{	
		\hline
		\multicolumn{4}{r}{\small\sl continued on next page}\\
		\hline 
	}
	
	\tablefirsthead{
		\toprule { \textbf{Prerequisite skills}} & { \textbf{Descriptions or examples}}         & { \textbf{Frequency SQuAD}} & { \textbf{Frequency MCTest}}   \\
		\hline}
	
	\tablelasttail{\bottomrule}
	
	\begin{supertabular}{p{4cm} p{6cm}p{1.5cm}  p{1.5cm} }	
		
		{ List/Enumeration}      & { Tracking, retaining, and list/enumeration of entities or states}   & { 5.00\%}       & { 11.70\%}       \\
		\hline { Mathematical operations}     & { Four basic operations and geometric comprehension}     & { 0.00\%}       & { 4.20\%}      \\
		\hline { Coreference resolution}      & { Detection and resolution of coreferences}        & { 6.20\%}       & { 57.50\%}       \\
		\hline { Logical reasoning}       & { Induction, deduction, conditional statement, and quantifier}   & { 1.20\%}       & { 0.00\%}      \\
		\hline { Analogy}         & { Trope in figures of speech, e.g.,metaphor}       & { 0.00\%}       & { 0.00\%}      \\
		\hline { Spatiotemporal relations}    & { Spatial and/or temporal relations of events}       & { 2.50\%}       & { 28.30\%}       \\
		\hline { Causal relations}      & { Why, because, the reason, etc.}          & { 6.20\%}       & { 18.30\%}       \\
		\hline { Commonsense reasoning}       & { Taxonomic/qualitative knowledge, action and event change}    & { 86.20\%}      & { 49.20\%}       \\
		\hline { Complex sentences}       & { Coordination or subordination of clauses}        & { 20.00\%}      & { 15.80\%}       \\
		\hline { Special sentence structure}    & { Scheme in figures of speech, constructions, and punctuation marks} & { 25.00\%}      & { 10.00\%} \\
		
	\end{supertabular}
\end{center}

\subsection{Citation Analysis}
The number of citations of the paper in which a dataset was proposed reveals the dataset’s impact to some extent. As shown in Table \ref{table: Citation }, we analyze how many times each paper was cited and make a statistical table. We count both the total number of citations and the monthly average citations since they were published. Except for the two PaperQA datasets \cite{hong2018learning,park2019paperqa}, the number of citations of all other papers have been found in Google Scholar. Besides, we make a Table \ref{table: Citation } in which the datasets are sorted by the monthly average citations. As expected, the dataset with the highest monthly average citations is SQuAD 1.1 \cite{rajpurkar2016squad}, followed by CNN/Daily Mail \cite{hermann2015cnn} and SQuAD 2.0 \cite{rajpurkar-etal-2018-know}. It shows that these datasets are widely used as a benchmark.

\begin{center}
	\renewcommand\arraystretch{1.2} 
	\topcaption{Citation analysis of the paper in which each dataset was proposed.}
	\label{table: Citation }
	\tablehead{
		\toprule {\textbf{Year}} & {\textbf{Datasets}} & {\textbf{Average Monthly Citations}} & {\textbf{Total Citations}} & {\textbf{Months after Publication}} & {\textbf{Date of Publication}} & {\textbf{Date of Statistics}} \\
	}
	\tabletail{	
		\hline
		\multicolumn{7}{r}{\small\sl continued on next page}\\
		\hline 
	}
	
	\tablefirsthead{
		\toprule {\textbf{Year}} & {\textbf{Datasets}} & {\textbf{Average Monthly Citations}} & {\textbf{Total Citations}} & {\textbf{Months after Publication}} & {\textbf{Date of Publication}} & {\textbf{Date of Statistics}} \\
		\midrule}
	
	\tablelasttail{\bottomrule}
	
	\begin{supertabular}{p{1cm} p{2.5cm} p{1.5cm} p{1.5cm} p{1.5cm} p{2cm} p{2cm}}	
		
             {2016}     & {SQuAD 1.1 \cite{rajpurkar2016squad}}     & {33.35}                & {1234}           & {37}                & {2016-10-10}           & {2019-12-01}  \\
  \hline {2015}     & {CNN/Daily Mail \cite{hermann2015cnn}}  & {25.21}                & {1210}           & {48}                & {2015-11-19}           & {2019-12-01}  \\
  \hline {2018}     & {SQuAD 2.0 \cite{rajpurkar-etal-2018-know}}     & {14.65}                & {249}            & {17}                & {2018-06-11}           & {2019-12-01}  \\
  \hline {2019}     & {Natural Questions \cite{kwiatkowski2019natural}} & {9.00}                 & {45}            & {5}                 & {2019-07-01}           & {2019-12-01}  \\
  \hline {2017}     & {TriviaQA \cite{joshi-etal-2017-triviaqa}}     & {7.97}                 & {239}            & {30}                & {2017-05-13}           & {2019-12-01}  \\
  \hline {2018}     & {CoQA \cite{reddy2019coqa}}       & {7.93}                 & {119}            & {15}                & {2018-08-21}           & {2019-12-01}  \\
  \hline {2016}     & {WikiMovies \cite{miller2016WikiMovies}}    & {7.73}                 & {286}            & {37}                & {2016-10-10}           & {2019-12-01}  \\
  \hline {2016}     & {CBT \cite{hill2016cbt}}   & {6.92}                 & {332}            & {48}                & {2015-11-07}           & {2019-12-01}  \\
  \hline {2016}     & {MS MARCO \cite{tri2016MARCO}}     & {6.65}                 & {246}            & {37}                & {2016-10-31}           & {2019-12-01}  \\
  \hline {2015}     & {WikiQA \cite{yang2015wikiqa}}      & {6.43}                 & {328}            & {51}                & {2015-09-01}           & {2019-12-01}  \\
  \hline {2018}     & {HotpotQA \cite{yang-etal-2018-hotpotqa}}     & {5.71}                 & {80}            & {14}                & {2018-09-25}           & {2019-12-01}  \\
  \hline {2016}     & {NewsQA \cite{trischler-etal-2017-newsqa}}      & {5.21}                 & {172}            & {33}                & {2017-02-07}           & {2019-12-01}  \\
  \hline {2016}     & {MovieQA \cite{tapaswi2016movieqa}}      & {5.00}                 & {235}            & {47}                & {2015-12-09}           & {2019-12-01}  \\
  \hline {2017}     & {RACE \cite{lai-etal-2017-race}}       & {4.87}                 & {151}            & {31}                & {2017-04-15}           & {2019-12-01}  \\
  \hline {2018}     & {QuAC \cite{choi-etal-2018-quac}}       & {4.73}                 & {71}            & {15}                & {2018-08-27}           & {2019-12-01}  \\
  \hline {2013}     & {MCTest \cite{richardson2013mctest}}      & {4.69}                 & {347}            & {74}                & {2013-10-01}           & {2019-12-01}  \\
  \hline {2017}     & {Qangaroo \cite{welbl2018qangaroo}}     & {4.59}                 & {78}            & {17}                & {2018-06-11}           & {2019-12-01}  \\
  \hline {2018}     & {SciTail \cite{khot2018scitail}}  & {4.16}                 & {79}            & {19}                & {2018-04-27}           & {2019-12-01}  \\
  \hline {2017}     & {NarrativeQA \cite{kovcisky2018narrativeqa}}    & {3.74}                 & {86}            & {23}                & {2017-12-19}           & {2019-12-01}  \\
  \hline {2019}     & {DROP \cite{Dua2019DROP}}       & {3.00}                 & {27}            & {9}                 & {2019-03-01}           & {2019-12-01}  \\
  \hline {2018}     & {ARC}        & {2.90}                 & {58}            & {20}                & {2018-03-14}           & {2019-12-01}  \\
  \hline {2017}     & {SearchQA \cite{dunn2017searchqa}}     & {2.81}                 & {87}            & {31}                & {2017-04-18}           & {2019-12-01}  \\
  \hline {2018}     & {OpenBookQA \cite{OpenBookQA2018}}    & {2.64}                 & {37}            & {14}                & {2018-09-08}           & {2019-12-01}  \\
  \hline {2016}     & {WikiReading \cite{hewlett-etal-2016-wikireading}}    & {2.41}                 & {77}            & {32}                & {2017-03-15}           & {2019-12-01}  \\
  \hline {2019}     & {CommonSenseQA \cite{talmor-etal-2019-commonsenseqa}}   & {2.33}                 & {28}            & {12}                & {2018-11-02}           & {2019-12-01}  \\
  \hline {2017}     & {Quasar \cite{Dhingra2017Quasar}}      & {1.82}                 & {51}            & {28}                & {2017-07-12}           & {2019-12-01}  \\
  \hline {2016}     & {Who-did-What \cite{onishi2016whodidwhat}}   & {1.69}                 & {66}            & {39}                & {2016-08-18}           & {2019-12-01}  \\
  \hline {2018}     & {MultiRC \cite{khashabi-etal-2018MultiRC}}      & {1.67}                 & {30}            & {18}                & {2018-06-01}           & {2019-12-01}  \\
  \hline {2017}     & {TQA \cite{kembhavi2017tqa}}        & {1.55}                 & {45}            & {29}                & {2017-07-01}           & {2019-12-01}  \\
  \hline {2019}     & {DREAM \cite{sun2019dream}}       & {1.50}                 & {15}            & {10}                & {2019-01-31}           & {2019-12-01}  \\
  \hline {2018}     & {ReCoRD \cite{zhang2018record}}      & {1.39}                 & {18}            & {13}                & {2018-10-30}           & {2019-12-01}  \\
  \hline {2016}     & {LAMBADA \cite{boleda2016lambada}}      & {1.29}                 & {53}            & {41}                & {2016-6-20}           & {2019-12-01}  \\
  \hline {2019}     & {ShARC \cite{saeidi2018sharc}}       & {1.27}                 & {19}            & {15}                & {2018-08-28}           & {2019-12-01}  \\
  \hline {2018}     & {MCScript \cite{ostermann-etal-2018-mcscript}}     & {1.10}                 & {22}            & {20}                & {2018-03-14}           & {2019-12-01}  \\
  \hline {2015}     & {CuratedTREC \cite{baudivs2015CuratedTREC}}    & {0.98}                 & {47}            & {48}                & {2015-11-20}           & {2019-12-01}  \\
  \hline {2018}     & {RecipeQA \cite{yagcioglu-etal-2018-recipeqa}}     & {0.93}                 & {13}            & {14}                & {2018-09-04}           & {2019-12-01}  \\
  \hline {2017}     & {COMICS \cite{iyyer2017COMICS}}      & {0.86}                 & {31}            & {36}                & {2016-11-16}           & {2019-12-01}  \\
  \hline {2018}     & {ProPara \cite{dalvi-etal-2018-ProPara}}      & {0.83}                 & {15}            & {18}                & {2018-05-17}           & {2019-12-01}  \\
  \hline {2017}     & {SciQ \cite{welbl2017SciQ}}   & {0.79}                 & {22}            & {28}                & {2017-07-19}           & {2019-12-01}  \\
  \hline {2016}     & {BookTest \cite{bajgar2016embracing}}     & {0.73}                 & {27}            & {37}                & {2016-10-04}           & {2019-12-01}  \\
  \hline {2018}     & {DuoRC \cite{saha-etal-2018-duorc}}       & {0.63}                 & {12}            & {19}                & {2018-04-21}           & {2019-12-01}  \\
  \hline {2018}     & {CliCR \cite{suster-daelemans-2018-clicr}}       & {0.55}                 & {11}            & {20}                & {2018-03-26}           & {2019-12-01}  \\
  \hline {2018}     & {CLOTH \cite{xie2018cloth}}       & {0.42}                 & {10}            & {24}                & {2017-11-09}           & {2019-12-01}  \\
  \hline {2018}     & {ReviewQA \cite{grail2018reviewqa}}     & {0.08}                 & {1}             & {13}                & {2018-10-29}           & {2019-12-01}    \\   
		
	\end{supertabular}
\end{center}

\begin{figure}[H]
	\centering
	\includegraphics[width=12cm]{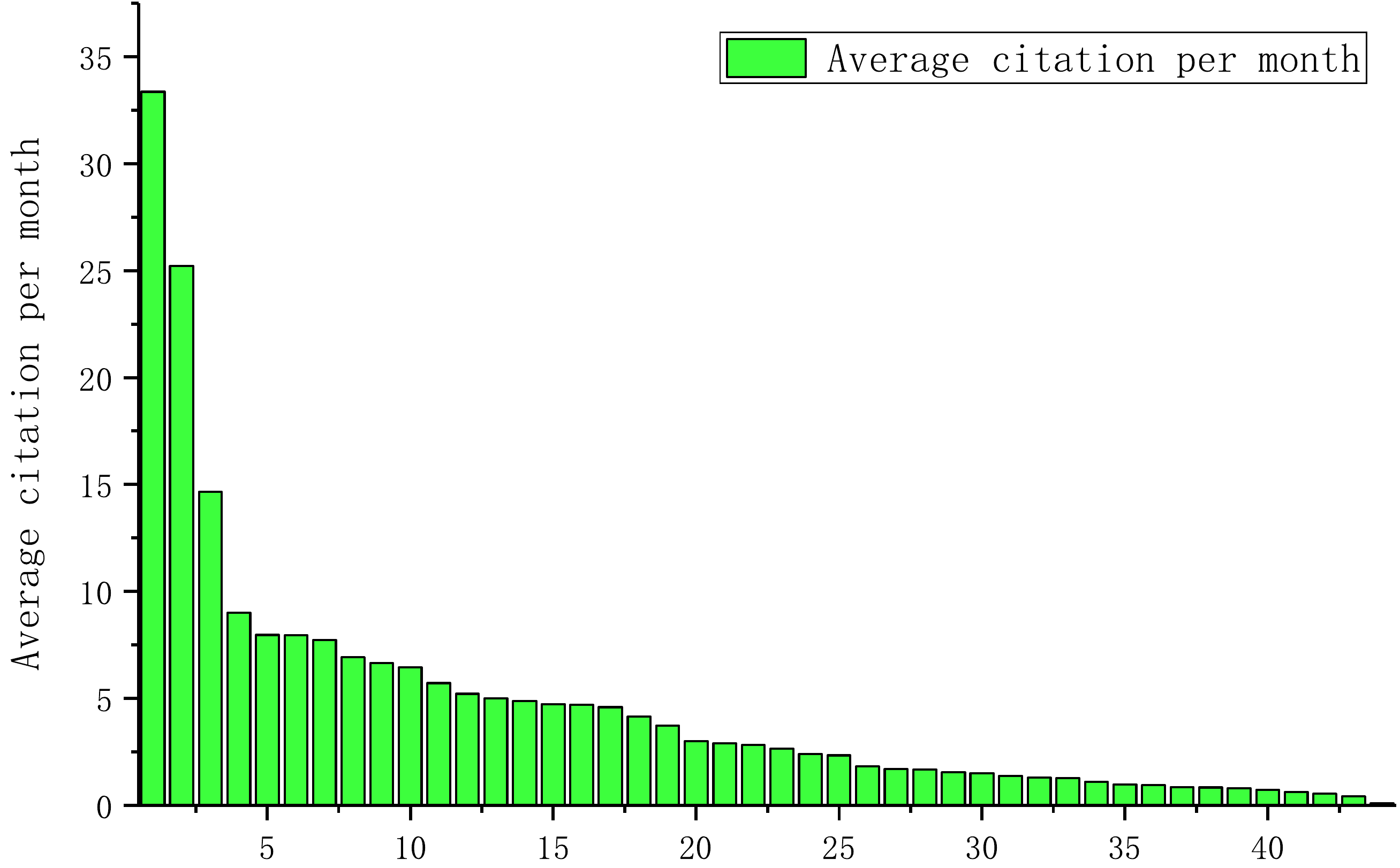}
	\caption{The average number of citations per month of the papers presenting the MRC datasets.}
	\label{figure: Citation }
\end{figure}

We also analyze the monthly average citations. As seen in Figure \ref{figure: Citation }, on the whole, there is a correlation between the monthly average citations and the total citations of the MRC dataset. For example, the top two citations of the total citations and the monthly average citations are the same which are SQuAD 1.1 \cite{rajpurkar2016squad} and CNN/Daily Mail \cite{hermann2015cnn}. However, some papers with lower total citations have higher monthly citations. This shows that these papers have been published for a short time, but they have received a lot of attention from the community, such as SQuAD 2.0 \cite{rajpurkar-etal-2018-know}. In addition, some papers with higher total citations have relatively low monthly average citations. Because these datasets have been published for a long time, but are rarely used in recent years. 

\subsection{Characteristics of Datasets}
\label{sec:Characteristics}
\subsubsection{Overview}
In recent years, various large-scale MRC datasets have been created. The growth of large-scale datasets greatly promoted the research process of the machine reading comprehension.\\
In this section, we analyze the characteristics of existing MRC datasets, including MRC with unanswerable questions, multi-hop MRC, MRC with paraphrased paragraph, MRC which require commonsense (world knowledge), complex reasoning MRC, large-scale dataset, domain-specific dataset, multi-modal MRC, MRC dataset for open-domain QA, and conversational MRC.\\
It should be noted that many MRC datasets have multiple characteristics. A typical example is the DuoRC \cite{saha-etal-2018-duorc} dataset, which has the following four characteristics: 1. DuoRC contains two versions of context, and the meanings of different versions of context are the same while the authors try to avoid words overlap between the two versions, so the DuoRC is a MRC dataset with paraphrased paragraphs. 2. DuoRC requires the use of commonsense and world knowledge. 3. It requires complex reasoning across multiple sentences to infer the answer. 4.There are unanswerable questions in DuoRC \cite{saha-etal-2018-duorc}.\\
Finally, we summarize the characteristics of each dataset in Table \ref{table: characteristics}. In the following sections, we will describe each of them separately.

\begin{center}
	\renewcommand\arraystretch{1.2} 
	\topcaption{The characteristics of each MRC dataset.}
	\label{table: characteristics}
	\tablehead{
		\toprule { \textbf{Year}} & {\textbf{Datasets}}        & {\textbf{Characteristics}} \\
	}
	\tabletail{	
		\hline
		\multicolumn{3}{r}{\small\sl continued on next page}\\
		\hline 
	}
	
	\tablefirsthead{
		\toprule { \textbf{Year}} & {\textbf{Datasets}}        & {\textbf{Characteristics}} \\
		\midrule}
	
	\tablelasttail{\bottomrule}
	
	\begin{supertabular}{p{1cm} p{3cm} p{8cm}}	
		
		{2015}     & {WikiQA \cite{yang2015wikiqa}}             & {With  Unanswerable Questions}      \\
		\hline {2018}     & {SQuAD 2.0 \cite{rajpurkar-etal-2018-know}}            & {With  Unanswerable Questions}      \\
		\hline {2019}     & {Natural Question \cite{kwiatkowski2019natural}}       & {With  Unanswerable Questions}      \\
		\hline {2016}     & {MS  MARCO \cite{tri2016MARCO}}           & {With  Unanswerable Questions ; Multi-hop MRC}       \\
		\hline {2018}     & {DuoRC \cite{saha-etal-2018-duorc}}              & {With  Paraphrased Paragraph ; Require Commonsense (World knowledge) ; Complex  Reasoning ; With Unanswerable Questions} \\
		\hline {2016}     & {Who-did-What \cite{onishi2016whodidwhat}}          & {With  Paraphrased Paragraph ; Complex Reasoning}     \\
		\hline {2018}     & {ARC \cite{Clark2018ARC}}               & {Require  Commonsense (World knowledge) ; Complex Reasoning}       \\
		\hline {2018}     & {MCScript \cite{ostermann-etal-2018-mcscript}}            & {Require  Commonsense (World knowledge)}          \\
		\hline {2018}     & {OpenBookQA \cite{OpenBookQA2018}}           & {Require  Commonsense (World knowledge)}          \\
		\hline {2018}     & {ReCoRD \cite{zhang2018record}}             & {Require  Commonsense (World knowledge)}          \\
		\hline {2019}     & {CommonSenseQA \cite{talmor-etal-2019-commonsenseqa}}          & {Require  Commonsense (World knowledge)}          \\
		\hline {2016}     & {WikiReading \cite{hewlett-etal-2016-wikireading}}           & {Require  Commonsense (External knowledge) ; Large Scale Dataset}     \\
		\hline {2016}     & {WikiMovies \cite{miller2016WikiMovies}}           & {Require  Commonsense (External knowledge) ; Domain-specific}       \\
		\hline {2016}     & {MovieQA \cite{tapaswi2016movieqa}}             & {Multi-Modal  MRC}            \\
		\hline {2017}     & {COMICS \cite{iyyer2017COMICS}}             & {Multi-Modal  MRC}            \\
		\hline {2017}     & {TQA \cite{kembhavi2017tqa}}               & {Multi-Modal  MRC}            \\
		\hline {2018}     & {RecipeQA \cite{yagcioglu-etal-2018-recipeqa}}            & {Multi-Modal  MRC}            \\
		\hline {2018}     & {HotpotQA \cite{yang-etal-2018-hotpotqa}}            & {Multi-hop  MRC ; Complex Reasoning}            \\
		\hline {2017}     & {NarrativeQA \cite{kovcisky2018narrativeqa}}           & {Multi-hop  MRC ; Complex Reasoning}            \\
		\hline {2017}     & {Qangaroo \cite{welbl2018qangaroo}}            & {Multi-hop  MRC}             \\
		\hline {2018}     & {MultiRC \cite{khashabi-etal-2018MultiRC}}             & {Multi-hop  MRC}             \\
		\hline {2015}     & {CNN/Daily Mail \cite{hermann2015cnn}}        & {Large-scale Dataset}          \\
		\hline {2016}     & {BookTest \cite{bajgar2016embracing}}            & {Large-scale Dataset}          \\
		\hline {2013}     & {MCTest \cite{richardson2013mctest}}             & {For  Open-domain QA}          \\
		\hline {2015}     & {CuratedTREC \cite{baudivs2015CuratedTREC}}           & {For  Open-domain QA}          \\
		\hline {2017}     & {Quasar \cite{Dhingra2017Quasar}}             & {For Open-domain  QA}          \\
		\hline {2017}     & {SearchQA \cite{dunn2017searchqa}}            & {For  Open-domain QA}          \\
		\hline {2017}     & {SciQ \cite{welbl2017SciQ}}              & {Domain-specific}             \\
		\hline {2018}     & {CliCR \cite{suster-daelemans-2018-clicr}}              & {Domain-specific}             \\
		\hline {2018}     & {PaperQA(Hong et al.) \cite{hong2018learning}} & {Domain-specific}             \\
		\hline {2018}     & {PaperQA(Park et al.) \cite{park2019can}}    & {Domain-specific}             \\
		\hline {2018}     & {ReviewQA \cite{grail2018reviewqa}}            & {Domain-specific}             \\
		\hline {2018}     & {SciTail \cite{khot2018scitail}}             & {Domain-specific}             \\
		\hline {2019}     & {DROP \cite{Dua2019DROP}}              & {Complex  Reasoning}           \\
		\hline {2016}     & {Facebook CBT \cite{hill2016cbt}}         & {Complex  Reasoning}           \\
		\hline {2016}     & {Google  MC-AFP}         & {Complex  Reasoning}           \\
		\hline {2016}     & {LAMBADA \cite{boleda2016lambada}}             & {Complex  Reasoning}           \\
		\hline {2016}     & {NewsQA \cite{trischler-etal-2017-newsqa}}             & {Complex  Reasoning}           \\
		\hline {2016}     & {SQuAD 1.1 \cite{rajpurkar2016squad}}            & {Complex  Reasoning}           \\
		\hline {2017}     & {RACE \cite{lai-etal-2017-race}}              & {Complex  Reasoning}           \\
		\hline {2017}     & {TriviaQA \cite{joshi-etal-2017-triviaqa}}            & {Complex  Reasoning}           \\
		\hline {2018}     & {CLOTH \cite{xie2018cloth}}              & {Complex  Reasoning}           \\
		\hline {2018}     & {ProPara \cite{dalvi-etal-2018-ProPara}}             & {Complex  Reasoning}     \\
		\hline {2019}     & {DREAM \cite{sun2019dream}}              & {Conversational  MRC ; Require Commonsense (World knowledge)}       \\
	\hline {2018}     & {CoQA \cite{reddy2019coqa}}              & {Conversational  MRC ; With Unanswerable Questions}          \\
	\hline {2018}     & {QuAC \cite{choi-etal-2018-quac}}              & {Conversational  MRC ; With Unanswerable Questions}          \\
	\hline {2019}     & {ShARC \cite{saeidi2018sharc}}              & {Conversational  MRC}          \\		
		
	\end{supertabular}
\end{center}

\subsubsection{MRC with Unanswerable Questions}
The existing MRC datasets often lack training sets for unanswerable questions, which weaken the robustness of the MRC systems. As a result, when the MRC models answer unanswerable questions, the models always try to give a most likely answer, rather than refuse to answer these unanswered questions. In this way, no matter how the model answers, the answers must be wrong.\\
To solve this problem, the researchers proposed many MRC datasets with unanswerable questions which were more challenging. Among the datasets collected by us, the datasets that contain unanswerable questions include: SQuAD 2.0, MS MARCO \cite{tri2016MARCO}, Natural Questions \cite{kwiatkowski2019natural} and NewsQA \cite{trischler-etal-2017-newsqa}. We will give a detailed description of these datasets in section in section~\ref{sec: Descriptions}.\\

\subsubsection{Multi-hop Reading Comprehension}
In most MRC dataset, the answer to a question usually can be found in a single paragraph or a document. However, in real human reading comprehension, when reading a novel, we are very likely to extract answers from multiple paragraphs. Compared with single passage MRC, the multi-hop machine reading comprehension is more challenging and requires multi-hop searching and reasoning over confusing passages or documents. \\
In different papers, multi-hop MRC is named in different ways such as multi-document machine reading comprehension \cite{yan2019deep}, multi-paragraph machine reading comprehension \cite{wang-etal-2018-multi-passage}, multi-sentence machine reading comprehension \cite{khashabi-etal-2018MultiRC}. Compared with single paragraph MRC, multi-hop MRC is more challenging and is naturally suitable for unstructured information processing.
Among the datasets collected by us, the datasets that contain unanswerable questions including SQuAD 2.0 \cite{rajpurkar-etal-2018-know}, MS MARCO \cite{tri2016MARCO}, Natural Questions \cite{kwiatkowski2019natural}, and NewsQA \cite{trischler-etal-2017-newsqa}. 

\subsubsection{Multi-modal Reading Comprehension}
When humans read, they often do it in a multi-modal way. For example, in order to understand the information and answer the questions, sometimes, we need to read both the texts and illustrations, and we also need to use our brains to imagine, reconstruct, reason, calculate, analyze or compare. Currently, most of the existing machine reading comprehension datasets belong to plain textual machine reading comprehension, which has some limitations. some complex or precise concepts can not be described or communicated only via text. For example, if we need the computer to answer some precise questions related to aircraft engine maintenance, we may have to input the image of the aircraft engine. \\
Multi-modal machine reading comprehension is a dynamic interdisciplinary field that has great application potential. Considering the heterogeneity of data, multi-modal machine reading comprehension brings unique challenges to NLP researchers, because the model has to understand both texts and images. In recent years, due to the availability of large-scale internet data, many multi-modal MRC datasets have been created, such as TQA \cite{kembhavi2017tqa}, RecipeQA \cite{yagcioglu-etal-2018-recipeqa}, COMICS \cite{iyyer2017COMICS}, and MovieQA \cite{tapaswi2016movieqa}. 

\subsubsection{Reading Comprehension Require Commonsense or World knowledge}
Human language is complex. When answering questions, we often need to draw upon our commonsense or world knowledge. Moreover, in the process of human language, many conventional puns and polysemous words have been formed. The use of the same words in different scenes also requires the computer to have a good command of the relevant commonsense or world knowledge.\\
Conventional MRC tasks usually focus on answering questions about given passages. In the existing machine reading comprehension datasets, only a small proportion of questions need to be answered with commonsense knowledge. In order to build MRC models with commonsense or world knowledge, many Commonsense Reading Comprehension (CRC) datasets have been created, such as CommonSenseQA \cite{talmor-etal-2019-commonsenseqa}, ReCoRD \cite{zhang2018record} and OpenBookQA \cite{OpenBookQA2018}.

\subsubsection{Complex Reasoning MRC}
The reasoning is an innate ability of human beings, which can be embodied in logical thinking, reading comprehension, and other activities. The reasoning is also a key component in artificial intelligence and a fundamental goal of MRC. In recent years, reasoning has been an essential topic among the MRC community. We hope that the MRC system can not only read and learn the representation of the language but also can really understand the context and answer complex questions. In order to push towards complex reasoning MRC system, many datasets have been generated, such as Facebook bAbI \cite{weston2016babi}, DROP \cite{Dua2019DROP},RACE \cite{lai-etal-2017-race}, and CLOTH \cite{xie2018cloth}.

\subsubsection{Conversational Reading Comprehension}
It is a natural way for human beings to exchange information through a series of conversations. In the typical MRC tasks, different question and answer pairs are usually independent of each other. However, in real human language communication, we often achieve an efficient understanding of complex information through a series of interrelated conversations. Similarly, in human communication scenarios, we often ask questions on our own initiative, to obtain key information that helps us understand the situation. In the process of conversation, we need to have a deep understanding of the previous conversations in order to answer each other's questions correctly or ask meaningful new questions. Therefore, in this process, historical conversation information also becomes a part of the context.\\
In recent years, conversational machine reading comprehension (CMRC) has become a new research hotspot in the NLP community, and there emerged many related datasets, such as CoQA \cite{reddy2019coqa}, QuAC \cite{choi-etal-2018-quac}, DREAM \cite{sun2019dream} and ShARC \cite{saeidi2018sharc}.

\subsubsection{Domain-specific Datasets}
In this paper, a domain-specific dataset refers to the MRC dataset whose context comes from a particular domain, such as science examinations, movies, clinical reports. Therefore, the neural network models trained by those datasets usually can be directly applied to a certain field. For example, CliCR \cite{suster-daelemans-2018-clicr} is a cloze MRC dataset in the medical domain. There are approximately 100,000 cloze questions about the clinical case reports. SciQ \cite{welbl2017SciQ} is a multiple-choice MRC dataset containing 13.7K crowdsourced science exam questions about physics, chemistry and biology, and others. The context and questions of SciQ are derived from scientific exam questions. In addition, domain-specific datasets also include ReviewQA \cite{grail2018reviewqa}, SciTail \cite{khot2018scitail}, WikiMovies \cite{miller2016WikiMovies}, PaperQA \cite{park2019can}. 

\subsubsection{MRC with Paraphrased Paragraph}
Paragraph paraphrasing refers to rewriting or rephrasing a paragraph using different words, while still conveying the same messages as before. The MRC dataset with paraphrased paragraph has at least two versions of context which expresses the same meanings while there is little word overlap between the different versions of context. The task of paraphrased MRC requires the computer to answer questions about contexts. To answer these questions correctly, the computer needs to understand the true meaning of different versions of context. So far, we only find that the DuoRC \cite{saha-etal-2018-duorc} and Who-did-What \cite{onishi2016whodidwhat} are datasets of this type.

\subsubsection{Large-scale MRC Dataset}
The size of the early MRC dataset is usually not very large, such as QA4MRE, CuratedTREC \cite{baudivs2015CuratedTREC}, MCTest \cite{richardson2013mctest}. With the emergence of large-scale datasets, MRC is greatly promoted due to the possibility of neural network models training.

\subsubsection{MRC dataset for Open-Domain QA}
The open-domain question answering was originally defined as finding answers in collections of unstructured documents \cite{chen-etal-2017-reading}. With the development of MRC research, many MRC datasets tend to be used to solve open-domain QA. The release of new MRC datasets such as MCTest \cite{richardson2013mctest}, CuratedTREC \cite{baudivs2015CuratedTREC}, Quasar \cite{Dhingra2017Quasar}, SearchQA \cite{dunn2017searchqa} greatly promotes open-domain QA recently.

\subsection{Descriptions of each MRC dataset}
\label{sec: Descriptions}

In section~\ref{sec:Characteristics}, we introduced the characteristics of various machine reading comprehension datasets. In this section, we will give a detailed description of the 47 MRC datasets collected in our survey with their download links available. Then we will describe them according to the order of datasets in Table \ref{table: characteristics}.

\subsubsection{WikiQA}
The WikiQA \cite{yang2015wikiqa} dataset contains a large number of real Bing query logs as the question-answer pair and provided links to Wikipedia passages that might have answers in the dataset. Differs from previous datasets such as QASENT, questions in WIKIQA were sampled from real queries of Bing without editorial revision. The WikiQA dataset also contains questions that can not actually be answered from the given passages, so the machine is required to detect these unanswerable questions. The WikiQA was created by crowd-workers and contains 3,047 questions and 29,258 sentences, in which 1,473 sentences were marked as answer sentences for the question \cite{yang2015wikiqa}. The WikiQA dataset is available on \href{https://www.microsoft.com/en-us/download/details.aspx?id=52419}{https://www.microsoft.com/en-us/download/details.aspx?id=52419}.

\subsubsection{SQuAD 2.0}
SQuAD 2.0 \cite{rajpurkar-etal-2018-know} is the latest version of the Stanford Question Answering Dataset (SQuAD). SQuAD 2.0 combines the data from the existing version of SQuAD 1.1 \cite{rajpurkar2016squad} with more than 50,000 unanswerable questions written by crowd workers. To acquire a good performance on SQuAD 2.0, the MRC model not only needs to answer questions when possible, but also needs to identify issues without correct answers in the context and not to answer them \cite{rajpurkar-etal-2018-know}. For existing models, SQuAD 2.0 is a challenging natural language understanding task. The author also compares the test data of similar model architecture in SQuAD 1.1. Compared with SQuAD 1.1, the gap between human accuracy and machine accuracy in SQuAD 2.0 is much larger, which confirms that square2.0 is a more difficult data set for existing models. As mentioned in the authors' paper, the powerful nervous model that achieved 86\% F1 on SQuAD 1.1 received only 66\% of F1 on SQuAD 2.0. Data for both SQuAD 1.1 and SQuAD 2.0 are available on \href{https://rajpurkar.github.io/SQuAD-explorer/}{https://rajpurkar.github.io/SQuAD-explorer/}.

\subsubsection{Natural Questions}
Natural Questions \cite{kwiatkowski2019natural} is a MRC dataset with unanswerable questions. The samples in this dataset come from real anonymous questions and answers in the Google search engine. The dataset is manually generated by the crowd workers. The MRC model presents the crowd worker with a question and related Wikipedia pages and requires the crowd worker to mark a long answer (usually a paragraph) and a short answer (usually one or more entities) on the page or mark null if there is no correct answer. The Natural Questions dataset consists of 307,373 training samples with single annotations, 7,830 samples with 5-way annotations for development data, and 7,842 test examples with 5-way annotations \cite{kwiatkowski2019natural}. The dataset can be downloaded at \href{https://github.com/google-research-datasets/natural-questions}{https://github.com/google-research-datasets/natural-questions}, which also has a link to the leaderboard.

\subsubsection{MS MARCO}
MS MARCO \cite{tri2016MARCO} is a large-scale machine reading comprehension dataset containing unanswerable questions. The dataset consists of 1,010,916 questions and answers collected from Bing's search query logs. Besides, the dataset contains 8,841,823 paragraphs extracted from 3,563,535 Web documents retrieved by Bing, which provide the information for answering questions. MS MARCO contains three different tasks: (1) Identify unanswerable questions; (2) Answer the question if it is answerable; (3) Rank a set of retrieved passages given a question \cite{tri2016MARCO}. The MRC model needs to estimate whether these paragraphs contain correct answers, and then sort them depending on how close they are to the answers. The dataset and leaderboard of MS MARCO are available on \href{http://www.msmarco.org/}{http://www.msmarco.org/}.

\subsubsection{DuoRC}
DuoRC \cite{saha-etal-2018-duorc} is a MRC dataset which contains 186,089  question-answer pairs generated from 7,680 pairs of movie plots. Each pair of movie plots reflects two versions of the same movie: one from Wikipedia and the other from IMDb. The texts of these two versions are written by two different authors. In the process of building question-answer pairs, the authors require crowd workers to create questions from one version of the story and a different set of crowd workers to extract or synthesize answers from another version. This is the unique feature of DuoRC in which there is almost no vocabulary overlap between the two versions. Additionally, the narrative style of the paragraphs generated from the movie plots (compare to the typical descriptive paragraphs in the existing dataset) indicates the need for complex reasoning of events in multiple sentences \cite{tri2016MARCO}. DuoRC is a challenging dataset, and the authors observed that the state-of-the-art model on the SQuAD 1.1 \cite{rajpurkar2016squad} also performed poorly on DuoRC, with F1 score of 37.42\% while 86\% on SQuAD 1.1. The dataset, paper and, leaderboard of DuoRC can be obtained at \href{https://duorc.github.io/}{https://duorc.github.io/}.

\subsubsection{Who-did-What}
The Who-did-What \cite{onishi2016whodidwhat} dataset contains more than 200,000 fill-in-the-gap (cloze) multiple-choice reading comprehension questions constructed from the LDC English Gigaword newswire corpus. Compared to other existing machine reading comprehension datasets, such as CNN/Daily Mail \cite{hermann2015cnn}, the Who-did-What dataset avoided using the same article summaries to create a sample in the dataset. Instead, each sample is formed by two separate articles. One article is given as the passage to be read and the other article on the same events is used to form the question. Second, the authors avoided anonymization — each choice is a person named entity. Third, the questions that can be easily solved by simple baselines have been removed, while humans can still solve 84\% of the questions \cite{onishi2016whodidwhat}. The dataset and leaderboard of Who-did-What are available on \href{https://tticnlp.github.io/who_did_what/index.html}{https://tticnlp.github.io/who\_did\_what/index.html}.

\subsubsection{ARC}
AI2 Reasoning Challenge (ARC) \cite{Clark2018ARC} is a MRC dataset and task to encourage AI research in question answering that requires deep reasoning. To finish the ARC task, the MRC model requires far more powerful knowledge and reasoning than previous challenges such as SQuAD \cite{rajpurkar2016squad,rajpurkar-etal-2018-know} or SNLI \cite{snli2015SNLI}. The ARC dataset contains 7,787 elementary-level scientific questions that are in the form of multiple-choices. The dataset is divided into a Challenge Set and an Easy Set, where the Challenge Set only contains questions that are not correctly answered by both a retrieval-based algorithm and a word co-occurrence algorithm. The ARC dataset contains only natural, primary-level science questions (written for the human exam) and is the largest collection of such datasets. The authors tested several baselines on the Challenge Set, including state-of-the-art models from the SQuAD and SNLI, and found that none of them were significantly better than the random baseline, reflecting the difficulty of the task. The author also publishes the ARC corpus, which is a corpus of 14M scientific sentences related to this task, and the implementation of three neural baseline models tested \cite{Clark2018ARC}. Information about the ARC dataset and leaderboards is available on \href{http://data.allenai.org/arc/}{http://data.allenai.org/arc/}.

\subsubsection{MCScript}
MCScript \cite{ostermann-etal-2018-mcscript} is a large-scale MRC dataset with narrative texts and questions that require reasoning using commonsense knowledge. The dataset focuses on narrative texts about everyday activities, and the commonsense knowledge are required to answer multiple-choice questions based on these texts. The feature of the MCScript dataset is to evaluate the contribution of script knowledge to machine understanding. A script is a series of events (also called scenarios) that describe human behavior. The MCScript dataset also forms the basis of a shared task on commonsense and script knowledge organized at SemEval 2018 \cite{ostermann-etal-2018-mcscript}. The official web page and CodaLab competition page of the SemEval 2018 Shared Task 11 are available on \href{https://competitions.codalab.org/competitions/17184}{https://competitions.codalab.org/competitions/17184}.

\subsubsection{OpenBookQA}
OpenBookQA \cite{OpenBookQA2018} consists of about 6,000 elementary level science questions in the form of multi-choice (4,957 training sets, 500 validation sets, and 500 test sets). Answering the questions in OpenBookQA requires broad common knowledge. OpenBookQA also requires a deeper understanding of both the topic (in the context of common knowledge) and the language it is expressed in \cite{OpenBookQA2018}. The baseline model provided by the authors has reached about 50\% in this dataset, but many state-of-the-art pre-trained QA methods perform surprisingly even worse \cite{OpenBookQA2018}. Dataset and leaderboard of OpenBookQA are available on \href{https://leaderboard.allenai.org/open_book_qa/}{https://leaderboard.allenai.org/open\_book\_qa/}.

\subsubsection{ReCoRD}
ReCoRD \cite{zhang2018record} is a large-scale MRC dataset that requires deep commonsense reasoning. Experiments on the ReCoRD show that the performance of the state-of-the-art MRC model lags far behind human performance. The ReCoRD represents the challenge of future research to bridge the gap between human and machine commonsense reading comprehension. The ReCoRD dataset contains more than 120,000 queries from over 70,000 news articles. Each query has been verified by crowd workers \cite{zhang2018record}. The feature of the data set is that all queries and paragraphs in the records are automatically mined from news articles, which minimizes the artificially induced bias. So most records need deep commonsense reasoning. Since July 2019, the ReCoRD has been added to SuperGLUE as an evaluation suite. The ReCoRD dataset and leaderboard are available on \href{https://sheng-z.github.io/ReCoRD-explorer/}{https://sheng-z.github.io/ReCoRD-explorer/}.

\subsubsection{CommonSenseQA}
CommonSenseQA \cite{talmor-etal-2019-commonsenseqa} is a MRC dataset that requires different types of commonsense knowledge to predict the correct answer. It contains 12,247 questions. The CommonSenseQA dataset is split into a training set, validation set and, test set. The authors performed two types of splits: "Random split" which is the main evaluation split, and "Question token split" where each of the three sets has disjoint question concepts \cite{talmor-etal-2019-commonsenseqa}. To capture common sense beyond association, the authors of CommonSenseQA extracted multiple target concepts from Conceptnet 5.5 \cite{speer2017conceptnet} that have the same semantic relationship to a single source concept. Crowd workers were asked to propose multiple-choice questions, mention source concepts, and then distinguished each goal concept. This encouraged crowd workers to ask questions with complex semantics that often require prior knowledge \cite{talmor-etal-2019-commonsenseqa}. The dataset and leaderboard of CommonSenseQA are available on \href{https://www.tau-nlp.org/commonsenseqa}{https://www.tau-nlp.org/commonsenseqa}.

\subsubsection{WikiReading}
WikiReading \cite{hewlett-etal-2016-wikireading} is a large-scale machine reading comprehension dataset that contains 18 million instances. The dataset consists of 4.7 million unique Wikipedia articles, which means that about 80\% of the English language Wikipedia is represented. The WikiReading dataset is composed of a variety of challenging classification and extraction subtasks, which makes it very suitable for neural network models. In the WikiReading dataset, multiple instances can share the same document, with an average of 5.31 instances per article (median: 4, maximum: 879). The most common document categories are humans, categories, movies, albums, and human settlements, accounting for 48.8\% of documents and 9.1\% of instances respectively. The average and median document lengths are 489.2 and 203 words \cite{hewlett-etal-2016-wikireading}. The WikiReading dataset is available on \href{https://github.com/google-research-datasets/wiki-reading}{https://github.com/google-research-datasets/wiki-reading}.

\subsubsection{WikiMovies}
WikiMovies \cite{miller2016WikiMovies} is a MRC dataset with Wikipedia documents. To compare using Knowledge Bases (KBs), information extraction or Wikipedia documents directly in a single framework, the author built the WikiMovies dataset which contains raw texts and preprocessed KBs. WikiMovies is part of FaceBook's bAbI project, and information about the BABi project is available on \href{https://research.fb.com/downloads/babi/}{https://research.fb.com/downloads/babi/}, and the WikiMovies dataset is available on \href{http://www.thespermwhale.com/jaseweston/babi/movieqa.tar.gz}{http://www.thespermwhale.com/jaseweston/babi/movieqa.tar.gz}.

\subsubsection{MovieQA}
The MovieQA \cite{tapaswi2016movieqa} dataset is a multi-modal machine reading comprehension dataset designed to evaluate the automatic understanding of both pictures and texts. The dataset contains 14,944 questions from 408 movies. The types of questions in the MovieQA dataset are multiple-choice, and the questions range from simpler "Who" did "What" to "Whom", to "Why" and "How" certain events occurred. The MovieQA dataset is unique because it contains multiple sources of information-video clips, episodes, scripts, subtitles, and DVS \cite{tapaswi2016movieqa}. Download links and evaluation benchmarks of the MovieQA dataset can be obtained for free from \href{http://movieqa.cs.toronto.edu/home/}{http://movieqa.cs.toronto.edu/home/}.

\subsubsection{COMICS}
COMICS \cite{iyyer2017COMICS} is a multi-modal machine reading comprehension dataset, which is composed of more than 1.2 million comic panels (120 GB) and automatic text box transcriptions. In the COMICS task, the machine is required to read and understand the text and images in the comic panels at the same time. Besides the traditional textual cloze tasks, the authors also designed two novel MRC tasks (visual cloze, and character coherence) to test the model's ability to understand narratives and characters in a given context \cite{iyyer2017COMICS}. The dataset and baseline of COMICS are available on \href{https://obj.umiacs.umd.edu/comics/index.html}{https://obj.umiacs.umd.edu/comics/index.html}.

\subsubsection{TQA}
The TQA \cite{kembhavi2017tqa} (Textbook Question Answering) challenge encourages multi-modal machine reading (M3C) tasks. Compared with Visual Question Answering (VQA) \cite{antol2015vqa}, the TQA task provides the multi-modal context and question-answer pair which consists of text and images. TQA dataset is constructed from the science curricula of middle school. The textual and diagrammatic content in middle school science reference fairly complex phenomena that occur in the world. Many questions need not only simple search, but also complex analysis and reasoning of multi-mode context.\\
The TQA dataset consists of 1,076 courses and 26,260 multi-modal questions \cite{kembhavi2017tqa}. The analysis shows that a high proportion of questions in the TQA dataset require complex text analysis, graphing, and reasoning, which indicates that the TQA dataset is related to previous machine understanding and VQA dataset \cite{antol2015vqa} The TQA dataset and leaderboards are available on \href{http://vuchallenge.org/tqa.html}{http://vuchallenge.org/tqa.html}.

\subsubsection{RecipeQA}
RecipeQA \cite{yagcioglu-etal-2018-recipeqa} is a MRC dataset for multi-modal comprehension of recipes. It consists of about 20K instructional recipes with both texts and images and more than 36K automatically generated question-answer pairs. RecipeQA is a challenging multi-modal dataset for evaluating reasoning on real-life cooking recipes. The RecipeQAtask consists of many specific tasks. A sample in RecipeQA contains a multi-modal context, such as headings, descriptions, or images. To find an answer, the model needs (i) a joint understanding of the pictures and texts; (ii) capturing the temporal flow of events; and (iii) understanding procedural knowledge \cite{yagcioglu-etal-2018-recipeqa}. The dataset and leaderboard of RecipeQA are available on \href{http://hucvl.github.io/recipeqa}{http://hucvl.github.io/recipeqa}.

\subsubsection{HotpotQA}
HotpotQA \cite{yang-etal-2018-hotpotqa} is a multi-hop MRC dataset with multi-paragraphs. There are 113k Wikipedia-based QA pairs in HotpotQA. Different from other MRC datasets, In the HotpotQA, the model is required to perform complex reasoning and provide explanations for answers from multi-paragraphs. HotpotQA has four key features: (1) the questions require the machine to read and reason over multiple supporting documents to find the answer; (2) The questions are diverse and not subject to any pre-existing knowledge base; (3) The authors provided sentence-level supporting facts required for reasoning; (4) The authors offered a new type of factoid comparison questions to test QA systems’ ability to extract relevant facts and perform necessary comparison \cite{yang-etal-2018-hotpotqa}. Dataset and leaderboard of HotpotQA are publicly available on \href{https://hotpotqa.github.io/}{https://hotpotqa.github.io/}.

\subsubsection{NarrativeQA}
NarrativeQA \cite{kovcisky2018narrativeqa} is a multi-paragraph machine reading comprehension dataset and a set of tasks. To encourage progress on deeper comprehension of language, the authors designed the NarrativeQA dataset. Unlike other datasets in which the questions can be solved by selecting answers using superficial information, in the NarrativeQA, the machine is required to answer questions about the story by reading the entire book or movie script. In order to successfully answer questions, the model needs to understand the underlying narrative rather than relying on shallow pattern matching or salience \cite{kovcisky2018narrativeqa}. NarrativeQA is available on \href{https://github.com/deepmind/narrativeqa}{https://github.com/deepmind/narrativeqa}.

\subsubsection{Qangaroo}
Qangaroo \cite{welbl2018qangaroo} is a multi-hop machine reading comprehension dataset. Most reading comprehension methods limit themself to questions that can be answered using a single sentence, paragraph, or document \cite{welbl2018qangaroo}. Therefore, the authors of Qangaro proposed a new task and dataset to encourage the development of text understanding models across multiple documents and to study the limitations of existing methods. In the Qangaroo task, the model is required to seek and combine evidence – effectively performing multihop,
alias multi-step, inference \cite{welbl2018qangaroo}. The dataset, papers, and leaderboard of Qangaroo are publicly available on \href{http://qangaroo.cs.ucl.ac.uk/index.html}{http://qangaroo.cs.ucl.ac.uk/index.html}.

\subsubsection{MultiRC}
MultiRC (Multi-Sentence Reading Comprehension) \cite{khashabi-etal-2018MultiRC} is a MRC dataset in which questions can only be answered by considering information from multiple sentences. The purpose of creating this dataset is to encourage the research community to explore more useful methods than complex lexical matching. MultiRC consists of about 6,000 questions from more than 800 paragraphs across 7 different areas (primary science, news, travel guides, event stories, etc.) \cite{khashabi-etal-2018MultiRC}. MultiRC is available on \href{http://cogcomp.org/multirc/}{http://cogcomp.org/multirc/}. Since May 2019, MultiRC is part of SuperGLUE, so the authors will no longer provide the leaderboard on the above website. 

\subsubsection{CNN/Daily Mail}
In order to solve the problem of lack of large-scale datasets, Hermann et al. \cite{hermann2015cnn} created a new dataset generation method that provided a large-scale supervised reading comprehension dataset in 2015. They also extracted text from the websites of CNN and Daily Mail and created two MRC datasets, which is the CNN/Daily Mail \cite{hermann2015cnn} dataset. In the CNN dataset, there are 90,266 documents and 380,298 questions. The Daily Mail dataset consist of 196,961 documents and 879,450 questions. The creation of the CNN/Daily Mail dataset allows the community to develop a class of attention based deep neural networks that learn to read real documents and answer complex questions with minimal prior knowledge of language structure \cite{hermann2015cnn}. The CNN/Daily Mail dataset and related materials are available on \href{https://github.com/deepmind/rc-data}{https://github.com/deepmind/rc-data}.

\subsubsection{BookTest}
The BookTest \cite{bajgar2016embracing} is a large-scale MRC dataset with 14,140,825 training examples and 7,917,523,807 tokens. The BookTest dataset is derived from books available through the project Gutenberg \cite{hartproject}. The training dataset contains the original CBT NE and CN data \cite{hill2016cbt} and extends the new NE and CN examples. The authors of BookTest extracted 10,507 books for NE instances from the project Gutenberg and also used 3,555 copyright-free books to extract CN instances \cite{bajgar2016embracing}. The BookTest dataset can be downloaded from \href{https://ibm.biz/booktest-v1}{https://ibm.biz/booktest-v1}.

\subsubsection{MCTest}
In MCTest \cite{richardson2013mctest} dataset, the model is required to answer multiple-choice questions about fictional stories, directly tackling the high-level goal of open-domain machine comprehension. The stories and questions of MCTest are also carefully limited to those a young child would understand, reducing the world knowledge that is required for the task \cite{richardson2013mctest}. The data in MCTest was gathered using Amazon Mechanical Turk. Since the answer is a fictional story, the content of the answer is very broad and not limited to a certain field. Therefore, the MRC model trained by MCTest is helpful for the open-domain question answering research \cite{richardson2013mctest}. The MCTest dataset and leaderboards are available on \href{https://mattr1.github.io/MCTest/}{https://mattr1.github.io/MCTest/}.

\subsubsection{CuratedTREC}
The CuratedTREC \cite{baudivs2015CuratedTREC} dataset is a curated version of the TREC corpus \cite{TRECproject}. The Text REtrieval Conference (TREC) \cite{TRECproject} was started in 1992 by the U.S. Department of Defense and the National Institute of Standards and Technology (NIST). Its purpose was to support the research of the information retrieval system. The large version of CuratedTREC is based on the QA tasks of TREC 1999, 2000, 2001 and 2002 which have been curated by Baudiš and JŠedivý \cite{baudivs2015CuratedTREC} and contains a total of 2,180 questions. CuratedTREC is also used to evaluate the ability of the machine reading comprehension model to answer open-domain questions \cite{chen-etal-2017-reading,raison2018weaver,sampathqa}. The TREC corpus is available in \href{https://github.com/brmson/dataset-factoid-curated}{https://github.com/brmson/dataset-factoid-curated}.

\subsubsection{Quasar}
Quasar \cite{Dhingra2017Quasar} is a MRC dataset for open-domain questions, it contains two sub-datasets: Quasar-T and Quasar-S. Quasar is designed to evaluate the model's ability of understanding natural language queries and extract answers from large amounts of texts. The Quasar-S dataset consists of 37,000 cloze-style questions, and the Quasar-T dataset contains 43,000 open-domain trivia issues questions. ClueWeb09 \cite{callan2009clueweb09} serves as a background corpus for extracting these answers. 
The Quasar dataset is a challenge to two related sub-tasks of the factoid questions: (1) searching for relevant text segments containing the correct answers to the query, and (2) reading the retrieved passages to answer the questions \cite{Dhingra2017Quasar}. The dataset and paper of Quasar are available on \href{https://github.com/bdhingra/quasar}{https://github.com/bdhingra/quasar}.

\subsubsection{SearchQA}
SearchQA \cite{dunn2017searchqa} is a MRC dataset with retrieval systems. To answer open-domain questions in SearchQA, the model needs to read the text retrieved by the search engine, so it can also be regarded as a machine reading comprehension dataset. The question-answer pairs in the SearchQA dataset are all collected from the J!Archive, and the context is retrieved from Google. SearchQA consists of more than 140k QA pairs, with an average of 49.6 clips per pair. Each QA environment tuple in SearchQA comes with additional metadata, such as the URL of the fragment, which the authors believe will be a valuable resource for future research. The authors perform a manual evaluation on SearchQA and test two baseline methods, one simple word selection, and another deep learning \cite{dunn2017searchqa}. The paper suggests that the SearchQA can be obtained at \href{https://github.com/nyu-dl/SearchQA}{https://github.com/nyu-dl/SearchQA}.

\subsubsection{SciQ}
SciQ \cite{welbl2017SciQ} is a domain-specific multiple-choice MRC dataset containing 13.7K crowdsourced science questions about Physics, Chemistry, and Biology, etc. The context and questions are derived from real 4th and 8th-grade exam questions. The questions are in the form of multiple-choices, with an average of four choices for each question. For the majority of the questions, an additional paragraph with supporting evidence for the correct answer is provided. In addition, the authors proposed a new method for generating domain-specific multiple-choice MRC datasets from crowd workers \cite{welbl2017SciQ}. The SciQ dataset can be downloaded at \href{http://data.allenai.org/sciq/}{http://data.allenai.org/sciq/}.

\subsubsection{CliCR}
CliCR \cite{suster-daelemans-2018-clicr} is a cloze MRC dataset in the medical domain. There are approximately 100,000 cloze questions about the clinical case reports. The authors applied several baselines and a state-of-the-art neural model and observed the performance gap (20\% F1) between the human and the best neural models \cite{suster-daelemans-2018-clicr}. They also analyzed the skills required to correctly answer the question and explained how the model's performance changes based on the applicable skills, and they found that reasoning using domain knowledge and object tracking is the most frequently needed skill, and identifying missing information and spatiotemporal reasoning is the most difficult for machines \cite{suster-daelemans-2018-clicr}. The code of the baseline project can be publicly available on \href{https://github.com/clips/clicr}{https://github.com/clips/clicr}, where the author claims that the CliCR dataset can be obtained by contacting the author via email.

\subsubsection{PaperQA (Hong et al.,2018)}
PaperQA \cite{hong2018learning} created by Hong et al. is a MRC dataset containing more than 6,000 human-generated question-answer pairs about academic knowledge. To build the PaperQA, crowd workers have provided questions based on more than 1,000 abstracts of the research paper on deep learning, and their answers that consist of text spans of the related abstracts. The authors collected the PaperQA through a four-stage process to acquire QA pairs that require reasoning. And they have proposed a semantic segmentation model to solve this task \cite{hong2018learning}. PaperQA is publicly available on \href{http://bit.ly/PaperQA}{http://bit.ly/PaperQA}.

\subsubsection{PaperQA (Park et al.,2018)}
In order to measure the machine's ability of understanding professional-level scientific papers, a domain-specific MRC dataset called PaperQA \cite{park2019can} was created. PaperQA consists of over 80,000 cloze questions from research papers. The authors of PaperQA performed ﬁne-grained linguistic analysis and evaluation to compare PaperQA and other conventional question and answering (QA) tasks on general literature (e.g., books, news, and Wikipedia), and the results indicated that the PaperQA task is difficult, showing there is ample room for future research \cite{park2019can}. According to the authors' paper, PaperQA had been published on \href{http://dmis.korea.ac.kr/downloads?id=PaperQA}{http://dmis.korea.ac.kr/downloads?id=PaperQA}, but when we visited this website, it was not available at that moment.

\subsubsection{ReviewQA}
ReviewQA \cite{grail2018reviewqa} is a domain-specific MRC dataset about hotel reviews. ReviewQA contains over 500,000 natural questions and 100,000 hotel reviews. The authors hope to improve the relationship understanding ability of the machine reading comprehension model by constructing the ReviewQA dataset. Each question in ReviewQA is related to a set of relationship understanding capabilities that the model is expected to master \cite{grail2018reviewqa}. The ReviewQA dataset, summary of the tasks, and results of models are available on \href{https://github.com/qgrail/ReviewQA/}{https://github.com/qgrail/ReviewQA/}.

\subsubsection{SciTail}
The SciTail \cite{khot2018scitail} is a textual entailment dataset which consists of multiple-choices QA pairs about scientific exams and web sentences. The dataset consists of 27,026 examples, of which 10,101 examples contain entails labels and 16,925 examples contain neutral labels. Different from existing datasets, SciTail was created solely from natural sentences that already exist independently "in the wild" rather than sentences authored speciﬁcally for the entailment task \cite{khot2018scitail}. The authors also generated hypotheses from questions, the relevant answer options, and premises from related web sentences from a large corpus \cite{khot2018scitail}. Baseline and leaderboard of SciTail are available on \href{https://leaderboard.allenai.org/scitail/submissions/public}{https://leaderboard.allenai.org/scitail/submissions/public}. The SciTail dataset is available on \href{http://data.allenai.org/scitail/}{http://data.allenai.org/scitail/}.

\subsubsection{DROP}
DROP \cite{Dua2019DROP} is an English MRC dataset that requires the Discrete Reasoning Over the content of Paragraphs. The DROP dataset contains 96k questions created by crowd workers. Unlike the existing MRC task, in the DROP, the MRC model is required to resolve references in a question, and perform discrete operations on them (such as adding, counting, or sorting) \cite{Dua2019DROP}. These operations require a deeper understanding of the content of paragraphs than what was necessary for prior datasets \cite{Dua2019DROP}. The dataset of DROP can be downloaded at \href{https://s3-us-west-2.amazonaws.com/allennlp/datasets/drop/drop_dataset.zip}{https://s3-us-west-2.amazonaws.com/allennlp/datasets/drop/drop\_dataset.zip}. The Leaderboard is available on \href{https://leaderboard.allenai.org/drop}{https://leaderboard.allenai.org/drop}.

\subsubsection{Facebook CBT}
Children's Book Test (CBT) \cite{hill2016cbt} is a MRC dataset that uses children's books as context. Each sample in the CBT dataset contains 21 consecutive sentences, the first 20 sentences become the context, and a word is deleted from the 21st sentence, so it becomes a cloze question. MRC model is required to identify the answer word among a selection of 10 candidate answers appearing in the context sentences and the question. 
Different from standard language-modeling tasks, CBT distinguishes the task of predicting syntactic function words from that of predicting lower-frequency words, which carry greater semantic content \cite{hill2016cbt}. The CBT dataset is part of FaceBook's bAbI project which is available on \href{https://research.fb.com/downloads/babi/}{https://research.fb.com/downloads/babi/}.The Children’s Book Test (CBT) dataset can be downloaded at \href{http://www.thespermwhale.com/jaseweston/babi/CBTest.tgz}{http://www.thespermwhale.com/jaseweston/babi/CBTest.tgz}.

\subsubsection{Google MC-AFP}
Google MC-AFP \cite{soricut2016building} is a MRC dataset which has about 2 million examples. It is generated from the AFP portion of LDC’s English Gigaword corpus \cite{graff2003english}. The authors of MC-AFP also provided a new method for creating large-scale MRC datasets using paragraph vector models. In the MC-AFP, the upper limit of accuracy achieved by human testers is approximately 91\%. Among all models tested by the authors, the authors' hybrid neural network architecture achieves the highest accuracy of 83.2\%. The remaining gap to the human-performance ceiling provides enough room for future model improvements \cite{soricut2016building}. Google MC-AFP is available on \href{https://github.com/google/mcafp}{ https://github.com/google/mcafp}.

\subsubsection{LAMBADA}
The main task of the LAMBADA \cite{boleda2016lambada} is to read the text and predict the missing last word. The authors of LAMBADA hoped to encourage the development of new models capable of genuine understanding of broad context in natural language text \cite{boleda2016lambada}, therefore, it can also be understood as a MRC task. The LAMBADA dataset consists of narrative passages in which human subjects could guess the last word if they read the whole paragraph, but not if they only read the last sentence preceding the answer word. In order to get high scores in LAMBADA, the models have to track information in a wider discourse. For the above reasons, LAMBADA as a challenging dataset and exempliﬁes a wide range of linguistic phenomena \cite{boleda2016lambada}. LAMBADA can be obtained at \href{https://zenodo.org/record/2630551}{ https://zenodo.org/record/2630551}.

\subsubsection{NewsQA}
NewsQA \cite{trischler-etal-2017-newsqa} is a new MRC dataset that contains more than 100,000 natural instances. Crowd workers provide questions and answers based on more than 10,000 news articles from CNN, in which the answer is a text span of the related news article. The authors collected NewsQA through a four-stage process to seek exploratory question-answer pairs that require reasoning. The authors also stratified reasoning categories in NewsQA, including word matching, paraphrasing, inference, synthesis, ambiguous/insufficient. The NewsQA requires the ability to go beyond simple word matching and recognizing textual entailment. The authors measured human performance on NewsQA and compared it to several powerful neural models. The performance gap between humans and the MRC model (0.198 in F1) suggested that significant progress could be made on NewsQA through future research \cite{trischler-etal-2017-newsqa}. The NewsQA dataset and model leaderboards are available for free at \href{https://www.microsoft.com/en-us/research/project/newsqa-dataset/}{https://www.microsoft.com/en-us/research/project/newsqa-dataset/}.

\subsubsection{SQuAD 1.1}
The Stanford Question Answering Dataset (SQuAD) \cite{rajpurkar2016squad} is a well-known machine reading comprehension dataset that contains more than 100,000 questions generated by crowd-workers, in which the answer of each question is a segment of text from the related paragraph \cite{rajpurkar2016squad}. Since it was released in 2016, SQuAD 1.1 quickly became the most widely used MRC dataset. Now it has been updated to SQuAD 2.0 \cite{rajpurkar-etal-2018-know}. In the leaderboards of SQuAD 1.1 and SQuAD 2.0, we have witnessed the birth of a series of state-of-the-art neural models, such as BiDAF \cite{seo2016bidirectional}, BERT \cite{devlin2018bert}, RoBERTa \cite{liu2019roberta} and XLNet \cite{yang2019xlnet}, etc.  The data and leaderboard of SQuAD 1.1 and SQuAD 2.0 are available on \href{https://rajpurkar. github.io/SQuAD-explorer/}{https://rajpurkar. github.io/SQuAD-explorer/}.

\subsubsection{RACE}
RACE \cite{lai-etal-2017-race} is a MRC dataset collected from the English exams for Chinese students. There are approximately 28,000 articles and 100,000 questions provided by humans (English teachers), covering a variety of carefully designed topics to test students' understanding and reasoning ability. Different from the existing MRC dataset, the proportion of questions that need reasoning ability in RACE is much large than other MRC datasets, and there is a great gap between the performance of the state-of-the-art models (43\%) and the best human performance (95\%) \cite{lai-etal-2017-race}. The authors hope that this new dataset can be used as a valuable resource for machine understanding research and evaluation \cite{lai-etal-2017-race}. The dataset of RACE is available on \href{http://www.cs.cmu.edu/~glai1/data/race/}{http://www.cs.cmu.edu/~glai1/data/race/}.The baseline project is available on \href{https://github.com/qizhex/RACE_AR_baselines}{https://github.com/qizhex/RACE\_AR\_baselines}.

\subsubsection{TriviaQA}
TriviaQA \cite{joshi-etal-2017-triviaqa} is a challenging MRC dataset, which contains more than 650k question-answer pairs and their evidence. TriviaQA has many advantages over other existing MRC datasets: (1) relatively complex combinatorial questions; (2) considerable syntactic and lexical variability between the questions and the related passages; (3) more cross sentence reasoning is required to answer the question \cite{joshi-etal-2017-triviaqa}. The TriviaQA dataset and baseline project are available on \href{http://nlp.cs.washington.edu/triviaqa/}{http://nlp.cs.washington.edu/triviaqa/} and information about the Codalab competition of TriviaQA is available on \href{https://competitions.codalab.org/competitions/17208}{https://competitions.codalab.org/competitions/17208}.

\subsubsection{CLOTH}
CLOTH \cite{xie2018cloth} is a large-scale cloze MRC dataset with 7,131 passages and 99,433 questions collected from English examinations. CLOTH requires a deeper language understanding of multiple aspects of natural language including reasoning, vocabulary and grammar. In addition, CLOTH can be used to evaluate language models' abilities in modeling long text \cite{xie2018cloth}. CLOTH's leaderboard is available on \href{http://www.qizhexie.com/data/CLOTH_leaderboard}{ http://www.qizhexie.com/data/CLOTH\_leaderboard} and dataset can be downloaded from \href{http://www.cs.cmu.edu/~glai1/data/cloth/}{http://www.cs.cmu.edu/~glai1/data/cloth/}. The code of baseline project can be downloaded at \href{https://github.com/qizhex/Large-scale-Cloze-Test-Dataset-Created-by-Teachers}{https://github.com/qizhex/Large-scale-Cloze-Test-Dataset-Created-by-Teachers}.

\subsubsection{ProPara}
ProPara \cite{dalvi-etal-2018-ProPara} is a MRC dataset for understanding contexts about processes (such as photosynthesis). In the ProPara task, the model is required to identify the actions described in the procedural text and tracking the state changes that have occurred to the entities involved. The ProPara dataset contains 488 paragraphs and 3,300 sentences (about 81,000 notes) generated by crowd workers. The purpose of creating ProPara is to predict the presence and location of each participant based on the sentences in the context \cite{dalvi-etal-2018-ProPara}. 
The dataset of Propara can be downloaded from \href{http://data.allenai.org/propara}{http://data.allenai.org/propara}, and the leaderboard of Propara is available on \href{https://leaderboard.allenai.org/propara/submissions/public}{https://leaderboard.allenai.org/propara/submissions/public}.

\subsubsection{DREAM}
DREAM \cite{sun2019dream} is a conversational, multiple-choice MRC dataset. The dataset was collected from English exam questions designed by human experts to evaluate the reading comprehension level of English learners. The DREAM dataset consists of 10,197 questions in the form of multiple-choice with a total of 6,444 dialogues. Compared to the existing conversational reading comprehension (CRC) dataset, DREAM is the first to focus on in-depth multi-turn multi-party dialogue understanding \cite{sun2019dream}. In the DREAM dataset, 84\% of answers are non-extractive, 85\% require more than one sentence of reasoning, and 34\% of questions involve common sense knowledge.
DREAM's authors applied several neural models on DREAM that used surface information in the text and found that they could barely surpass rule-based methods. In addition, the authors also studied the effects of incorporating dialogue structures and different types of general world knowledge into several models on the DREAM dataset. The experimental results demonstrated the effectiveness of the dialogue structure and general world knowledge \cite{sun2019dream}. DREAM is available on: \href{https://dataset.org/dream/}{https://dataset.org/dream/}.

\subsubsection{CoQA}
CoQA \cite{reddy2019coqa} is a conversational MRC dataset that contains 127K questions and answers from 8k dialogues in 7 different fields. Through an in-depth analysis of CoQA, the authors showed that conversational questions in CoQA have challenging phenomena that are not presented in existing MRC datasets, such as coreference and pragmatic reasoning. The authors also evaluated a set of state-of-the-art conversational MRC models on CoQA. The best F1 score achieved by those models is 65.1\%, and human performance is 88.8\%, indicating that there was plenty of room for future advance \cite{reddy2019coqa}. Dataset and leaderboard of CoQA can be found at \href{https://stanfordnlp.github.io/coqa/}{https://stanfordnlp.github.io/coqa/}.

\subsubsection{QuAC}
QuAC \cite{choi-etal-2018-quac} is a conversational MRC dataset containing about 100K questions from 14K information-seeking QA dialogs. Each dialogue in QuAC involves two crowd workers: (1) One act like a student who asks a few questions to learn a hidden passage from Wikipedia, and (2) the other one act as a teacher to answer questions by providing a brief excerpt from the Wikipedia passage. The QuAC dataset introduced the challenges that not present in existing MRC datasets: its questions are often more open-ended, unanswerable, or meaningful only in a dialog environment \cite{choi-etal-2018-quac}. The authors also reported the performance of many state-of-the-art models on QuAC, and the best result was 20\% lower than human F1, suggesting there was ample room for future research \cite{choi-etal-2018-quac}. Dataset, baseline and leaderboard of QuAC can be found at \href{http://quac.ai}{http://quac.ai}.

\subsubsection{ShARC}
ShARC \cite{saeidi2018sharc} is a conversational MRC dataset. Unlike existing conversational MRC datasets, when answering questions in the ShARC, the model needs to use background knowledge that is not in the context to get the correct answer. The first question in a ShARC conversation is usually not fully explained and does not provide enough information to answer directly. Therefore, the model needs to take the initiative to ask the second question, and after the model has got enough information, it then answers the first question \cite{saeidi2018sharc}. The dataset, paper, and leaderboard of ShARC are available on \href{https://sharc-data.github.io}{https://sharc-data.github.io}.

\section{Open Issues}
\label{sec:Issues}
In recent years, great progress has been made in the field of MRC due to large-scale datasets and effective deep neural network approaches. However, there are still many issues remaining in this field. In this section, we describe these issues in the following aspects:

\subsection{What needs to be improved?}
Nowadays, the neural machine reading models have exceeded the human performance scores on many MRC datasets. However, state-of-the-art models are still far from human-level language understanding. What needs to be improved on existing tasks and datasets? We believe that there are many important aspects that have been overlooked which merit additional research. Here we list several areas as below:

\subsubsection{Multi-modal MRC}
A fundamental characteristic of human language understanding is multimodality. Psychologists examined the role of mental imagery skills on story comprehension in fifth graders (10- to 12-year-olds). Experiments showed that children with higher mental imagery skills outperformed children with lower mental imagery skills on story comprehension after reading the experimental narrative \cite{boerma2016reading}.
Our observation and experience of the world bring us a lot of common sense and world knowledge, and the multi-modal information is extremely important for us to acquire such common sense and world knowledge. 
However, it is currently not clear how our brains store, encode, represent, and process knowledge, which is an important scientific problem in cognitive neuroscience, philosophy, psychology, artificial intelligence and other fields. At present, the research in the field of natural language processing mainly focuses on the pure textual corpus, but in neuroscience, the research methods are very different. Since the 1990s, cognitive neuroscientists have found that knowledge extraction could activate the widely distributed cerebral cortex, including the sensory cortex and the motor cortex \cite{kemmerer2014cognitive}. More and more cognitive neuroscientists believe that concepts are rooted in modality-specific representations \cite{kemmerer2014cognitive}. This is usually called Grounded Cognition Model \cite{barsalou2008grounded, pecher2005grounding}, or Embodied Cognition Model \cite{kemmerer2014cognitive,semin2008embodied,gibbs2005embodiment}. The key idea is that semantic knowledge does not exist in an abstract domain completely separated from perception and behavior, but overlaps these capabilities to some extent.
\cite{kemmerer2014cognitive, barsalou1999perceptual, shapiro2019embodied}.
In that case, can we still make computers really understand human languages only by the neural network training of pure textual corpus? Nowadays, although there are already a few of multi-modal MRC datasets, the related research is still insufficient. The number of current multi-modal MRC datasets are still small, and these datasets simply put pictures and texts together, lacking detailed annotations and internal connections. How to make better use of multi-modal information is an important research area in the future.

\subsubsection{Commonsense and World Knowledge}
Commonsense and world knowledge are the main bottlenecks in machine reading comprehension. Among different kinds of commonsense and world knowledge, two types of commonsense knowledge are considered fundamental for human reasoning and decision making: intuitive psychology and intuitive physics \cite{storks2019commonsense}.
Although there are some MRC datasets about commonsense, such as CommonSenseQA \cite{talmor-etal-2019-commonsenseqa}, ReCoRD \cite{zhang2018record}, DREAM \cite{sun2019dream}, OpenBookQA \cite{OpenBookQA2018}, this field is still in a very early stage. In these datasets, there is no strict division of commonsense types, nor research on commonsense acquisition methods combined with psychology.
Understanding how the commonsense knowledge is acquired in the process of human growth may help to reveal the computing model of commonsense.
Observing the world is the first step for us to acquire commonsense and world knowledge. For example, "this book can't be put into a school bag, it's too small" and "this book can't be put into a schoolbag, it's too big". In these two sentences, human beings can know from commonsense that the former "it" refers to a school bag, and the latter "it" refers to a book. But this is not intuitive for computers.
Human beings receive a great deal of multi-modal information in our daily life, which forms commonsense. When the given information is insufficient, we can make up the gap by predicting. Correct prediction is the core function of our commonsense.
In order to gain real understanding ability comparable to human beings, machine reading comprehension models must need massive data to provide commonsense and world knowledge. Algorithms are needed to get a better commonsense corpus and we need to create multi-modal MRC datasets to help machines acquire commonsense and world knowledge.

\subsubsection{Complex Reasoning}
Many of the existing MRC datasets are relatively simple. In these datasets, the answers are short, usually a word or a phrase. Many of the questions can be answered by understanding a single sentence in the context, and there are very few datasets that need multi-sentences reasoning \cite{chen2018neural}. This shows that most of the samples in existing MRC datasets are lack of complex reasoning.
In addition, researchers found that after input-ablation, many of the answers in existing MRC datasets are still correct \cite{sugawara2019assessing}. This shows that many existing benchmark datasets do not really require the machine reading comprehension model to have reasoning skills. From this perspective, high-quality MRC datasets that need complex reasoning is needed to test the reasoning skill of MRC modals.

\subsubsection{Robustness}

Robustness is one of the key desired properties of a MRC model. Jia and Liang \cite{jia-liang-2017-adversarial} found that existing benchmark datasets are overly lenient on models that rely on superficial cues \cite{chen2018neural,liu2019neural}. They tested whether MRC systems can answer questions that contain distracting sentences. In their experiment, a distracting sentence that contains words that overlap with the question was added at the end of the context. These distracting sentences will not mislead human understanding, but the average scores of the sixteen models on SQuAD will be significantly reduced. This shows that these state-of-the-art MRC models still rely too much on superficial cues, and there is still a huge gap between MRC and human-level reading comprehension \cite{jia-liang-2017-adversarial}. How to avoid the above situation and improve the robustness of MRC model is an important challenge.

\subsubsection{Interpretability}
In the existing MRC tasks, the model is only required to give the answer to the question directly, without explaining why it gets the answer. So it is very difficult to really understand how the model makes decisions \cite{chen2018neural,liu2019neural}. Regardless of whether the complete interpretability of these models is absolutely necessary, it is fair to say that a certain degree of understanding of the internal model can greatly guide the design of neural network structure in the future. In future MRC datasets, sub-tasks can be set up to let the model give the reasoning process, or the evidence used in reasoning.

\subsubsection{Evaluation of the quality of MRC datasets?}
There are many evaluation metrics for machine reading comprehension models, such as F1, EM, accuracy, etc. However, different MRC datasets also need to be evaluated. How to evaluate the quality of MRC datasets? 
One metric of MRC dataset is the readability. The classical measures of readability are based on crude approximations of the syntactic complexity (using the average sentence length as a proxy) and lexical complexity (average length in characters or syllables of words in a sentence). One of the most well-known measures along these lines is the Flesch-Kincaid readability index \cite{kincaid1975derivation} which combines these two measures into a global score \cite{benzahra2019measuring}. 
However, recent studies have shown that the readability of MRC dataset is not directly related to the question difficulty \cite{sugawara2017evaluation}.
The experiment results suggest that while the complexity of datasets is decreasing, the performance of MRC model will not be improved to the same extent and the correlation is quite small \cite{benzahra2019measuring}. Another possible metric is the frequencies of different prerequisite skills needed in MRC datasets. Sugawara et al. defined 10 prerequisite skills \cite{sugawara2017evaluation}, including Object tracking, Mathematical reasoning, Coreference resolution, Analog, Causal relation, etc. However, the definition of prerequisite skills is often arbitrary and changeable. Different definitions can be drawn from different perspectives \cite{sugawara2016analysis,sugawara2017evaluation,sugawara2019assessing}.
Moreover, at present, the frequency of prerequisite skills is still manually counted, and there is no automated statistical method. In summary, how to evaluate the quality of MRC datasets is still an unsolved problem.

\subsection{Have we understood understanding?}
The word "understanding" has been used by human beings for thousands of years \cite{hough2019understanding,jackson2020toward}. But, what is the exact meaning of understanding? What are the specific neural processes of understanding? 

Many researchers attempted to give definitions of understanding. For example, Hough and Gluck \cite{hough2019understanding} conducted an extensive survey of literature about understanding. They summarized: 
\begin{quote}
	"In an attempt to summarize the preceding review, we propose the following general definition for the process and outcome of understanding: The acquisition, organization, and appropriate use of knowledge to produce a response directed towards a goal, when that action is taken with awareness of its perceived purpose."
\end{quote}

But understanding is too natural and complex for us as it is difficult to define, especially from different perspectives such as philosophy, psychology, pedagogy, neuroscience, computer science, etc. In the field of NLP, we still lack a comprehensive definition of understanding of language and also lack of specific metrics to evaluate the real understanding capabilities of MRC models.

In recent years, great progress has been made in the field of cognitive neuroscience of language. Thanks to the advanced neuroimaging technologies such as PET and fMRI, contemporary cognitive neuroscientists have been able to study and describe large-scale cortical networks related to language in various ways, and they have found many interesting findings. Just taking understanding object nouns as an example. How are these object nouns represented in the brain? As David Kemmerer summarized in his book \cite{kemmerer2014cognitive}:
\begin{quote}
"From roughly the 1970s through the 1990s, the dominant theory of conceptual knowledge was the Amodal Symbolic Model. It emerged from earlier developments in logic, formal linguistics, and computer science, and its central claim was that concepts, including word meanings, consist entirely of abstract symbols that are represented and processed in an autonomous semantic system that is completely separate from the modality specific systems for perception and action\cite{fodor1975language,smith1978theories,pylyshyn1984computation}.\\
Since 1990s, the Grounded Cognition Model has been attracting increasing interest. The key idea is that semantic knowledge does not reside in an abstract realm that is totally segregated from perception and action, but instead overlaps with those capacities to some degree.
To return to the banana example mentioned above, understanding this object noun is assumed to involve activating modality-specific records in long-term memory that capture generalizations about how bananas look, how they taste, how they feel in one’s hands, how they are manipulated, etc. This theory maintains that conceptual processing amounts to recapitulating modality-specific states, albeit in a manner that draws mainly on high-level rather than low-level components of the perceptual and motor systems \cite{kemmerer2014cognitive}."
\end{quote}

In addition, a recent study \cite{wang2020two} published in the Cell reveals that the two hypothesis theories mentioned above are both right. The authors studied the brain basis of color knowledge in sighted individuals and congenitally blind individuals whose color knowledge can only be obtained through language descriptions. Their experiments show that congenitally blind individuals can obtain knowledge representation similar to healthy people through language without any sensory experience. And more importantly, they also found that there are two different coding systems in the brain of sighted individuals: one is directly related to the sense, in the visual color processing brain area; the other is in the left anterior temporal lobe dorsal side, the same as the memory brain area of knowledge obtained only through language in congenitally blind individuals \cite{wang2020two}. According to their study, there are (at least) two forms of object knowledge representations in the human brain: sensory-derived and cognitively-derived knowledge, supported by different brain systems \cite{wang2020two}. It also shows that human language is not only used to express symbols for communication, but also to encode conceptual knowledge. 

So, can we get a more effective MRC model through training multi-modal corpus? Probably. But, due to the complexity of the human brain, cognitive neuroscientists are still unable to fully understand the details of natural language understanding. But these cognitive neuroscience studies have brought a lot of inspiration to the NLP community. We could make full use of the existing research results of cognitive neuroscience to design novel MRC systems.


\section{Conclusions}
\label{sec:Conclusions}
We conducted a comprehensive survey of recent efforts on the tasks, evaluation metrics, and benchmark datasets of machine reading comprehension (MRC). We discussed the definition and taxonomy of MRC tasks and proposed a new classification method for MRC tasks. The computing methods of different MRC evaluation metrics have been introduced with their usage in each type of MRC tasks also analyzed. We also introduced attributes and characteristics of MRC datasets, with 47 MRC datasets described in detail. 
Finally, we discussed the open issues for future research of MRC and we argued that high-quality multi-modal MRC datasets and the research findings of cognitive neuroscience may help us find better ways to construct more challenging datasets and develop related MRC algorithms to achieve the ultimate goal of human-level machine reading comprehension. To facilitate the MRC community, we have published the above data on the companion website \href{https://mrc-datasets.github.io/}{(https://mrc-datasets.github.io/)}, from where MRC researchers could directly access the MRC datasets, papers, baseline projects and browse the leaderboards.

\vspace{6pt} 



\authorcontributions{Conceptualization, C.Z., J.H., and S.L.; methodology, C.Z. ;  investigation, C.Z, J.H.; resources, S.L.; data curation, C.Z.; writing--original draft preparation, C.Z., J.H.; writing--review and editing, C.Z, J.H., Q.L.; visualization, C.Z. and Jie.H.; supervision, S.L. and J.H.; project administration, S.L. and J.H.; funding acquisition, S.L. All authors have read and agreed to the published version of the manuscript.}

\funding{
Research reported in this publication was partially supported by the Major research project of the National Natural Science Foundation of China under grant number 91746116 and National Major Scientific and Technological Special Project of China under grant number 2018AAA0101803. This work is also partially supported by Guizhou Province Science and Technology Project under grant number [2015] 4011. 
}

\conflictsofinterest{The authors declare no conflict of interest.}

\reftitle{References}
\externalbibliography{yes}
\bibliography{MyLibrary}






\end{document}